\documentclass{article}

\PassOptionsToPackage{numbers, compress}{natbib}

\usepackage[preprint, eandd]{neurips_2026}

\usepackage[utf8]{inputenc} 
\usepackage[T1]{fontenc}    
\usepackage{hyperref}       
\hypersetup{
    hidelinks
}
\usepackage{url}            
\usepackage{booktabs}       
\usepackage{amsfonts}       
\usepackage{nicefrac}       
\usepackage{microtype}      
\usepackage{xcolor}         

\usepackage{graphicx}
\usepackage{tabularx}
\usepackage{array}
\usepackage{pifont}
\usepackage{makecell}
\usepackage{multirow}
\usepackage{arydshln}
\usepackage{amsmath}
\usepackage{subcaption}
\usepackage[table]{xcolor}
\usepackage{colortbl}
\usepackage{color}
\usepackage{titletoc}
\definecolor{WrongOrange}{HTML}{E55300}
\definecolor{CorrectGreen}{HTML}{74AF44}
\definecolor{PromptGray}{HTML}{6B6B6B}
\title{SenseBench: A Benchmark for Remote Sensing Low-Level Visual Perception and Description in Large Vision-Language Models}

\author{
    Chen Zhong$^{1*}$~
    Xiao An$^{1*}$,~
    Jiaxing Sun$^{2}$,~
    Zihan Gui$^{1}$,~
    Guangyi Yang$^{1}$~
    Wei He$^{1\dag}$ \\
    $^1$ Wuhan University~
    $^2$ Shanghai Artificial Intelligent Laboratory \\
    \texttt{sunjiaxing@pjlab.org.cn}\\
    \texttt{\{chen\_zhong, anxiao, sunjiax0323, syro\_gzh, weihe1990\}@whu.edu.cn}
}

\begin{document}

\maketitle

\begin{abstract}
  \renewcommand{\thefootnote}{}
    \makeatletter\def\Hy@Warning#1{}\makeatother
    \footnotetext{
    $^*$Equal contribution. 
    $^\dag$Corresponding author.}
Low-level visual perception underpins reliable remote sensing (RS) image analysis, yet current image quality assessment (IQA) methods output uninterpretable scalar scores rather than characterizing physics-driven RS degradations, deviating markedly from the diagnostic needs of RS experts. While Vision-Language Models (VLMs) present a compelling alternative by delivering language-grounded IQA, their visual priors are heavily biased toward ground-level natural images. Consequently, whether VLMs can overcome this domain gap to perceive and articulate RS artifacts remains insufficiently studied. To bridge this gap, we propose \textbf{SenseBench}, the first dedicated diagnostic benchmark for RS low-level visual perception and description. Driven by a physics-based hierarchical taxonomy that unifies both non-reference and reference-based paradigms, SenseBench features over 10K meticulously curated instances across 6 major and 22 fine-grained RS degradation categories. Specifically, two complementary protocols are designed for evaluation: objective low-level visual \textit{perception} and subjective diagnostic \textit{description}. Comprehensive evaluation of 29 state-of-the-art VLMs reveals not only skewed domain priors and multi-distortion collapse, but also \textit{fluency illusion} and a \textit{perception-description inversion} effect. We hope SenseBench provides a robust evaluation testbed and high-quality diagnostic data to advance the development of VLMs in RS low-level perception. Code and datasets are available \href{https://github.com/Zhong-Chenchen/SenseBench}{\textcolor{blue}{here}}.
\end{abstract}
\section{Introduction}

Robust low-level visual perception is a fundamental prerequisite for reliable remote sensing (RS) image analysis~\cite{lu2025vision}. Throughout acquisition, transmission, and processing, RS imagery is inevitably corrupted by highly complex, physics-driven degradations~\cite{zhu2017deep, tuia2022perspectives}. Domain-specific distortions, ranging from atmospheric interference to specialized sensor noise, severely compromise the vital underlying visual structure~\cite{wei2025bpme}, making accurate image quality assessment (IQA) essential for successful downstream interpretation. While conventional IQA methods attempt to address this challenge by regressing scalar quality scores~\cite{musiq, liqe, clipiqa}, they fundamentally suffer from two critical limitations. First, their focus remains narrowly confined to generic distortions (e.g., Gaussian noise, motion blur, or JPEG compression), inherently failing to capture the intricate, compound degradations endemic to diverse RS scenarios. Second, practical RS workflows demand granular degradation analysis to guide downstream restoration or image selection. By collapsing multifaceted visual quality into a single numerical proxy, conventional methods bypass explicit perceptual reasoning, thereby losing semantic interpretability and failing to provide the actionable diagnostic insights required by RS practitioners to make informed, reliable decisions in complex real-world scenarios~\cite{li2024vision}.

Large Vision-Language Models (VLMs) emerge as a powerful paradigm to bridge this interpretability gap. Empowered by massive multimodal pretraining and instruction tuning, general-domain models such as Q-Instruct~\cite{qinstruct} and DepictQA~\cite{depictqa} excel at detecting artifacts, reasoning about degradation severity, and generating fine-grained linguistic diagnostics. While these capabilities make VLMs theoretically ideal for RS quality assessment, their internal visual priors are overwhelmingly anchored to ground-level, natural photography. Conversely, RS imagery presents drastically different spatial statistics, spectral signatures~\cite{an2024pretrain}, and complex, physics-driven degradation mechanisms that lack counterparts in standard datasets~\cite{huang2025task}. Consequently, it remains completely unresolved whether current VLMs can overcome this profound domain shift to reliably perceive and articulate specialized RS distortions. As Table~\ref{tab:compare_table} illustrates, existing RS benchmarks predominantly target high-level semantic recognition or constrain themselves to superficial artifact categories. This conspicuous absence of a dedicated diagnostic benchmark leaves the low-level perceptual limits of VLMs in the RS domain entirely unquantified, thereby obstructing systematic progress in this critical area~\cite{reobench, vrsbench}.


\begin{table}[!t]
\centering
\caption{Comparison of SenseBench with Existing Benchmarks. \textit{Dimension} means the number of fine-grained evaluation capabilities. Abbreviations adopted: MCQ for Multi-Choice Question; FF for Free-Form; BBox for Bounding Box; Seg for Segmentation Mask; L-L for Low Level.}
\vspace{0.5\baselineskip}
\label{tab:compare_table}

\scriptsize
\renewcommand{\arraystretch}{1.15}
\setlength{\tabcolsep}{3pt}

\resizebox{\textwidth}{!}{%
\begin{tabular}{c c c c c c c c}
\Xhline{1.2pt}
\textbf{Benchmark} &
\textbf{Domain} &
\textbf{Data Sources} &
\textbf{New-Data} &
\textbf{Geospatial Coverage} &
\textbf{Answer Type} &
\textbf{Multi-image} &
\textbf{Dims.} \\ 
\Xhline{1.2pt}

MMBench~\cite{mmbench} & General & Multiple Public Datasets & \ding{55} & N/A & MCQ & \ding{55} & 20 \\
MMStar~\cite{mmstar} & General & Multiple Public Datasets & \ding{55} & N/A & MCQ & \ding{55} & 18 \\
Q-Bench(+)~\cite{qbench+} & General L-L & Multiple Public Datasets & \ding{55} & N/A & MCQ, FF & \ding{51} & N/A \\
DepictQA-Wild~\cite{you2025enhancing} & General L-L & Multiple Public Datasets & \ding{51} & N/A & MCQ, FF & \ding{51} & 4 \\
MICBench~\cite{wu2024towards} & General L-L & Multiple Public Datasets & \ding{51} & N/A & MCQ, FF & \ding{51} & N/A \\ 
\hline

LHRS-Bench~\cite{lhrs} & RS & Google Earth+OSM & \ding{51} & N/A & MCQ & \ding{55} & 11 \\
GeoChat-Bench~\cite{geochat} & RS & SAMRS & \ding{55} & Multiple Regions & FF, BBox & \ding{55} & 6 \\
VRSBench~\cite{vrsbench} & RS & DOTA, DIOR & \ding{55} & N/A & FF, BBox & \ding{55} & 3 \\
CHOICE~\cite{choice} & RS & Diverse Platforms & \ding{51} & 50 Cities Worldwide & MCQ, BBox, Seg & \ding{51} & 23 \\
DisasterM3~\cite{disasterm3} & RS Disaster & xBD, BRIGHT, \emph{etc.} & \ding{51} & N/A & MCQ, BBox, Seg, FF & \ding{51} & 12 \\
DynamicVL~\cite{dynamicvl} & RS City & NAIP & \ding{51} & USA & MCQ, Seg, FF & \ding{51} & 7 \\
REOBench~\cite{reobench} & RS Robust & AID, DIOR, \emph{etc.} & \ding{55} & N/A & MCQ, BBox, Seg, FF & \ding{55} & 6 \\ 
\Xhline{1.2pt}

\textbf{SenseBench (Ours)} &
\textbf{RS L-L} &
\textbf{Diverse Platforms} &
\ding{51} &
\textbf{Worldwide} &
\textbf{MCQ, FF} &
\ding{51} &
\textbf{22} \\ 
\Xhline{1.2pt}
\end{tabular}%
}
\vspace{-5mm}
\end{table}

To systematically address these challenges and answer the core research question---\textit{To what extent can current VLMs robustly perceive and understand physics-driven RS degradations?}---we propose \textbf{SenseBench}, the first diagnostic benchmark explicitly designed to quantify the low-level visual perception and description capabilities of VLMs in the RS domain. As illustrated in Figure~\ref{fig:worldwide_taxonomy}, SenseBench covers 6 major categories and 22 fine-grained sub-categories, and unifies both no-reference and reference-based paradigms through a rigorous multi-stage construction pipeline. Specifically, (1) \textbf{Global Data Sampling:} To mitigate regional bias, pristine reference images are systematically curated across six continents via population-weighted geospatial sampling. (2) \textbf{Physics-Based Distortion Synthesis:} We employ a parameter-controlled degradation engine to inject diverse artifacts, guaranteeing deterministic and fully traceable ground-truth labels for objective perception assessment. (3) \textbf{Dual-Perspective Protocol:} Moving beyond conventional single-image inspection (non-reference evaluation), we introduce pairwise comparative analysis (full-reference evaluation), faithfully reflecting real-world demands for selecting optimal multi-temporal observations and evaluating restoration algorithms. (4) \textbf{GPT-Assisted Diagnostic Generation:} Beyond simple multiple-choice questions, we utilize a constrained LLM pipeline to synthesize open-ended quality inspection reports, which subsequently undergo strict human-in-the-loop verification to serve as semantic ground truths.

Leveraging this comprehensive testing framework, we conduct a systematic zero-shot evaluation of 29 state-of-the-art VLMs. The experimental results reveal three critical structural limitations in current models: (1) \textit{Skewed domain priors:} General VLMs heavily favor natural photographic degradations, while RS-specialized models overfit to RS-centric distortions, with neither achieving a balanced visual foundation. (2) \textit{Multi-distortion collapse:} Model robustness degrades exponentially when confronting compound degradations, indicating that entangled artifact understanding is not a trivial extension of isolated distortion recognition. (3) \textit{Fluency illusion and perception-description inversion effect:} VLMs frequently generate linguistically confident but visually ungrounded diagnostic reports, and paradoxically exhibit a severe disconnect between their objective perceptual accuracy and subjective descriptive capabilities, exposing a profound misalignment between textual priors and authentic visual reasoning.

\section{Related Work}

\textbf{Image Quality Assessment.} Blind IQA has advanced from handcrafted statistical models~\cite{brisque, niqe} to transformer architectures~\cite{musiq}, recently further enhanced with vision-language pretraining that enables zero-shot estimation and cross-dataset generalization (e.g., CLIP-IQA~\cite{clipiqa}, LIQE~\cite{liqe}). However, these methods are developed exclusively on natural photographic content, creating a fundamental mismatch with RS imagery. RS degradations are physics-driven~\cite{zhu2017deep}, including atmospheric interference, sensor noise, and spectral distortions, with acquisition and scene statistics fundamentally distinct from natural images. Existing RS IQA work addresses narrow degradation subsets via feature engineering~\cite{yan2019nriqa}, block-matching~\cite{bm_iqe}, task-specific databases~\cite{prsiqd}, and domain models~\cite{nriqa_uav_hsi}. Yet all reduce quality to scalar scores, forfeiting semantic interpretability and diagnostic reasoning. This is precisely what RS practitioners require to guide restoration and image selection.

\textbf{Vision-Language Models.} VLMs now extend beyond high-level semantic reasoning to fine-grained low-level visual understanding~\cite{qwen3vl, internvl3.5, llava}. Through instruction tuning, models such as Q-Instruct~\cite{qinstruct} and DepictQA~\cite{depictqa} detect artifacts, reason about degradation severity, and produce diagnostic descriptions aligned with human judgment~\cite{qalign, Q-Scorer}. Such capabilities position VLMs as promising candidates for interpretable RS quality assessment, yet general VLMs inherit biases toward natural images. Specialized VLMs for RS have emerged across visual grounding~\cite{geochat, hu2025rsgpt}, multi-sensor fusion~\cite{soni2025earthdial, yao2025falcon}, temporal reasoning~\cite{irvin2024teochat}, and scalable architectures~\cite{lin2025rs, liu2024rsunivlm, liu2025skymoe}, but remain trained under the implicit assumption of clean imaging conditions. Whether current VLMs can perceive and articulate the physics-driven RS degradations remains insufficiently studied.

\textbf{Evaluation Benchmarks.} For natural images, Q-Bench(+)~\cite{qbench+} and DepictQA-Wild~\cite{you2025enhancing} established rigorous evaluation protocols for low-level perception and description, systematically exposing critical gaps between model capabilities and human judgment. Conversely, RS benchmarks have proliferated across high-level semantic tasks, including hierarchical reasoning~\cite{choice}, object-level understanding~\cite{vrsbench}, disaster response~\cite{disasterm3}, and dynamic scenes~\cite{dynamicvl}, but remain silent on fine-grained low-level perception. Even rare exceptions like REOBench~\cite{reobench} treat image quality as a mere confounding variable rather than the true subject of evaluation. Because complex RS degradations arise from distinct physical mechanisms, the direct transfer of natural-image benchmarks proves fundamentally inadequate. SenseBench elegantly fills this paramount void as the pioneering dedicated framework for evaluating VLM low-level perception and diagnostic description on RS imagery.

\vspace{-1mm}
\section{SenseBench}
\vspace{-1mm}

\label{sec:construction}

\begin{figure}[!t]
    \centering
    \resizebox{\textwidth}{!}{%
        \includegraphics{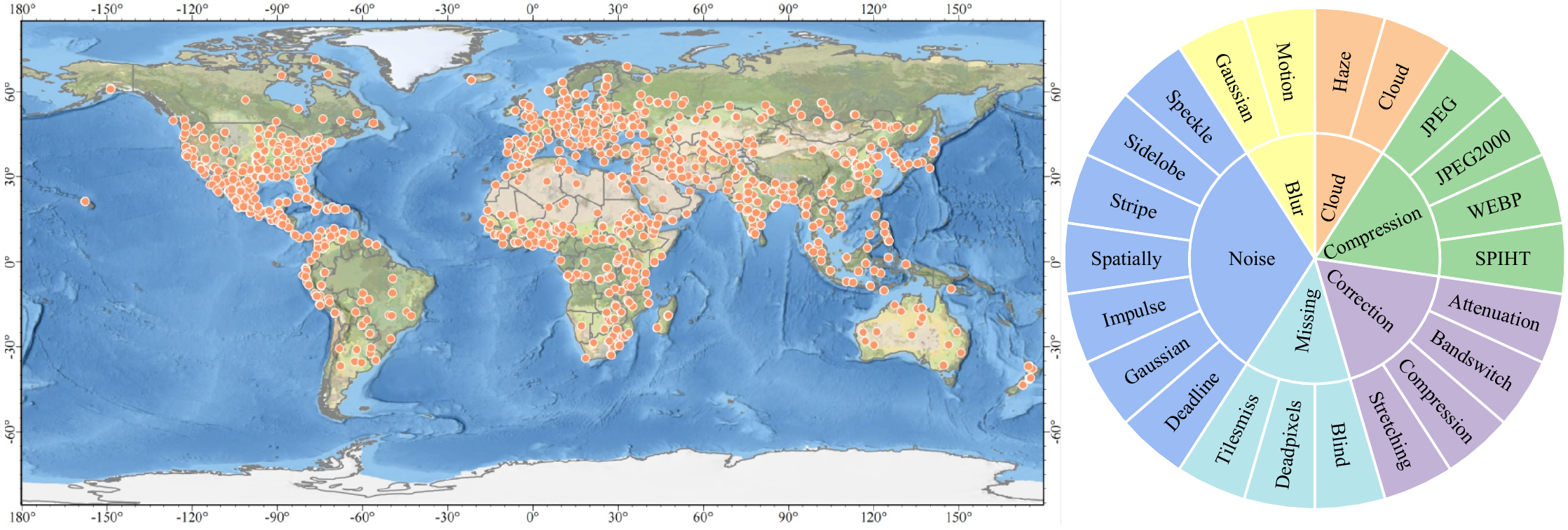}
    }
    \caption{
    Overview of the dataset coverage and degradation taxonomy. 
    The left panel shows the global distribution of sampled locations in our dataset, with orange dots denoting individual sampling points. 
    The right panel presents the hierarchical taxonomy of image degradations used in our benchmark.
    }
    \label{fig:worldwide_taxonomy}
    \vspace{-4mm}
\end{figure}

\subsection{Hierarchical Distortion Taxonomy}
Anchored in the physics of the remote sensing imaging chain and established technical standards (detailed in Appendix~\ref{benchmark_taxonomy}), \textbf{SenseBench} introduces a hierarchical distortion taxonomy spanning the complete acquisition-to-processing pipeline. Integrating classical low-level vision formulations~\cite{superresolution, denoising, deblur, coompression, inpainting}, the taxonomy defines \textbf{6 major categories (L-1)} systematically stratified into \textbf{22 fine-grained sub-categories (L-2)}, as illustrated in Figure~\ref{fig:worldwide_taxonomy}. This granular architecture ensures rigorous coverage of diverse real-world degradation scenarios. Formal definitions are provided in \textbf{Appendix~\ref{Appendix_A_1}}.

\begin{figure}[t]
    \centering
    \includegraphics[width=1\linewidth]{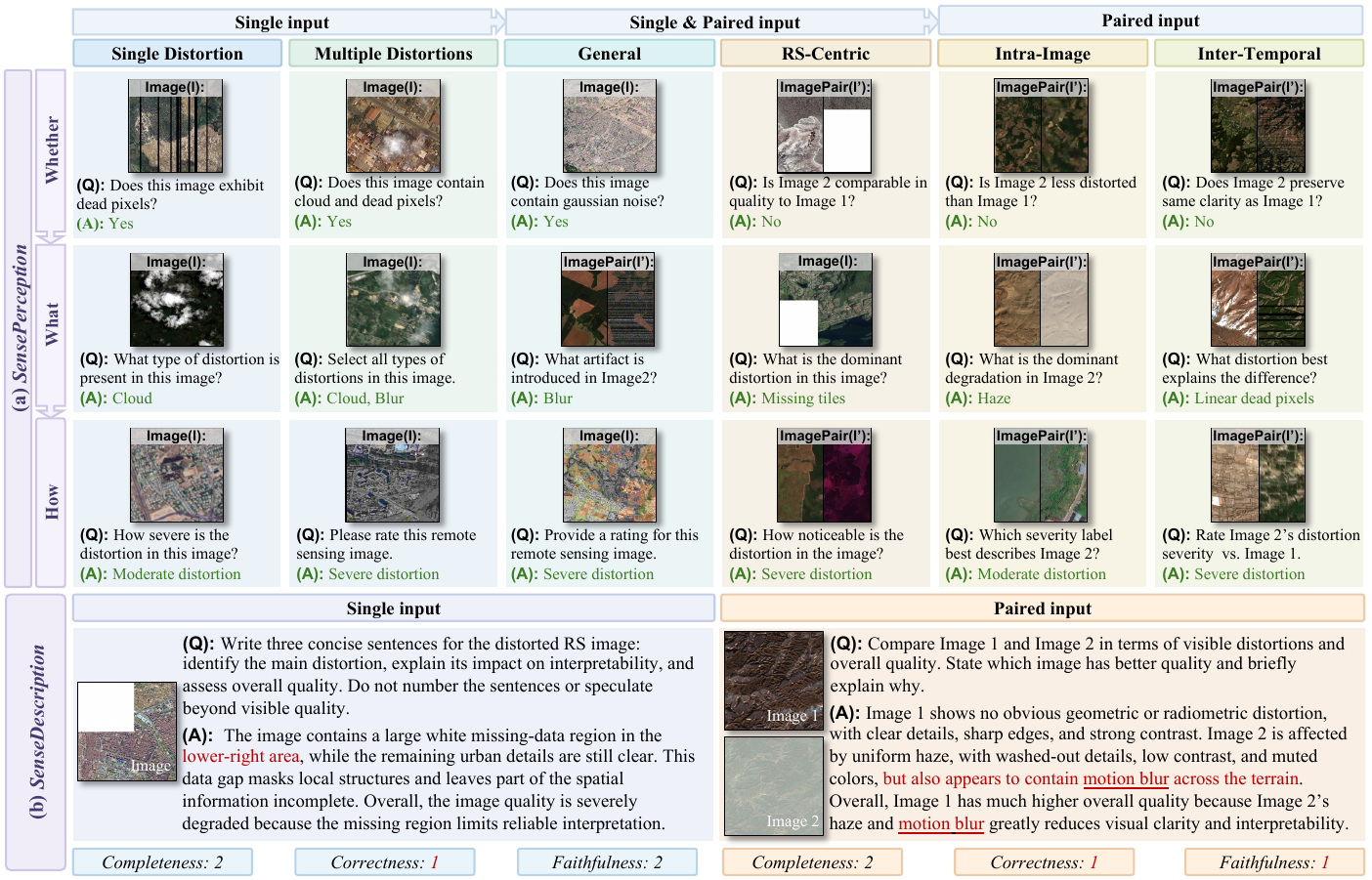}
    \caption{
    Overview of the SenseBench evaluation framework. 
    The upper part shows the \textit{SensePerception} task taxonomy across input formats, distortion settings, and \textit{whether}/\textit{what}/\textit{how} question types. 
    The lower part illustrates \textit{SenseDescription} examples for single and paired inputs, where responses are evaluated by \textit{completeness}, \textit{correctness}, and \textit{faithfulness}, with \textcolor{red}{red} text indicating incorrect or unsupported statements.
    }
    \label{fig:perception}
    \vspace{-4mm}
\end{figure}

\subsubsection{SensePerception}
\label{sec: SensePerception}
Remote sensing imagery exhibits physics-driven artifacts fundamentally distinct from natural photography, a domain gap persistently unverified by existing low-level benchmarks~\cite{qbench+}. To quantify this capability, \textbf{SensePerception} establishes a rigorous multiple-choice question (MCQ) protocol governed by two orthogonal dimensions: \textit{question types} that codify progressive perceptual skills, and \textit{evaluation scenarios} that stratify contextual complexity. Figure~\ref{fig:perception} illustrates representative evaluation instances. Both structural dimensions are formally detailed below.

\textbf{Question Types.} We postulate that an ideal VLM capable of performing reliable Remote Sensing IQA must exhibit progressive perceptual capabilities, advancing from basic degradation detection to fine-grained categorization and severity quantification. Consequently, we formulate three discrete question types, structured along the vertical axis of Figure~\ref{fig:perception}. \textbf{Whether} questions assess base sensitivity to perceivable distortions and quality deviations. \textbf{What} questions demand precise categorization of the dominant degradation (e.g., distinguishing \textit{Gaussian} from sensor-specific \textit{Speckle Noise}). \textbf{How} questions quantify severity across three standardized levels (\textit{No/Slight}, \textit{Moderate}, \textit{Severe}). Extending beyond strict no-reference evaluation, these question types seamlessly adapt to full-reference and pairwise comparative protocols by accepting paired inputs, forcing models to reason over relative artifact existence, typological differences, or severity gaps.

\textbf{Evaluation Scenarios.} To systematically expose domain biases and robustness against escalating \textit{contextual complexity}, samples are stratified across three distinct axes, structured along the horizontal axis of Figure~\ref{fig:perception}. \textbf{General vs.\ RS-Centric} (Domain Context): \textit{General} distortions (e.g., blur, compression) benchmark the transferability of natural-image priors, while \textit{RS-Centric} artifacts (e.g., cloud obstruction, scan-line missing) probe specialized domain adaptation. \textbf{Single vs.\ Multiple} (Degradation Context): \textit{Single} distortions provide isolated diagnostics, whereas \textit{Multiple} superimposed distortions simulate the highly complex, compound degradations of authentic RS workflows. \textbf{Intra-image vs.\ Inter-temporal} (Comparative Context, paired inputs only): \textit{Intra-image} pairings juxtapose pristine and degraded views of identical scenes, echoing full-reference IQA mechanics; \textit{Inter-temporal} pairings evaluate identical geolocations across timestamps, modeling the critical observation selection process in temporal Earth analysis.

\subsubsection{SenseDescription}
\label{sec:three_dimensions}
While multiple-choice paradigms quantify baseline perception, they cannot ascertain if a model synthesizes coherent visual narratives or merely exploits surface-level correlations. Practical RS workflows inherently demand free-form diagnostics, such as rigorous quality inspection notes or comparative justifications for multi-temporal analysis. Operating across the identical evaluation scenarios defined for \textit{SensePerception}, \textit{SenseDescription} challenges models to generate comprehensive natural-language reports for both full-reference and non-reference. Surpassing generic VQA and captioning, these diagnostics must jointly identify degradation typologies, quantify severity, and articulate impacts on visual interpretability.

\textbf{Evaluation Dimensions.} A trustworthy diagnostic report demands exhaustive coverage, factual precision, and reference-grounded reliability. Accordingly, we evaluate candidate generations along three orthogonal dimensions. \textbf{Completeness} measures the recall of all reference degradation factors, penalizing artifact omissions. \textbf{Correctness} assesses strict factual alignment with the reference, penalizing contradictions in visual attribution. \textbf{Faithfulness} measures reference-grounded reliability, penalizing hallucinated, unsupported, or irrelevant claims absent from the verified reference.
\begin{figure}[t]
    \centering
    \includegraphics[width=\textwidth]{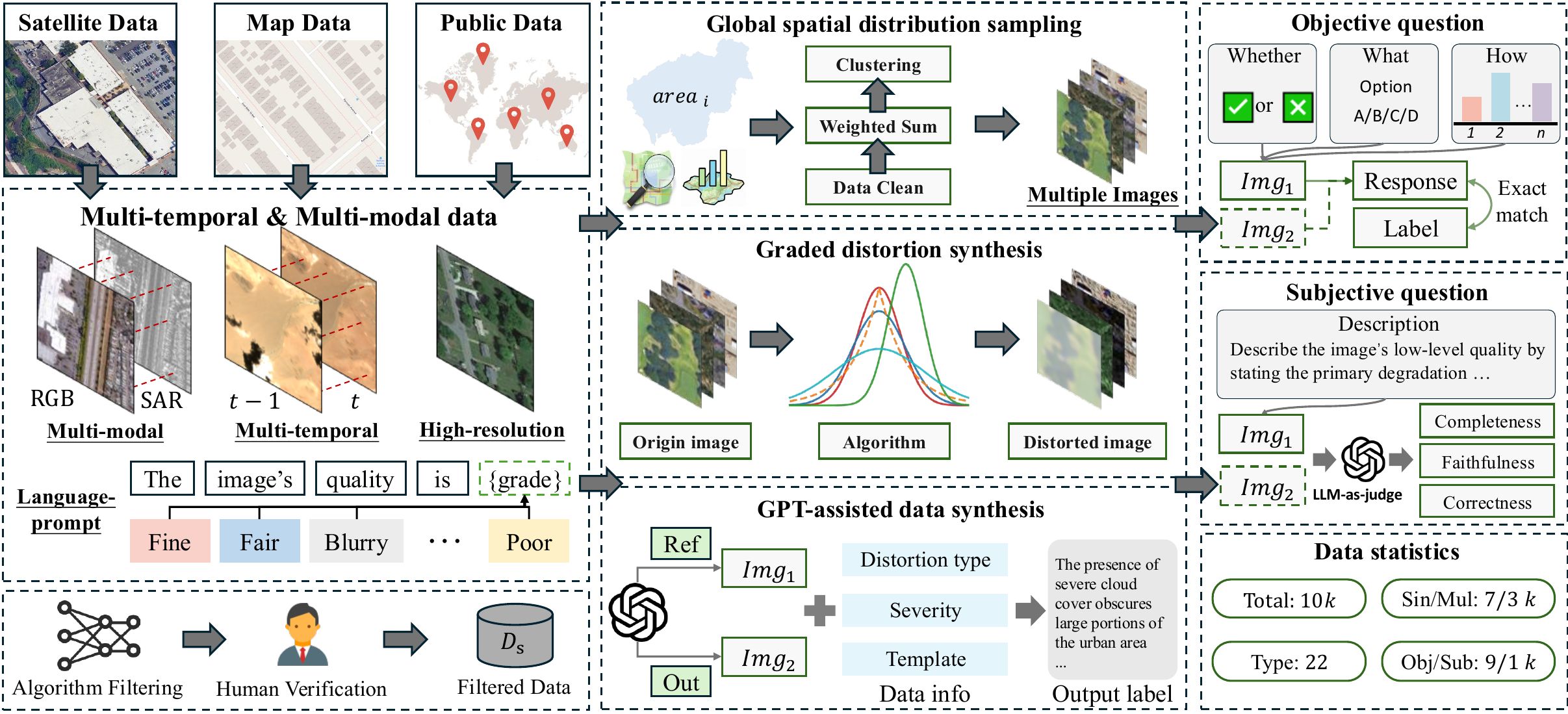}
    \caption{Overview of SenseBench construction. Remote sensing images are sourced globally from various satellite, map, and public datasets. Multiple approaches, including graded distortion synthesis and GPT-assisted data synthesis with language prompts, generate problems, with algorithm filtering and human verification ensuring accuracy of degradations, labels, and data quality.}
    \label{fig:main_architecture}
    \vspace{-5mm}
\end{figure}

\subsection{Benchmark Construction}
\label{benchmark_construction}
Configured strictly as a zero-shot evaluation protocol, \textbf{SenseBench} encompasses 10{,}815 meticulously curated instances, comprising 7{,}300 isolated single-view assessments and 3{,}515 paired comparative ones. This extensive corpus fully covers the 6 major and 22 fine-grained taxonomy categories. As illustrated in Figure~\ref{fig:main_architecture}, the construction pipeline enforces absolute \textit{traceability} and \textit{coverage}: every evaluation instance is directly anchored to parameter-controlled synthesis or a meticulously sourced real-world observation. Crucially, \textit{SenseDescription} reference reports are programmatically generated from the explicit metadata orchestrating the \textit{SensePerception} MCQ labels, thereby structurally eliminating human subjective drift between the two tasks.

\textbf{Global Geospatial Curation.} To mitigate the regional bias prevalent in existing RS datasets, we adopt a rigorous three-stage global sampling strategy. First, continent boundaries from \textit{Natural Earth} are used to enforce continent-wise quotas across all six inhabited continents. Second, representative cities are drawn by population-weighted sampling. Third, geographic coordinates are randomly sampled within \textit{OpenStreetMap} city boundaries, and the corresponding high-resolution tiles are retrieved from \textit{Google Earth}~\cite{gee}. For distortion categories defying synthetic replication (e.g., SAR speckle, cloud obscuration), images are sourced from curated real-world datasets to preserve authentic physical signatures. To comply with provider Terms of Service, all source imagery is curated exclusively for non-commercial academic research.

\textbf{Physics-Based Synthesis \& Traceable Labeling.} Each \textit{SensePerception} sample is generated by a specialized physics-based engine governed by a continuous severity parameter $s$, yielding a functionally degraded image paired with structured, traceable metadata (distortion category, sub-category, severity level). To reduce human annotation subjectivity, task labels are mathematically projected from $s$: \textit{Whether} labels are assigned positive iff $s$ surpasses an empirically calibrated perceptual visibility threshold; \textit{What} labels map directly to the injected distortion category; and \textit{How} labels discretize $s$ into three qualitative severity tiers via distortion-specific thresholds, grounding the entire evaluation protocol in objective and reproducible labeling protocol.

\textbf{Constraint-Driven Reference Generation.} \textit{SenseDescription} references are synthesized by prompting GPT-5.2 with a triplet $\langle\text{distorted image, high-quality reference, structured metadata}\rangle$, designed as detailed visual inspection reports rather than generic captions. Two strict principles constrain generation: (i)~\textit{Perceptual Visibility}, limiting narratives to visible phenomena to eradicate semantic hallucinations; and (ii)~\textit{Comprehensive Coverage}, forcing exhaustive alignment with the taxonomy in Figure~\ref{fig:worldwide_taxonomy}. To structurally prevent inference-time leakage, high-quality references are strictly withheld post-generation, yielding completely self-contained target texts. These generated candidates then undergo rigorous \textit{human-in-the-loop verification} to systematically rectify unsupported claims and remove latent leakage residues, ultimately outputting definitive semantic ground truths.

\textbf{Multi-Stage Quality Auditing.} Auditing is conducted by 12 evaluators with backgrounds in RS and computer vision via a three-phase protocol: (i)~algorithmic filtering removes retrieval errors and corrupted tiles; (ii)~human verification audits consistency among metadata, labels, and task definitions; and (iii)~perceptual auditing on a stratified 8\% sample evaluates degradation recognizability, severity alignment, and synthesis naturalness, with each sample cross-checked by two evaluators and disagreements adjudicated by a third expert. Samples failing any stage are discarded and re-synthesized. Full statistics, prompt templates, annotation guidelines, and rubrics are provided in Appendix~\ref{Appendix_A_2} and Appendix~\ref{sec:quality_control}.

\subsection{Evaluation Strategy}
\textbf{Evaluation of \textit{SensePerception}.} Both single-image and paired-image \textit{SensePerception} tasks enforce a strictly zero-shot, multiple-choice formulation containing 2 to 4 candidates per query. VLMs are instructed to emit either the canonical choice letter or the verbatim text, enabling a deterministic extraction pipeline to reliably recover the predicted index. To rigorously neutralize positional priors, the ground-truth locations are uniformly randomized, whereas distractors adhere to a fixed semantic ordering. We report absolute exact-match (EM) accuracy as the primary quantitative metric.


\textbf{Evaluation of \textit{SenseDescription}.}
For open-ended diagnostic reporting, static n-gram metrics such as BLEU~\cite{bleu} and ROUGE~\cite{rouge} fail to capture fine-grained semantic consistency and low-level diagnostic accuracy. We therefore adopt an \textit{LLM-as-a-Judge} framework~\cite{gu2024survey}. To reduce self-preference bias, we use \textit{Prometheus}~\cite{kim2024prometheus}, an open-source evaluation-oriented LLM, as the independent judge, which is architecturally distinct from the model used for reference generation. Given a candidate report $\hat{R}_i$ and the gold-standard reference $R_i$, the judge assigns a discrete score $s_i^d \in \{0,1,2\}$ for each evaluation dimension $d \in \{\textit{Completeness}, \textit{Correctness}, \textit{Faithfulness}\}$ defined in Sec.~\ref{sec:three_dimensions}, where 0, 1, and 2 denote critical failure, partial alignment, and full alignment, respectively. For each dimension, we report both the mean score and the per-score proportions $\{P_0, P_1, P_2\}$. The composite \textit{Sum} score is computed as the unweighted sum of the three mean dimension scores.

\section{Experiment}

\begin{table*}[!t]
    \centering
    \small 
    \renewcommand\arraystretch{1} 
    \renewcommand\tabcolsep{4.5pt}
    
    \caption{Quantitative results of VLMs on low-level perception for single images.}
    
    \label{perceptionsingle}
    \newcolumntype{C}{>{\centering\arraybackslash}p{1.3cm}}
    \resizebox{0.95\textwidth}{!}{%
    \begin{tabular}{l | CCC | CC | CC | C}
    \toprule
    
    \textbf{Sub-categories} & \multicolumn{3}{c|}{\textbf{Question Type}} & \multicolumn{2}{c|}{\textbf{Domain}} & \multicolumn{2}{c|}{\textbf{Context}} & \multirow{2}{*}{\textit{Average$\uparrow$}} \\ 
    
    \cdashline{1-8} 
    
    \rule{0pt}{12pt}\textbf{Model} & \textit{Whether$\uparrow$} & \textit{What$\uparrow$} & \textit{How$\uparrow$} & \textit{General$\uparrow$} & \textit{RS$\uparrow$} & \textit{Single$\uparrow$} & \textit{Multi$\uparrow$} & \\ 
    
    \midrule
    \textcolor{gray}{\textsc{Human Evaluator}} & \textcolor{gray}{90.58\%} & \textcolor{gray}{88.62\%} & \textcolor{gray}{90.26\%} & \textcolor{gray}{90.11\%} & \textcolor{gray}{89.32\%} & \textcolor{gray}{89.75\%} & \textcolor{gray}{90.00\%} & \textcolor{gray}{89.82\%}\\
    \noalign{\vskip 1pt}
    \cdashline{1-9}
    \noalign{\vskip 1pt}

    GPT-5.4 & 68.46\% & 60.43\% & 58.54\% & 64.83\% & \underline{62.87\%} & 64.23\% & 41.50\% & 62.48\% \\
    Gemini-3.1-pro-preview & \textbf{87.17\%} & \textbf{80.17\%} & \underline{59.92\%} & \textbf{79.47\%} & \textbf{80.06\%} & \textbf{79.68\%} & 41.17\% & \textbf{75.75\%} \\

    \midrule
    Ovis2.5-9B\cite{ovis2.5} & 50.39\% & 41.48\% & 37.31\% & 46.33\% & 41.48\% & 43.87\% & 29.50\% & 43.06\% \\
    Llama3.2-11B\cite{llama3.2} & 54.17\% & 42.21\% & 38.62\% & 49.70\% & 42.02\% & 45.57\% & 46.67\% & 45.00\% \\
    Phi4-vision\cite{phi4} & 51.75\% & 37.33\% & 38.38\% & 43.97\% & 42.19\% & 42.42\% & 44.83\% & 42.49\% \\
    LLaVA2-ov-7B\cite{llava-ov} & \underline{70.58\%} & \underline{67.57\%} & 54.31\% & \underline{71.07\%} & 61.46\% & \underline{67.26\%} & 46.00\% & 64.15\% \\
    InternVL3.5-8B\cite{internvl3.5} & 56.59\% & 40.70\% & 42.38\% & 48.89\% & 44.52\% & 47.19\% & 46.83\% & 46.56\% \\
    InternVL3.5-14B\cite{internvl3.5} & 53.14\% & 41.77\% & 42.23\% & 46.47\% & 44.64\% & 46.47\% & 39.67\% & 45.71\% \\
    InternVL3.5-38B\cite{internvl3.5} & 57.70\% & 50.27\% & 25.46\% & 49.92\% & 50.52\% & 46.43\% & 36.67\% & 44.48\% \\
    InternVL3.5-30B-A3B\cite{internvl3.5} & 58.99\% & 47.96\% & 47.77\% & 53.30\% & 51.61\% & 52.26\% & 44.17\% & 51.57\% \\
    Qwen3VL-8B\cite{qwen3vl} & 64.35\% & 54.20\% & 52.38\% & 59.80\% & 53.29\% & 58.02\% & 50.33\% & 56.98\% \\
    Qwen3VL-32B\cite{qwen3vl} & 69.94\% & 63.65\% & \textbf{60.15\%} & 70.06\% & 56.40\% & 65.92\% & \underline{58.33\%} & \underline{64.58\%} \\
    Qwen3VL-30B-A3B\cite{qwen3vl} & 65.66\% & 54.75\% & 50.46\% & 60.93\% & 54.83\% & 57.13\% & \textbf{58.50\%} & 56.96\% \\
    KimiVL-A3B\cite{kimivl} & 66.11\% & 51.73\% & 49.31\% & 58.82\% & 55.89\% & 56.85\% & 47.17\% & 55.72\% \\
    
    \midrule 
    DepictQA\cite{depictqa} & 50.06\% & 31.49\% & 21.00\% & 39.00\% & 33.20\% & 35.87\% & 41.00\% & 34.18\% \\
    Q-instruct-llava-13b\cite{qinstruct} & 53.95\% & 36.41\% & 48.69\% & 45.37\% & 46.98\% & 46.11\% & 40.83\% & 46.35\% \\
    Q-instruct-llava-7b\cite{qinstruct} & 47.17\% & 28.11\% & 34.31\% & 37.40\% & 36.31\% & 36.57\% & 37.50\% & 36.53\% \\
    
    \midrule 
    GeoChat\cite{geochat} & 46.13\% & 29.80\% & 20.92\% & 34.60\% & 33.83\% & 33.57\% & 33.50\% & 32.28\% \\
    TEOChat\cite{teochat} & 54.57\% & 29.99\% & 27.38\% & 37.70\% & 41.52\% & 38.51\% & 24.50\% & 37.31\% \\
    EarthDial\cite{soni2025earthdial} & 58.96\% & 48.06\% & 30.31\% & 46.23\% & 52.87\% & 48.89\% & 20.50\% & 45.78\% \\
    LHRS-Bot\cite{lhrs} & 54.96\% & 27.01\% & 26.15\% & 35.47\% & 40.56\% & 37.40\% & 33.17\% & 36.04\% \\
    LHRS-Bot-nova\cite{li2025lhrsbotnava} & 47.83\% & 23.54\% & 21.54\% & 30.83\% & 33.85\% & 30.85\% & 41.33\% & 30.97\% \\
    \bottomrule
    \end{tabular}
    }
    \vspace{-4mm}
\end{table*}

\label{04_experiment}

We conducted a comprehensive evaluation of 29 mainstream VLMs on SenseBench. The main paper reports 22 representative models, covering 14 general-domain VLMs, 5 RS vision-language models, and 3 IQA vision-language models. The general-domain VLMs reported in the main tables include two closed-source models, GPT-5.4 and Gemini-3.1-Pro-Preview, as well as Ovis2.5-9B~\cite{ovis2.5}, Llama3.2-11B~\cite{llama3.2}, Phi4-vision~\cite{phi4}, LLaVA2-ov-7B~\cite{llava-ov}, InternVL3.5-8B/14B/38B/30B-A3B~\cite{internvl3.5}, Qwen3VL-8B/32B/30B-A3B~\cite{qwen3vl}, and KimiVL-A3B~\cite{kimivl}. The RSVLMs include GeoChat~\cite{geochat}, TEOChat~\cite{teochat}, EarthDial~\cite{soni2025earthdial}, and LHRS-Bot/LHRS-Bot-nova~\cite{lhrs, li2025lhrsbotnava}. The IQA-VLMs include DepictQA~\cite{depictqa} and Q-Instruct-LLaVA-7B/13B~\cite{qinstruct}. Further details on the evaluation protocol, implementation settings, and metric computation are provided in Appendix~\ref{evaluation_details}, while complete results are reported in Appendix~\ref{sec:additional_results}.

\subsection{Findings on SensePerception}
\label{sec: findings_senseperception}
The quantitative results for the \textit{SensePerception} task are detailed in Tables~\ref{perceptionsingle} and~\ref{perceptionmulti}. By dissecting performance across our hierarchical taxonomy, SenseBench unveils several critical characteristics of current multimodal architectures.

\begin{figure}[!ht] 
    \centering
    \includegraphics[width=1\textwidth]{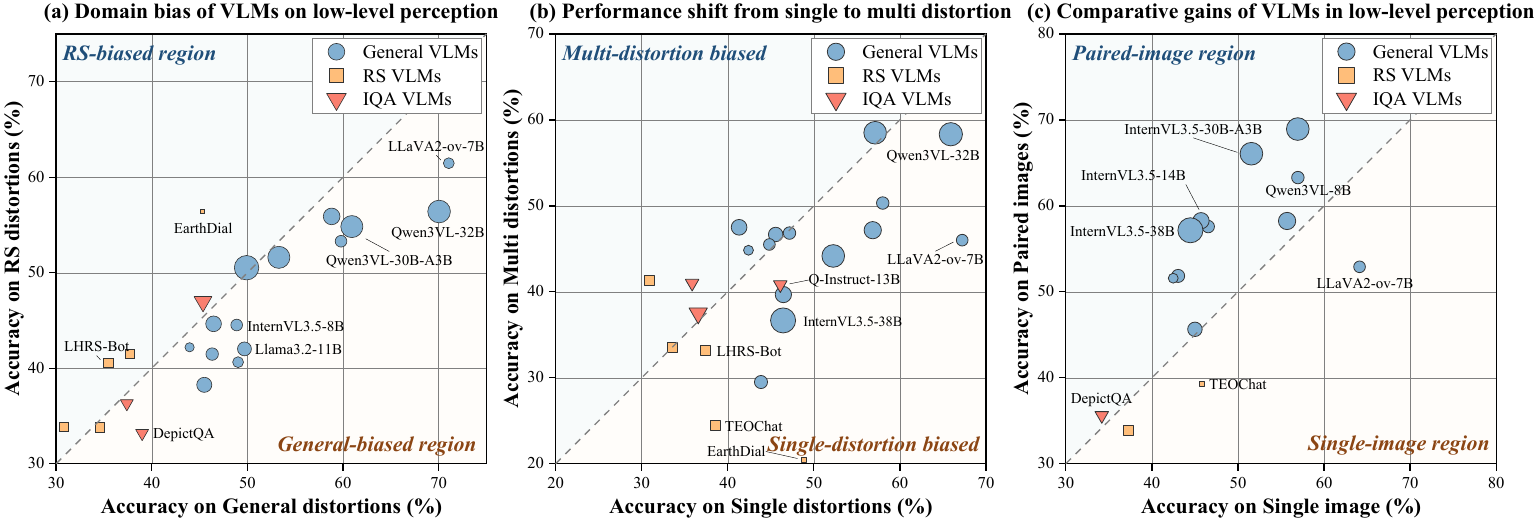}
    \caption{Cross-perspective analysis of VLMs on SenseBench. Circles, squares, and triangles denote general-domain, RS-specialized, and IQA-VLMs, respectively; marker size indicates parameter count, and the dashed diagonal marks parity.}
    \label{fig:Experiment}
    \vspace{-5mm}
\end{figure}

\textbf{The Domain Gap in Foundational Priors.}
Figure~\ref{fig:Experiment} (a) provides direct empirical evidence of a severe, bipartite domain bias in low-level perceptual priors. General-domain VLMs systematically cluster below the diagonal, demonstrating robust perception of general degradations while failing to generalize to physics-driven RS artifacts. Conversely, RS-specialized models exhibit an inverse bias, shifting above the diagonal due to domain-specific instruction tuning that seemingly compromises generic robustness. Crucially, as shown in Table~\ref{perceptionsingle}, this dichotomy is orthogonal to raw model capacity: even state-of-the-art architectures exhibit profound domain penalties, with Qwen3VL-32B degrading from $70.06\%$ on general to $56.40\%$ on RS-centric distortions. The absence of any model in the upper-right quadrant (the area of balanced, high-accuracy perception) clearly demonstrates that general visual capabilities do not automatically transfer to specialized low-level tasks. To achieve true visual universality, future models must explicitly incorporate physical degradation distributions during training.

\begin{table*}[!t]
    \centering
    \small 
    \renewcommand\arraystretch{1} 
    \renewcommand\tabcolsep{4.5pt}
    
    \caption{Quantitative results of VLMs on low-level perception for paired images.}
    \label{perceptionmulti}
    \newcolumntype{C}{>{\centering\arraybackslash}p{1.3cm}}
    
    \resizebox{\textwidth}{!}{%
    \begin{tabular}{l | CCC | CC | CC | C}
    \toprule
    
    \textbf{Sub-categories} & \multicolumn{3}{c|}{\textbf{Question Type}} & \multicolumn{2}{c|}{\textbf{Domain}} & \multicolumn{2}{c|}{\textbf{Context}} & \multirow{2}{*}{\textit{Average$\uparrow$}} \\ 
    
    \cdashline{1-8} 

    \rule{0pt}{12pt}\textbf{Model} & \textit{Whether$\uparrow$} & \textit{What$\uparrow$} & \textit{How$\uparrow$} & \textit{General$\uparrow$} & \textit{RS$\uparrow$} & \textit{Intra.$\uparrow$} & \textit{Inter.$\uparrow$} & \\ 

    \midrule
    \textcolor{gray}{\textsc{Human Evaluator}} & \textcolor{gray}{95.28\%} & \textcolor{gray}{89.44\%} & \textcolor{gray}{82.05\%} & \textcolor{gray}{89.81\%} & \textcolor{gray}{89.87\%} & \textcolor{gray}{81.92\%} & \textcolor{gray}{89.62\%} & \textcolor{gray}{88.92\%} \\
    \noalign{\vskip 1pt}
    \cdashline{1-9}
    \noalign{\vskip 1pt}

    GPT-5.4 & 85.64\% & \underline{80.45\%} & 62.36\% & \underline{76.73\%} & 75.58\% & \underline{77.50\%} & 78.00\% & \underline{76.15\%} \\
    Gemini-3.1-pro-preview & \underline{89.64\%} & \textbf{90.00\%} & \textbf{64.21\%} & \textbf{81.94\%} & \textbf{80.63\%} & \textbf{82.55\%} & \textbf{84.33\%} & \textbf{81.28\%} \\

    \midrule
    Ovis2.5-9B\cite{ovis2.5} & 58.36\% & 60.55\% & 36.64\% & 51.52\% & 56.92\% & 51.13\% & 61.13\% & 51.85\% \\
    Llama3.2-11B\cite{llama3.2} & 72.73\% & 32.82\% & 31.33\% & 46.00\% & 49.49\% & 45.42\% & 53.08\% & 45.63\% \\
    Phi4-vision\cite{phi4} & 83.73\% & 34.64\% & 36.36\% & 52.24\% & 55.34\% & 52.64\% & 56.10\% & 51.58\% \\
    LLaVA2-ov-7B\cite{llava-ov} & 69.00\% & 52.09\% & 37.62\% & 57.21\% & 51.94\% & 51.70\% & 63.52\% & 52.90\% \\
    InternVL3.5-8B\cite{internvl3.5} & 51.27\% & 57.55\% & \underline{63.92\%} & 52.61\% & 62.13\% & 56.37\% & 57.74\% & 57.58\% \\
    InternVL3.5-14B\cite{internvl3.5} & 56.91\% & 61.27\% & 56.64\% & 54.85\% & 63.24\% & 57.92\% & 60.00\% & 58.27\% \\
    InternVL3.5-38B\cite{internvl3.5} & 56.00\% & 62.73\% & 52.73\% & 51.94\% & 65.30\% & 59.62\% & 52.70\% & 57.15\% \\
    InternVL3.5-30B-A3B\cite{internvl3.5} & 68.73\% & 66.64\% & 62.80\% & 63.58\% & 70.28\% & 65.52\% & 69.06\% & 66.06\% \\
    Qwen3VL-8B\cite{qwen3vl} & 72.00\% & 66.55\% & 51.33\% & 63.64\% & 66.48\% & 63.25\% & 69.18\% & 63.29\% \\
    Qwen3VL-32B\cite{qwen3vl} & 79.27\% & 75.82\% & 57.62\% & 70.30\% & \underline{75.73\%} & 70.57\% & 78.24\% & 70.90\% \\
    Qwen3VL-30B-A3B\cite{qwen3vl} & \textbf{91.55\%} & 60.55\% & 54.69\% & 69.82\% & 72.09\% & 66.60\% & \underline{82.01\%} & 68.93\% \\
    KimiVL-A3B\cite{kimivl} & 71.45\% & 53.73\% & 49.51\% & 58.18\% & 60.95\% & 54.95\% & 71.19\% & 58.23\% \\

    \midrule 
    DepictQA\cite{depictqa} & 60.18\% & 23.09\% & 23.55\% & 41.82\% & 29.39\% & 37.03\% & 42.44\% & 35.61\% \\
    \midrule
    TEOChat\cite{teochat} & 34.55\% & 30.00\% & 37.06\% & 31.64\% & 35.81\% & 35.33\% & 28.43\% & 33.87\% \\
    EarthDial\cite{soni2025earthdial} & 26.91\% & 50.73\% & 40.28\% & 38.06\% & 40.63\% & 43.92\% & 26.54\% & 39.31\% \\

    \bottomrule
    \end{tabular}
    
    \vspace{-5mm}
    }
\end{table*}

\textbf{Robustness Collapse Under Multi-distortion.}
Figure~\ref{fig:Experiment} (b) evaluates model resilience against compounded degradations. The resulting distribution overwhelmingly skews below the diagonal, confirming that severe vulnerability to superimposed artifacts is a population-level bottleneck across modern VLMs, rather than an isolated architectural flaw. Alarmingly, as shown in Table~\ref{perceptionsingle}, the steepest performance collapses occur among the most capable single-distortion performers: Gemini-3.1-pro-preview plummets from $79.68\%$ to $41.17\%$, and EarthDial collapses from $48.89\%$ to $20.50\%$. While a few low-performing models hover near the diagonal, their apparent robustness is merely a statistical floor effect, where near-random baseline inaccuracy masks the true compounding difficulty. These findings empirically demonstrate that multi-distortion robustness is a distinct capability rather than a simple extension of single-distortion competence. Models that successfully recognize isolated atomic artifacts fundamentally fail to disentangle them when superposed. This exposes a critical limitation in existing evaluation paradigms: standard single-distortion benchmarks provide a dangerously over-optimistic proxy for real-world reliability, making SenseBench's stratified multi-distortion testing mandatory.

\textbf{The Anchoring Effect in Paired Perception.}
Figure~\ref{fig:Experiment} (c) contrasts non-reference and full-reference perception to evaluate the impact of visual priors. While the model population generally shifts above the diagonal—confirming that a reference acts as a structural anchor to lower degradation detection thresholds—this anchoring benefit uncovers a stark domain dichotomy. General-domain VLMs systematically leverage the reference to boost accuracy, whereas RSVLMs paradoxically regress, exposing a critical deficit in multi-image spatial alignment within current domain-specific tuning recipes. Notably, as shown in Table~\ref{perceptionsingle} and~\ref{perceptionmulti}, strong single-image performers such as InternVL3.5-38B further benefit from comparative settings, improving by over 10\% when provided with paired visual evidence. This reveals a fundamental disconnect: single-image perception is not a reliable proxy for comparative reasoning. Therefore, evaluating both scenarios independently is essential, validating SenseBench's dual-protocol design.

\begin{table*}[!t]
    \centering
    \small
    \renewcommand\arraystretch{1}
    \setlength{\tabcolsep}{2.6pt}
    
    \caption{Quantitative results of VLMs on low-level description for single and paired images.}
    \label{tab:description_all}
    
    \newcolumntype{C}{>{\centering\arraybackslash}p{1.3cm}}
    \resizebox{0.93\textwidth}{!}{%
    \begin{tabular}{l | CCCC | CCCC}
    \toprule
    
    \textbf{Evaluation Settings} 
        & \multicolumn{4}{c|}{\textbf{Single Image}} 
        & \multicolumn{4}{c}{\textbf{Paired Images}} \\
    
    \cdashline{1-9}
    
    \rule{0pt}{12pt}\textbf{Model} 
        & \textit{Complete.$\uparrow$}
        & \textit{Correct.$\uparrow$}
        & \textit{Faithful.$\uparrow$}
        & \textit{Sum}$\uparrow$
        & \textit{Complete.$\uparrow$}
        & \textit{Correct.$\uparrow$}
        & \textit{Faithful.$\uparrow$}
        & \textit{Sum}$\uparrow$ \\
    \midrule
    
    \textcolor{gray}{\textsc{Human Evaluator}}
    & \textcolor{gray}{1.70} & \textcolor{gray}{1.57} & \textcolor{gray}{1.92} & \textcolor{gray}{5.19}
    & \textcolor{gray}{1.70} & \textcolor{gray}{1.45} & \textcolor{gray}{1.64} & \textcolor{gray}{4.79} \\
    \noalign{\vskip 1pt}
    \cdashline{1-9}
    \noalign{\vskip 1pt}

    GPT-5.4
        & \textbf{1.84} & \textbf{1.47} & 1.56 & \textbf{4.87}
        & \textbf{1.75} & 1.13 & 1.22 & 4.10 \\
    Gemini-3.1-pro-preview
        & \underline{1.73} & \underline{1.45} & 1.53 & 4.71
        & \underline{1.70} & \textbf{1.49} & 1.50 & \textbf{4.69} \\

    \midrule
    Ovis2.5-9B\cite{ovis2.5} 
        & 1.38 & 1.44 & \textbf{1.93} & \underline{4.75} 
        & 1.10 & \underline{1.35} & \textbf{1.76} & \underline{4.21} \\
    Llama-3.2-11B\cite{llama3.2} 
        & 0.79 & 0.63 & 1.11 & 2.53 
        & 0.97 & 0.67 & 0.94 & 2.58 \\
    Phi4-vision\cite{phi4} 
        & 1.05 & 0.75 & 1.73 & 3.53 
        & 0.83 & 0.09 & 0.62 & 1.54 \\
    LLaVA2-ov-7B\cite{llava-ov}
        & 0.92 & 1.19 & 1.76 & 3.87 
        & 1.01 & 0.46 & 0.95 & 2.42 \\
    InternVL3.5-8B\cite{internvl3.5} 
        & 1.19 & 0.85 & 1.74 & 3.78 
        & 0.80 & 0.37 & 0.75 & 1.92 \\
    InternVL3.5-14B\cite{internvl3.5} 
        & 1.20 & 0.81 & 1.88 & 3.89 
        & 0.70 & 0.35 & 0.74 & 1.79 \\
    InternVL3.5-38B\cite{internvl3.5} 
        & 1.01 & 0.71 & 1.73 & 3.45 
        & 0.84 & 0.64 & 0.89 & 2.37 \\
    InternVL3.5-30B-A3B\cite{internvl3.5} 
        & 0.97 & 0.66 & 1.60 & 3.23 
        & 0.82 & 0.42 & 0.86 & 2.10 \\
    Qwen3VL-8B\cite{qwen3vl} 
        & 1.24 & 0.99 & 1.85 & 4.08 
        & 1.23 & 0.66 & 1.24 & 3.13 \\
    Qwen3VL-32B\cite{qwen3vl} 
        & 1.32 & 1.06 & \underline{1.91} & 4.29 
        & 1.46 & 1.11 & \underline{1.61} & 4.18 \\
    Qwen3VL-30B-A3B\cite{qwen3vl} 
        & 1.29 & 1.09 & 1.82 & 4.20 
        & 1.15 & 0.81 & 1.22 & 3.18 \\
    KimiVL-A3B\cite{kimivl} 
        & 0.76 & 0.78 & 1.08 & 2.62 
        & 0.97 & 0.84 & 1.09 & 2.90 \\

    \midrule
    DepictQA\cite{depictqa} 
        & 1.50 & 0.97 & 1.19 & 3.66 
        & 1.11 & 0.31 & 1.05 & 2.47 \\
    Q-Instruct-LLaVA-13B\cite{qinstruct} 
        & 1.04 & 1.07 & 1.21 & 3.32
        & - & - & - & - \\
    Q-Instruct-LLaVA-7B\cite{qinstruct} 
        & 1.06 & 1.07 & 1.18 & 3.31 
        & - & - & - & - \\

    \midrule
    GeoChat\cite{geochat} 
        & 0.83 & 0.32 & 0.90 & 2.05 
        & - & - & - & - \\
    TEOChat\cite{teochat} 
        & 0.84 & 0.39 & 1.38 & 2.61 
        & 0.37 & 0.19 & 0.45 & 1.01 \\
    EarthDial\cite{soni2025earthdial} 
        & 0.22 & 0.35 & 0.52 & 1.09 
        & 0.00 & 0.00 & 0.23 & 0.23 \\

    \bottomrule
    \end{tabular}
    }
    \vspace{-5mm}
\end{table*}

\subsection{Findings on SenseDescription}
While \textit{SensePerception} rigorously quantifies baseline recognition accuracy, practical remote sensing workflows demand actionable, natural-language diagnostics. \textit{SenseDescription} elevates the evaluation from discrete classification to open-ended diagnostic reporting, challenging models to explicitly articulate degradation semantics. Table~\ref{tab:description_all} reports performance across the three axes defined in \ref{benchmark_construction}, unveiling critical nuances in how current VLMs translate visual priors into coherent narratives.

\textbf{The Fluency Illusion in Diagnostic Description.}
Table~\ref{tab:description_all} exposes a systematic asymmetry across evaluation axes: models achieve high Faithfulness and moderate Completeness, but consistently collapse on Correctness. This unveils a \textit{fluency illusion}—current VLMs generate diagnostic reports that appear linguistically confident and visually grounded, yet fundamentally misidentify the specific degradation type or misjudge its severity. The dominant failure mode is therefore not semantic hallucination, but imprecise low-level categorization (e.g., confusing SAR speckle with Gaussian noise) and inaccurate severity scaling. A monolithic scalar metric would entirely obscure these disparate bottlenecks, substantiating the necessity of SenseDescription's multi-dimensional rubric for dissecting genuine diagnostic capabilities.

\textbf{The Perception-Description Inversion Effect in Full-reference Evaluation.}
Table~\ref{tab:description_all} reveals a striking contradiction between discrete perceptual accuracy and open-ended generative competence. Models like InternVL3.5 series suffer a precipitous drop in composite \textit{Sum} scores (from $\sim3.6$ to $\sim2.0$), while several RSVLMs hit catastrophic $P_0$ failure rates approaching $100\%$ in Completeness, indicating a near-total inability to articulate cross-image discrepancies. We hypothesize this severe degradation stems from a critical lack of comparative instruction-tuning: while models possess sufficient single-image feature extraction capabilities, they lack the specific cross-attention alignment necessary to track and narrate subtle pixel-level deviations between two simultaneously presented high-resolution inputs. Consequently, perceiving and narrating comparative data face distinct bottlenecks. While paired perception benefits implicitly from structural anchoring, paired description is profoundly constrained by the scarcity of multi-image relational reasoning during instruction tuning. Exposing this critical capacity gap is essential, validating the necessity of SenseBench's dual-task design.
\section{Conclusion}
In this paper, we introduce \textbf{SenseBench}, the first rigorous benchmark dedicated to evaluating the low-level visual perception and diagnostic description capabilities of VLMs in remote sensing. Spanning 10,815 instances across a physics-driven degradation taxonomy, it thoroughly assesses architectures through complementary non-reference and full-reference protocols. Our extensive zero-shot evaluation of 29 state-of-the-art VLMs exposes critical structural bottlenecks: a severe bipartite domain bias that precludes visual universality, a population-level robustness collapse under compounded distortions, and a deceptive "fluency illusion" in textual diagnostics. Crucially, we identify a profound perception-description inversion effect when processing paired inputs, demonstrating that current instruction tuning severely lacks multi-image relational reasoning. Ultimately, these vulnerabilities reveal a systemic gap between high-level semantic alignment and low-level physical fidelity. SenseBench establishes the indispensable analytical framework required to forge genuinely reliable, degradation-aware visual foundations for Earth observation.

\bibliographystyle{unsrtnat}
\bibliography{bibtex/bib/ref}

@inproceedings{mmbench,
  title={Mmbench: Is your multi-modal model an all-around player?},
  author={Liu, Yuan and Duan, Haodong and Zhang, Yuanhan and Li, Bo and Zhang, Songyang and Zhao, Wangbo and Yuan, Yike and Wang, Jiaqi and He, Conghui and Liu, Ziwei and others},
  booktitle={European conference on computer vision},
  pages={216--233},
  year={2024},
  organization={Springer}
}

@article{mmstar,
  title={Are we on the right way for evaluating large vision-language models?},
  author={Chen, Lin and Li, Jinsong and Dong, Xiaoyi and Zhang, Pan and Zang, Yuhang and Chen, Zehui and Duan, Haodong and Wang, Jiaqi and Qiao, Yu and Lin, Dahua and others},
  journal={Advances in Neural Information Processing Systems},
  volume={37},
  pages={27056--27087},
  year={2024}
}

@inproceedings{lhrs,
  title={Lhrs-bot: Empowering remote sensing with vgi-enhanced large multimodal language model},
  author={Muhtar, Dilxat and Li, Zhenshi and Gu, Feng and Zhang, Xueliang and Xiao, Pengfeng},
  booktitle={European Conference on Computer Vision},
  pages={440--457},
  year={2024},
  organization={Springer}
}

@inproceedings{geochat,
  title={Geochat: Grounded large vision-language model for remote sensing},
  author={Kuckreja, Kartik and Danish, Muhammad Sohail and Naseer, Muzammal and Das, Abhijit and Khan, Salman and Khan, Fahad Shahbaz},
  booktitle={Proceedings of the IEEE/CVF Conference on Computer Vision and Pattern Recognition},
  pages={27831--27840},
  year={2024}
}

@article{vrsbench,
  title={Vrsbench: A versatile vision-language benchmark dataset for remote sensing image understanding},
  author={Li, Xiang and Ding, Jian and Elhoseiny, Mohamed},
  journal={Advances in Neural Information Processing Systems},
  volume={37},
  pages={3229--3242},
  year={2024}
}

@article{choice,
  title={CHOICE: Benchmarking the Remote Sensing Capabilities of Large Vision-Language Models},
  author={An, Xiao and Sun, Jiaxing and Gui, Zihan and He, Wei},
  journal={arXiv preprint arXiv:2411.18145},
  year={2024}
}

@article{reobench,
  title={REOBench: Benchmarking Robustness of Earth Observation Foundation Models},
  author={Li, Xiang and Tao, Yong and Zhang, Siyuan and Liu, Siwei and Xiong, Zhitong and Luo, Chunbo and Liu, Lu and Pechenizkiy, Mykola and Zhu, Xiao Xiang and Huang, Tianjin},
  journal={arXiv preprint arXiv:2505.16793},
  year={2025}
}

@article{qbench,
  title={Q-bench: A benchmark for general-purpose foundation models on low-level vision},
  author={Wu, Haoning and Zhang, Zicheng and Zhang, Erli and Chen, Chaofeng and Liao, Liang and Wang, Annan and Li, Chunyi and Sun, Wenxiu and Yan, Qiong and Zhai, Guangtao and others},
  journal={arXiv preprint arXiv:2309.14181},
  year={2023}
}

@article{qbench+,
  title={Q-bench: A benchmark for multi-modal foundation models on low-level vision from single images to pairs},
  author={Zhang, Zicheng and Wu, Haoning and Zhang, Erli and Zhai, Guangtao and Lin, Weisi},
  journal={IEEE Transactions on Pattern Analysis and Machine Intelligence},
  year={2024},
  publisher={IEEE}
}

@article{disasterm3,
  title={DisasterM3: A Remote Sensing Vision-Language Dataset for Disaster Damage Assessment and Response},
  author={Wang, Junjue and Xuan, Weihao and Qi, Heli and Liu, Zhihao and Liu, Kunyi and Wu, Yuhan and Chen, Hongruixuan and Song, Jian and Xia, Junshi and Zheng, Zhuo and others},
  journal={arXiv preprint arXiv:2505.21089},
  year={2025}
}

@article{dynamicvl,
  title={DynamicVL: Benchmarking Multimodal Large Language Models for Dynamic City Understanding},
  author={Xuan, Weihao and Wang, Junjue and Qi, Heli and Chen, Zihang and Zheng, Zhuo and Zhong, Yanfei and Xia, Junshi and Yokoya, Naoto},
  journal={arXiv preprint arXiv:2505.21076},
  year={2025}
}

@inproceedings{bleu,
  title={Bleu: a method for automatic evaluation of machine translation},
  author={Papineni, Kishore and Roukos, Salim and Ward, Todd and Zhu, Wei-Jing},
  booktitle={Proceedings of the 40th annual meeting of the Association for Computational Linguistics},
  pages={311--318},
  year={2002}
}

@inproceedings{rouge,
  title={Rouge: A package for automatic evaluation of summaries},
  author={Lin, Chin-Yew},
  booktitle={Text summarization branches out},
  pages={74--81},
  year={2004}
}

@inproceedings{kadid10k,
  title={KADID-10k: A large-scale artificially distorted IQA database},
  author={Lin, Hanhe and Hosu, Vlad and Saupe, Dietmar},
  booktitle={2019 Eleventh International Conference on Quality of Multimedia Experience (QoMEX)},
  pages={1--3},
  year={2019},
  organization={IEEE}
}

@article{TID2013,
  title={Image database TID2013: Peculiarities, results and perspectives},
  author={Ponomarenko, Nikolay and Jin, Lina and Ieremeiev, Oleg and Lukin, Vladimir and Egiazarian, Karen and Astola, Jaakko and Vozel, Benoit and Chehdi, Kacem and Carli, Marco and Battisti, Federica and others},
  journal={Signal processing: Image communication},
  volume={30},
  pages={57--77},
  year={2015},
  publisher={Elsevier}
}

@article{LIVE,
  title={A statistical evaluation of recent full reference image quality assessment algorithms},
  author={Sheikh, Hamid R and Sabir, Muhammad F and Bovik, Alan C and others},
  journal={IEEE Trans. Image Process.},
  volume={15},
  number={11},
  pages={3440--3451},
  year={2006}
}

@article{mmvet,
  title={Mm-vet: Evaluating large multimodal models for integrated capabilities},
  author={Yu, Weihao and Yang, Zhengyuan and Li, Linjie and Wang, Jianfeng and Lin, Kevin and Liu, Zicheng and Wang, Xinchao and Wang, Lijuan},
  journal={arXiv preprint arXiv:2308.02490},
  year={2023}
}

@article{lin2025rs,
  title={Rs-moe: A vision-language model with mixture of experts for remote sensing image captioning and visual question answering},
  author={Lin, Hui and Hong, Danfeng and Ge, Shuhang and Luo, Chuyao and Jiang, Kai and Jin, Hao and Wen, Congcong},
  journal={IEEE Transactions on Geoscience and Remote Sensing},
  year={2025},
  publisher={IEEE}
}

@article{superresolution,
  title={Remote sensing image super-resolution and object detection: Benchmark and state of the art},
  author={Wang, Yi and Bashir, Syed Muhammad Arsalan and Khan, Mahrukh and Ullah, Qudrat and Wang, Rui and Song, Yilin and Guo, Zhe and Niu, Yilong},
  journal={Expert Systems with Applications},
  volume={197},
  pages={116793},
  year={2022},
  publisher={Elsevier}
}

@article{denoising,
  title={RCST: Residual context-sharing transformer cascade to approximate Taylor expansion for remote sensing image denoising},
  author={Huang, Zhenghua and Yang, Yang and Yu, Hao and Li, Qian and Shi, Yu and Zhang, Yaozong and Fang, Hao},
  journal={IEEE Transactions on Geoscience and Remote Sensing},
  volume={63},
  pages={1--15},
  year={2025},
  publisher={IEEE}
}

@article{deblur,
  title={Jointly RS Image Deblurring and Super-Resolution With Adjustable-Kernel and Multi-Domain Attention},
  author={Zhang, Yan and Zheng, Pengcheng and Zeng, Chengxiao and Xiao, Bin and Li, Zhenghao and Gao, Xinbo},
  journal={IEEE Transactions on Geoscience and Remote Sensing},
  year={2024},
  publisher={IEEE}
}

@inproceedings{coompression,
  title={Remote sensing image compression: A review},
  author={Zhou, Shichao and Deng, Chenwei and Zhao, Baojun and Xia, Yatong and Li, Qisheng and Chen, Zhenzhong},
  booktitle={2015 IEEE International conference on multimedia big data},
  pages={406--410},
  year={2015},
  organization={IEEE}
}

@article{inpainting,
  title={Image Inpainting and Digital Camouflage: Methods, Applications, and Perspectives for Remote Sensing},
  author={Karwowska, Kinga and Wierzbicki, Damian and Kedzierski, Michal},
  journal={IEEE Journal of Selected Topics in Applied Earth Observations and Remote Sensing},
  year={2025},
  publisher={IEEE}
}

@article{gu2024survey,
  title={A survey on llm-as-a-judge},
  author={Gu, Jiawei and Jiang, Xuhui and Shi, Zhichao and Tan, Hexiang and Zhai, Xuehao and Xu, Chengjin and Li, Wei and Shen, Yinghan and Ma, Shengjie and Liu, Honghao and others},
  journal={The Innovation},
  year={2024},
  publisher={Elsevier}
}

@inproceedings{kim2024prometheus,
  title={Prometheus 2: An open source language model specialized in evaluating other language models},
  author={Kim, Seungone and Suk, Juyoung and Longpre, Shayne and Lin, Bill Yuchen and Shin, Jamin and Welleck, Sean and Neubig, Graham and Lee, Moontae and Lee, Kyungjae and Seo, Minjoon},
  booktitle={Proceedings of the 2024 Conference on Empirical Methods in Natural Language Processing},
  pages={4334--4353},
  year={2024}
}

@inproceedings{qinstruct,
  title={Q-instruct: Improving low-level visual abilities for multi-modality foundation models},
  author={Wu, Haoning and Zhang, Zicheng and Zhang, Erli and Chen, Chaofeng and Liao, Liang and Wang, Annan and Xu, Kaixin and Li, Chunyi and Hou, Jingwen and Zhai, Guangtao and others},
  booktitle={Proceedings of the IEEE/CVF conference on computer vision and pattern recognition},
  pages={25490--25500},
  year={2024}
}

@inproceedings{depictqa,
  title={Depicting beyond scores: Advancing image quality assessment through multi-modal language models},
  author={You, Zhiyuan and Li, Zheyuan and Gu, Jinjin and Yin, Zhenfei and Xue, Tianfan and Dong, Chao},
  booktitle={European Conference on Computer Vision},
  pages={259--276},
  year={2024},
  organization={Springer}
}

@inproceedings{Q-Scorer,
  title={Revisiting MLLM Based Image Quality Assessment: Errors and Remedy},
  author={Tang, Zhenchen and Yang, Songlin and Peng, Bo and Wang, Zichuan and Dong, Jing},
  booktitle={Proceedings of the AAAI Conference on Artificial Intelligence},
  volume={40},
  number={11},
  pages={9475--9483},
  year={2026}
}

@article{zhu2017deep,
  title={Deep learning in remote sensing: A comprehensive review and list of resources},
  author={Zhu, Xiao Xiang and Tuia, Devis and Mou, Lichao and Xia, Gui-Song and Zhang, Liangpei and Xu, Feng and Fraundorfer, Friedrich},
  journal={IEEE geoscience and remote sensing magazine},
  volume={5},
  number={4},
  pages={8--36},
  year={2017},
  publisher={IEEE}
}

@article{li2024vision,
  title={Vision-language models in remote sensing: Current progress and future trends},
  author={Li, Xiang and Wen, Congcong and Hu, Yuan and Yuan, Zhenghang and Zhu, Xiao Xiang},
  journal={IEEE Geoscience and Remote Sensing Magazine},
  volume={12},
  number={2},
  pages={32--66},
  year={2024},
  publisher={IEEE}
}

@article{tuia2022perspectives,
  title={Perspectives in machine learning for wildlife conservation},
  author={Tuia, Devis and Kellenberger, Benjamin and Beery, Sara and Costelloe, Blair R and Zuffi, Silvia and Risse, Benjamin and Mathis, Alexander and Mathis, Mackenzie W and Van Langevelde, Frank and Burghardt, Tilo and others},
  journal={Nature communications},
  volume={13},
  number={1},
  pages={792},
  year={2022},
  publisher={Nature Publishing Group UK London}
}

@article{wei2025bpme,
  title={BPME: A Blind Perceptual Metric-Driven Method for Enhancing Degraded Polar Optical Remote Sensing Imagery},
  author={Wei, Zijun and Lan, Chaozhen and Yao, Fushan and Wang, Longhao and Gao, Tian and Yu, Hanyang},
  journal={IEEE Transactions on Geoscience and Remote Sensing},
  volume={64},
  pages={1--13},
  year={2025},
  publisher={IEEE}
}

@article{an2024pretrain,
  title={Pretrain a remote sensing foundation model by promoting intra-instance similarity},
  author={An, Xiao and He, Wei and Zou, Jiaqi and Yang, Guangyi and Zhang, Hongyan},
  journal={IEEE Transactions on Geoscience and Remote Sensing},
  volume={62},
  pages={1--15},
  year={2024},
  publisher={IEEE}
}

@article{huang2025task,
  title={Task-Guided Prompting for Unified Remote Sensing Image Restoration},
  author={Huang, Wenli and Wu, Yang and Xin, Xiaomeng and Liu, Zhihong and Wang, Jinjun and Deng, Ye},
  journal={IEEE Transactions on Geoscience and Remote Sensing},
  volume={64},
  pages={1--17},
  year={2025},
  publisher={IEEE}
}

@article{qalign,
  title={Q-align: Teaching lmms for visual scoring via discrete text-defined levels},
  author={Wu, Haoning and Zhang, Zicheng and Zhang, Weixia and Chen, Chaofeng and Liao, Liang and Li, Chunyi and Gao, Yixuan and Wang, Annan and Zhang, Erli and Sun, Wenxiu and others},
  journal={arXiv preprint arXiv:2312.17090},
  year={2023}
}

@article{you2025enhancing,
  title={Enhancing Descriptive Image Quality Assessment With a Large-Scale Multi-Modal Dataset},
  author={You, Zhiyuan and Gu, Jinjin and Cai, Xin and Li, Zheyuan and Zhu, Kaiwen and Dong, Chao and Xue, Tianfan},
  journal={IEEE Transactions on Image Processing},
  volume={34},
  pages={8201--8215},
  year={2025},
  publisher={IEEE}
}

@inproceedings{wu2024towards,
  title={Towards open-ended visual quality comparison},
  author={Wu, Haoning and Zhu, Hanwei and Zhang, Zicheng and Zhang, Erli and Chen, Chaofeng and Liao, Liang and Li, Chunyi and Wang, Annan and Sun, Wenxiu and Yan, Qiong and others},
  booktitle={European Conference on Computer Vision},
  pages={360--377},
  year={2024},
  organization={Springer}
}

@article{brisque,
  title={No-reference image quality assessment in the spatial domain},
  author={Mittal, Anish and Moorthy, Anush Krishna and Bovik, Alan Conrad},
  journal={IEEE Transactions on image processing},
  volume={21},
  number={12},
  pages={4695--4708},
  year={2012},
  publisher={IEEE}
}

@article{niqe,
  title={Making a “completely blind” image quality analyzer},
  author={Mittal, Anish and Soundararajan, Rajiv and Bovik, Alan C},
  journal={IEEE Signal processing letters},
  volume={20},
  number={3},
  pages={209--212},
  year={2012},
  publisher={IEEE}
}

@inproceedings{musiq,
  title={Musiq: Multi-scale image quality transformer},
  author={Ke, Junjie and Wang, Qifei and Wang, Yilin and Milanfar, Peyman and Yang, Feng},
  booktitle={Proceedings of the IEEE/CVF international conference on computer vision},
  pages={5148--5157},
  year={2021}
}

@inproceedings{clipiqa,
  title={Exploring clip for assessing the look and feel of images},
  author={Wang, Jianyi and Chan, Kelvin CK and Loy, Chen Change},
  booktitle={Proceedings of the AAAI conference on artificial intelligence},
  volume={37},
  number={2},
  pages={2555--2563},
  year={2023}
}

@inproceedings{liqe,
  title={Blind image quality assessment via vision-language correspondence: A multitask learning perspective},
  author={Zhang, Weixia and Zhai, Guangtao and Wei, Ying and Yang, Xiaokang and Ma, Kede},
  booktitle={Proceedings of the IEEE/CVF conference on computer vision and pattern recognition},
  pages={14071--14081},
  year={2023}
}

@article{yan2019nriqa,
  title={No-reference remote sensing image quality assessment based on gradient-weighted natural scene statistics in spatial domain},
  author={Yan, Junhua and Bai, Xuehan and Xiao, Yongqi and Zhang, Yin and Lv, Xiangyang},
  journal={Journal of Electronic Imaging},
  volume={28},
  number={1},
  pages={013033--013033},
  year={2019},
  publisher={Society of Photo-Optical Instrumentation Engineers}
}

@article{bm_iqe,
  title={BM-IQE: An image quality evaluator with block-matching for both real-life scenes and remote sensing scenes},
  author={Xu, Ningshan and Ma, Dongao and Ren, Guoqiang and Huang, Yongmei},
  journal={Sensors},
  volume={20},
  number={12},
  pages={3472},
  year={2020},
  publisher={MDPI}
}

@article{prsiqd,
  title={A large-scale remote sensing database for subjective and objective quality assessment of pansharpened images},
  author={Xiong, Yiming and Shao, Feng and Meng, Xiangchao and Jiang, Qiuping and Sun, Weiwei and Fu, Randi and Ho, Yo-Sung},
  journal={Journal of Visual Communication and Image Representation},
  volume={73},
  pages={102947},
  year={2020},
  publisher={Elsevier}
}

@article{nriqa_uav_hsi,
  title={NR-IQA for UAV hyperspectral image based on distortion constructing, feature screening, and machine learning},
  author={Tian, Wenzhong and Sanchez-Azofeifa, Arturo and Kan, Za and Zhao, Qingzhan and Zhang, Guoshun and Wu, Yuzhen and Jiang, Kai},
  journal={International Journal of Applied Earth Observation and Geoinformation},
  volume={133},
  pages={104130},
  year={2024},
  publisher={Elsevier}
}

@article{lu2025vision,
  title={Vision foundation models in remote sensing: A survey},
  author={Lu, Siqi and Guo, Junlin and Zimmer-Dauphinee, James R and Nieusma, Jordan M and Wang, Xiao and VanValkenburgh, Parker and Wernke, Steven A and Huo, Yuankai},
  journal={IEEE Geoscience and Remote Sensing Magazine},
  volume={13},
  number={3},
  pages={190--215},
  year={2025},
  publisher={IEEE}
}

@article{gee,
  title={Google Earth Engine: Planetary-scale geospatial analysis for everyone},
  author={Gorelick, Noel and Hancher, Matt and Dixon, Mike and Ilyushchenko, Simon and Thau, David and Moore, Rebecca},
  journal={Remote sensing of Environment},
  volume={202},
  pages={18--27},
  year={2017},
  publisher={Elsevier}
}

@article{hu2025rsgpt,
  title={Rsgpt: A remote sensing vision language model and benchmark},
  author={Hu, Yuan and Yuan, Jianlong and Wen, Congcong and Lu, Xiaonan and Liu, Yu and Li, Xiang},
  journal={ISPRS Journal of Photogrammetry and Remote Sensing},
  volume={224},
  pages={272--286},
  year={2025},
  publisher={Elsevier}
}

@article{li2025lhrsbotnava,
  title={Lhrs-bot-nova: Improved multimodal large language model for remote sensing vision-language interpretation},
  author={Li, Zhenshi and Muhtar, Dilxat and Gu, Feng and He, Yanglangxing and Zhang, Xueliang and Xiao, Pengfeng and He, Guangjun and Zhu, Xiaoxiang},
  journal={ISPRS Journal of Photogrammetry and Remote Sensing},
  volume={227},
  pages={539--550},
  year={2025},
  publisher={Elsevier}
}

@inproceedings{soni2025earthdial,
  title={Earthdial: Turning multi-sensory earth observations to interactive dialogues},
  author={Soni, Sagar and Dudhane, Akshay and Debary, Hiyam and Fiaz, Mustansar and Munir, Muhammad Akhtar and Danish, Muhammad Sohail and Fraccaro, Paolo and Watson, Campbell D and Klein, Levente J and Khan, Fahad Shahbaz and others},
  booktitle={Proceedings of the Computer Vision and Pattern Recognition Conference},
  pages={14303--14313},
  year={2025}
}

@article{irvin2024teochat,
  title={Teochat: A large vision-language assistant for temporal earth observation data},
  author={Irvin, Jeremy Andrew and Liu, Emily Ruoyu and Chen, Joyce Chuyi and Dormoy, Ines and Kim, Jinyoung and Khanna, Samar and Zheng, Zhuo and Ermon, Stefano},
  journal={arXiv preprint arXiv:2410.06234},
  year={2024}
}

@article{yao2025falcon,
  title={Falcon: A remote sensing vision-language foundation model},
  author={Yao, Kelu and Xu, Nuo and Yang, Rong and Xu, Yingying and Gao, Zhuoyan and Kitrungrotsakul, Titinunt and Ren, Yi and Zhang, Pu and Wang, Jin and Wei, Ning and others},
  journal={arXiv preprint arXiv:2503.11070},
  year={2025}
}

@article{liu2024rsunivlm,
  title={Rsunivlm: A unified vision language model for remote sensing via granularity-oriented mixture of experts},
  author={Liu, Xu and Lian, Zhouhui},
  journal={arXiv preprint arXiv:2412.05679},
  year={2024}
}

@article{liu2025skymoe,
  title={SkyMoE: A Vision-Language Foundation Model for Enhancing Geospatial Interpretation with Mixture of Experts},
  author={Liu, Jiaqi and Fu, Ronghao and Sun, Lang and Liu, Haoran and Yang, Xiao and Zhang, Weipeng and Na, Xu and Duan, Zhuoran and Yang, Bo},
  journal={arXiv preprint arXiv:2512.02517},
  year={2025}
}

@article{qwen3vl,
    author = {Shuai Bai and Yuxuan Cai and Ruizhe Chen and Keqin Chen and Xionghui Chen and Zesen Cheng and Lianghao Deng and Wei Ding and Chang Gao and Chunjiang Ge and Wenbin Ge and Zhifang Guo and Qidong Huang and Jie Huang and Fei Huang and Binyuan Hui and Shutong Jiang and Zhaohai Li and Mingsheng Li and Mei Li and Kaixin Li and Zicheng Lin and Junyang Lin and Xuejing Liu and Jiawei Liu and Chenglong Liu and Yang Liu and Dayiheng Liu and Shixuan Liu and Dunjie Lu and Ruilin Luo and Chenxu Lv and Rui Men and Lingchen Meng and Xuancheng Ren and Xingzhang Ren and Sibo Song and Yuchong Sun and Jun Tang and Jianhong Tu and Jianqiang Wan and Peng Wang and Pengfei Wang and Qiuyue Wang and Yuxuan Wang and Tianbao Xie and Yiheng Xu and Haiyang Xu and Jin Xu and Zhibo Yang and Mingkun Yang and Jianxin Yang and An Yang and Bowen Yu and Fei Zhang and Hang Zhang and Xi Zhang and Bo Zheng and Humen Zhong and Jingren Zhou and Fan Zhou and Jing Zhou and Yuanzhi Zhu and Ke Zhu},
    title = {Qwen3-VL Technical Report},
    journal = {arXiv preprint arXiv:2511.21631},
    year = {2025}
}

@article{internvl3.5,
  title={Internvl3. 5: Advancing open-source multimodal models in versatility, reasoning, and efficiency},
  author={Wang, Weiyun and Gao, Zhangwei and Gu, Lixin and Pu, Hengjun and Cui, Long and Wei, Xingguang and Liu, Zhaoyang and Jing, Linglin and Ye, Shenglong and Shao, Jie and others},
  journal={arXiv preprint arXiv:2508.18265},
  year={2025}
}

@article{ovis1.6,
  title={Ovis: Structural embedding alignment for multimodal large language model},
  author={Lu, Shiyin and Li, Yang and Chen, Qing-Guo and Xu, Zhao and Luo, Weihua and Zhang, Kaifu and Ye, Han-Jia},
  journal={arXiv preprint arXiv:2405.20797},
  year={2024}
}

@article{ovis2.5,
  title={Ovis2. 5 technical report},
  author={Lu, Shiyin and Li, Yang and Xia, Yu and Hu, Yuwei and Zhao, Shanshan and Ma, Yanqing and Wei, Zhichao and Li, Yinglun and Duan, Lunhao and Zhao, Jianshan and others},
  journal={arXiv preprint arXiv:2508.11737},
  year={2025}
}

@article{llama3.2,
  title={The llama 3 herd of models},
  author={Grattafiori, Aaron and Dubey, Abhimanyu and Jauhri, Abhinav and Pandey, Abhinav and Kadian, Abhishek and Al-Dahle, Ahmad and Letman, Aiesha and Mathur, Akhil and Schelten, Alan and Vaughan, Alex and others},
  journal={arXiv preprint arXiv:2407.21783},
  year={2024}
}

@inproceedings{molmo,
  title={Molmo and pixmo: Open weights and open data for state-of-the-art vision-language models},
  author={Deitke, Matt and Clark, Christopher and Lee, Sangho and Tripathi, Rohun and Yang, Yue and Park, Jae Sung and Salehi, Mohammadreza and Muennighoff, Niklas and Lo, Kyle and Soldaini, Luca and others},
  booktitle={Proceedings of the Computer Vision and Pattern Recognition Conference},
  pages={91--104},
  year={2025}
}

@article{phi4,
  title={Phi-4-mini technical report: Compact yet powerful multimodal language models via mixture-of-loras},
  author={Abouelenin, Abdelrahman and Ashfaq, Atabak and Atkinson, Adam and Awadalla, Hany and Bach, Nguyen and Bao, Jianmin and Benhaim, Alon and Cai, Martin and Chaudhary, Vishrav and Chen, Congcong and others},
  journal={arXiv preprint arXiv:2503.01743},
  year={2025}
}

@article{sensenova,
  title={Scaling spatial intelligence with multimodal foundation models},
  author={Cai, Zhongang and Wang, Ruisi and Gu, Chenyang and Pu, Fanyi and Xu, Junxiang and Wang, Yubo and Yin, Wanqi and Yang, Zhitao and Wei, Chen and Sun, Qingping and others},
  journal={arXiv preprint arXiv:2511.13719},
  year={2025}
}

@article{deepseekvl,
  title={Deepseek-vl: towards real-world vision-language understanding},
  author={Lu, Haoyu and Liu, Wen and Zhang, Bo and Wang, Bingxuan and Dong, Kai and Liu, Bo and Sun, Jingxiang and Ren, Tongzheng and Li, Zhuoshu and Yang, Hao and others},
  journal={arXiv preprint arXiv:2403.05525},
  year={2024}
}

@inproceedings{llava,
  title={Improved baselines with visual instruction tuning},
  author={Liu, Haotian and Li, Chunyuan and Li, Yuheng and Lee, Yong Jae},
  booktitle={Proceedings of the IEEE/CVF conference on computer vision and pattern recognition},
  pages={26296--26306},
  year={2024}
}

@article{llava-ov,
  title={Llava-onevision: Easy visual task transfer},
  author={Li, Bo and Zhang, Yuanhan and Guo, Dong and Zhang, Renrui and Li, Feng and Zhang, Hao and Zhang, Kaichen and Zhang, Peiyuan and Li, Yanwei and Liu, Ziwei and others},
  journal={arXiv preprint arXiv:2408.03326},
  year={2024}
}

@article{kimivl,
  title={Kimi-vl technical report},
  author={Team, Kimi and Du, Angang and Yin, Bohong and Xing, Bowei and Qu, Bowen and Wang, Bowen and Chen, Cheng and Zhang, Chenlin and Du, Chenzhuang and Wei, Chu and others},
  journal={arXiv preprint arXiv:2504.07491},
  year={2025}
}

@article{teochat,
  title={Teochat: A large vision-language assistant for temporal earth observation data},
  author={Irvin, Jeremy Andrew and Liu, Emily Ruoyu and Chen, Joyce Chuyi and Dormoy, Ines and Kim, Jinyoung and Khanna, Samar and Zheng, Zhuo and Ermon, Stefano},
  journal={arXiv preprint arXiv:2410.06234},
  year={2024}
}







\clearpage
\appendix

\hypersetup{
linkcolor=black
}

\startcontents[appendix]
\printcontents[appendix]{}{0}{
    \section*{Appendix Contents}
}

\newpage
\appendix

\section{Benchmark Taxonomy Details}
\label{benchmark_taxonomy}
To support a thorough and unbiased evaluation of VLMs, we organize SenseBench into 6 L-1 dimensions and further 22 L-2 distortions, grounded in widely adopted remote sensing quality assessment standards (ISO 19157-1:2023\footnote{\url{https://www.iso.org/standard/78900.html}}, ISO 19115-2:2019\footnote{\url{https://www.iso.org/standard/67039.html}}, GB/T 24356--2023\footnote{\url{https://openstd.samr.gov.cn/bzgk/std/newGbInfo?hcno=2874EFAC7523FB293E6AF2E4068CEB02}}). We first present the detailed definition of each L-2 distortion, accompanied by representative examples, and then report the overall statistics of SenseBench.

\subsection{Definition of Each L-2 Distortions}
\label{Appendix_A_1}

\begin{figure*}[!ht]
  \centering
  \resizebox{0.95\linewidth}{!}{
   \includegraphics{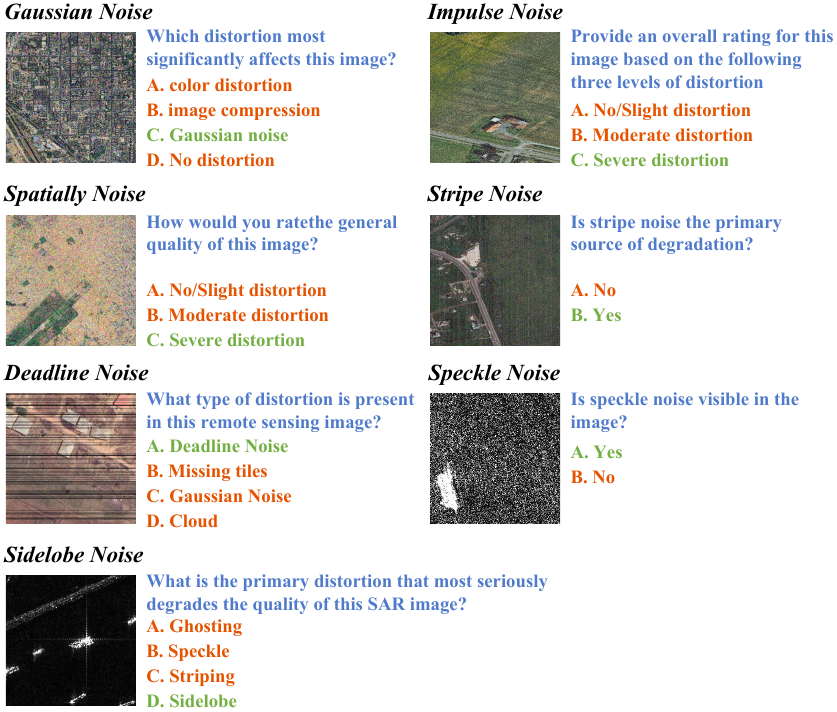}}
   \caption{Examples of L-2 distortions within the L-1 dimension of Image Noise.}
   \label{fig:L2_Noise}
   \vspace{-5mm}
\end{figure*}

\paragraph{Noise.} This category models sensor-inherent noise and signal transmission perturbations, comprising 7 L-2 distortions that span statistical sensor noise, detector array artifacts, and coherent imaging interference. In Figure~\ref{fig:L2_Noise}, we present some visualization examples for this dimension. 
\begin{itemize}
\item \textit{Gaussian Noise}: This distortion simulates thermal noise and readout noise in sensor electronics, modeled as additive perturbations following a Gaussian distribution. It represents one of the most prevalent noise types affecting RSI quality across all sensor platforms.
\item \textit{Impulse Noise}: This distortion replicates sudden pixel-level corruptions caused by transmission errors or sensor malfunctions, manifesting as sparsely distributed extreme values that disrupt local image statistics.
\item \textit{Spatially Correlated Noise}: This distortion models noise components that exhibit spatial dependence rather than pixel-wise independence, reflecting realistic sensor behaviors where neighboring pixels share correlated perturbations.
\item \textit{Stripe Noise}: This distortion arises from inter-detector calibration mismatches in push-broom scanners, producing line-structured artifacts aligned with the scanning direction.
\item \textit{Dead-line Noise}: This distortion corresponds to the complete failure of individual detector elements, resulting in continuous missing lines that severely compromise downstream interpretation.
\item \textit{Speckle Noise}: This distortion is an inherent granular pattern in \textbf{SAR} imagery caused by the coherent summation of backscattered signals from sub-resolution scatterers, exhibiting multiplicative rather than additive behavior.
\item \textit{Sidelobe Noise}: This distortion originates from the point spread function of \textbf{SAR} systems, where strong scatterers produce cross-shaped energy leakage into adjacent pixels and obscure neighboring targets.
\end{itemize}

\begin{figure*}[!t]
  \centering
  \resizebox{0.95\linewidth}{!}{
   \includegraphics{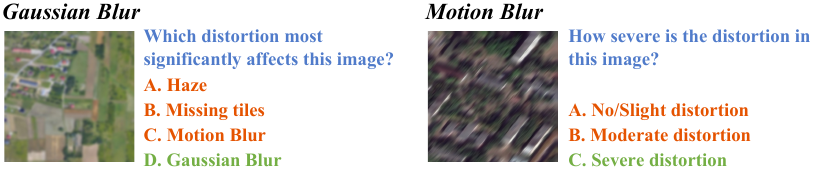}}
   \caption{Examples of L-2 distortions within the L-1 dimension of Image Blur}
   \label{fig:L2_Blur}
   \vspace{-4mm}
\end{figure*}

\begin{figure*}[!t]
  \centering
  \resizebox{0.95\linewidth}{!}{
   \includegraphics{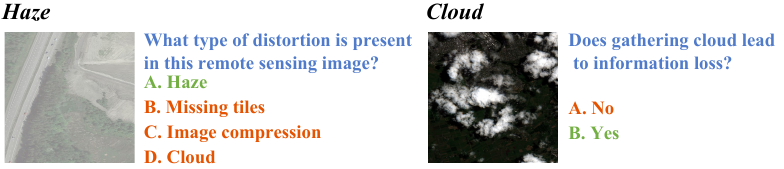}}
   \caption{Examples of L-2 distortions within the L-1 dimension of Image Cloud.}
   \label{fig:L2_Cloud}
   \vspace{-4mm}
\end{figure*}

\paragraph{Blur.} This category captures resolution loss arising from optical imperfections or platform–target relative motion In Figure~\ref{fig:L2_Blur}, we present some visualization examples for this dimension. 
\begin{itemize}
\item \textit{Gaussian Blur}: This distortion models isotropic point-spread function (PSF) broadening caused by optical defocus, atmospheric turbulence, and diffraction-limited imaging, leading to uniform attenuation of high-frequency details across the image.
\item \textit{Motion Blur}: This distortion models directional PSF elongation induced by satellite along-track velocity, attitude jitter, and target motion within the integration window, with kernel orientation and length sampled to reflect typical low-Earth-orbit acquisition geometries.
\end{itemize}

\paragraph{Cloud.} This category quantifies visibility degradation caused by atmospheric particulates and hydrometeors, comprising 2 L-2 distortions. In Figure~\ref{fig:L2_Cloud}, we present some visualization examples for this dimension.
\begin{itemize}
\item \textit{Cloud}: This distortion introduces opaque occlusions through a hybrid pipeline, where cloud masks are either harvested from publicly available cloud-annotated datasets or generated procedurally, and are then alpha-composited with clean scenes to produce semantically plausible occlusions over surface content.
\item \textit{Haze}: This distortion is synthesized via the standard atmospheric scattering model:
\begin{equation}
I(x) = J(x)t(x) + A\bigl(1 - t(x)\bigr),
\label{eq:atmospheric_scattering_model}
    \end{equation}
    where the transmission map $t(x)$ and global airlight $A$ are randomized across samples to span a realistic range of optical depths, jointly attenuating contrast and shifting the spectral signature of surface reflectance.
\end{itemize}

\begin{figure*}[!ht]
  \centering
  \resizebox{0.95\linewidth}{!}{
   \includegraphics{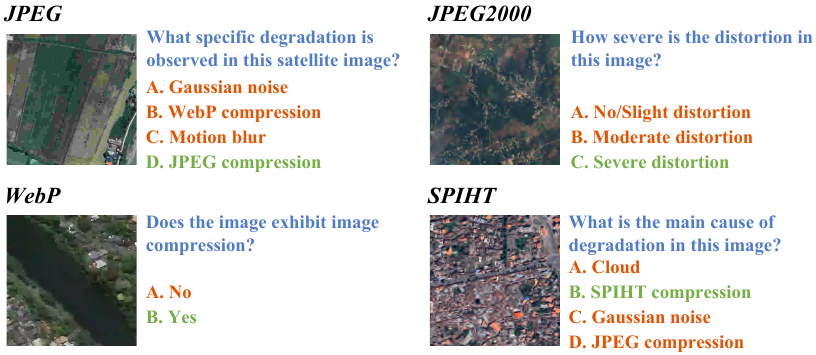}}
   \caption{Examples of L-2 distortions within the L-1 dimension of Image Compression.}
   \label{fig:L2_Compression}
   \vspace{-5mm}
\end{figure*}

\paragraph{Compression.} This category simulates lossy coding artifacts incurred during on-board compression, downlink transmission, and archival storage, comprising 4 L-2 distortions that span legacy and modern satellite codecs. In Figure~\ref{fig:L2_Compression}, we present some visualization examples for this dimension. 
\begin{itemize}
    \item \textit{JPEG}: This distortion produces DCT-based $ 8\times8 $ blocking artifacts caused by coarse quantization of frequency coefficients within independently coded macroblocks, representing the most widely deployed lossy compression standard in remote sensing pipelines.
    \item \textit{JPEG2000}: This distortion induces wavelet-based ringing and edge smearing arising from the truncation of high-frequency subbands, offering improved rate-distortion performance over JPEG at the cost of distinct boundary artifacts.
    \item \textit{WebP}: This distortion introduces predictive-block residuals from modern web-oriented compression, where intra-frame prediction errors and adaptive block partitioning give rise to artifact patterns absent in classical codecs.
    \item \textit{SPIHT}: This distortion produces progressive bit-plane truncation characteristic of embedded zerotree wavelet coders, which remain widely deployed in legacy spaceborne systems where progressive transmission is prioritized.
\end{itemize}

\begin{figure*}[!ht]
  \centering
  \resizebox{0.95\linewidth}{!}{
   \includegraphics{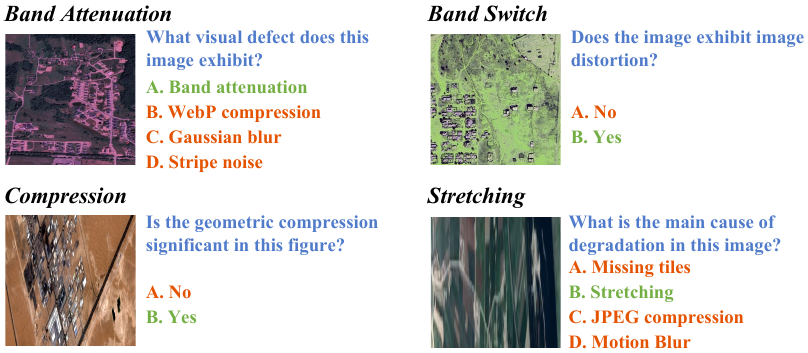}}
   \caption{Examples of L-2 distortions within the L-1 dimension of Image Correction.}
   \label{fig:L2_Correction}
   \vspace{-5mm}
\end{figure*}

\paragraph{Correction.} This category models residual artifacts left by imperfect spectral or geometric rectification during ground-segment processing, comprising 4 L-2 distortions that span spectral and geometric failure modes. In Figure~\ref{fig:L2_Correction}, we present some visualization examples for this dimension.
\begin{itemize}
\item \textit{Band Attenuation}: This distortion corresponds to per-channel radiometric gain loss from imperfect cross-calibration, where individual spectral bands are systematically darkened relative to the others, breaking the radiometric balance across channels.
\item \textit{Band Switch}: This distortion corresponds to channel permutation caused by metadata corruption during ground-segment processing, where spectral bands are assigned to incorrect color components and produce semantically implausible color renderings.
\item \textit{Compression}: This distortion models unintended axis-wise shrinkage of the image, where one spatial dimension is contracted relative to the other, typically arising from inaccurate orthorectification or sensor-model misregistration during geometric correction.
\item \textit{Stretching}: This distortion models unintended axis-wise elongation of the image, where one spatial dimension is expanded relative to the other, producing distorted aspect ratios that violate the metric consistency required for spatial reasoning.
\end{itemize}

\begin{figure*}[!ht]
  \centering
  \resizebox{0.95\linewidth}{!}{
   \includegraphics{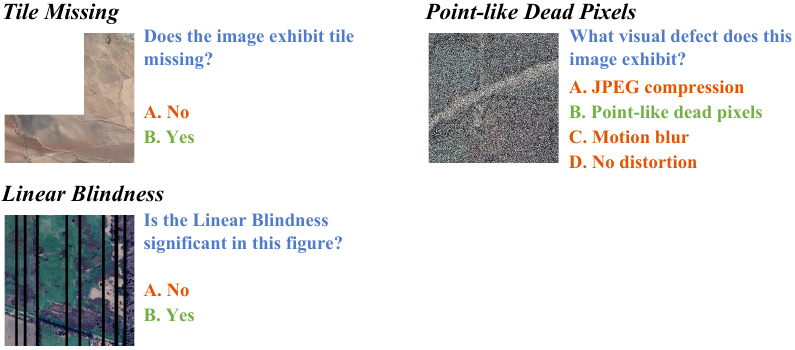}}
   \caption{Examples of L-2 distortions within the L-1 dimension of Image Missing.}
   \label{fig:L2_Missing}
   \vspace{-5mm}
\end{figure*}

\paragraph{Missing.} This category models data-loss patterns occurring across the acquisition–transmission chain, comprising 3 L-2 distortions that jointly cover the dominant operational dropout mechanisms. In Figure~\ref{fig:L2_Missing}, we present some visualization examples for this dimension.
\begin{itemize}
\item \textit{Tile Missing}: This distortion produces contiguous rectangular voids caused by packet loss during downlink transmission, where entire data blocks are dropped and replaced with zero-valued regions that disrupt the spatial continuity of the scene.
\item \textit{Point-like Dead Pixels}: This distortion introduces sparsely distributed isolated zero-valued pixels arising from random detector failures, manifesting as scattered point-wise voids that perturb local image statistics without affecting global structure.
\item \textit{Linear Blindness}: This distortion produces strip-shaped voids caused by systematic detector-array malfunction in push-broom imaging geometries, where the failure of one or more detector elements yields continuous missing lines aligned with the scanning direction.
\end{itemize}

\subsection{Statistics of SenseBench}
\label{Appendix_A_2}

All RS images in SenseBench are collected from globally distributed satellite imagery, public remote sensing datasets, and map platforms, covering diverse geographic regions and sensor modalities. To support balanced and comprehensive evaluation, we ensure an even distribution across distortion categories and task types during dataset construction, with dedicated subsets for both optical and SAR imagery. A detailed breakdown of data sources, distortion categories, and task-specific sample counts is provided in Table~\ref{tab:information_table}.


\newcommand{\attr}[4]{%
\makebox[1.8em][c]{#1}/%
\makebox[1.8em][c]{#2}/%
\makebox[1.8em][c]{#3}/%
\makebox[1.8em][c]{#4}%
}

\begin{table*}[!t]
\centering
\caption{The detailed statistics for SenseBench, a low-level perception benchmark for remote sensing VLMs. Abbreviations adopted: GEE for Google Earth Engine; W/W/H/D for Whether, What, How, and Description questions, respectively.}
\label{tab:information_table}

\setlength{\tabcolsep}{2pt}

\renewcommand{\arraystretch}{1}
\footnotesize

\begin{tabular}{lllcclc}
\toprule
\textbf{Modality} & \textbf{Level-1} & \textbf{Level-2} & \textbf{Data Source} & \textbf{Resolution} & \textbf{Image Size} & \textbf{Attributes (W/W/H/D)} \\
\midrule

\multirow{24}{*}{RGB} 
    & \multirow{3}{*}{Missing} & Missing tiles & \multirow{3}{*}{GEE} & \multirow{3}{*}{0.3$\sim$10m} & \multirow{3}{*}{512$\times$512} & \attr{155}{155}{--}{60} \\
    & & Blind flickering & & & & \attr{155}{155}{--}{60} \\
    & & Linear dead pixels & & & & \attr{155}{155}{--}{60} \\
    \cmidrule{2-7} 

    & \multirow{5}{*}{Noise} & Stripe noise & \multirow{5}{*}{GEE} & \multirow{5}{*}{0.3$\sim$10m} & \multirow{5}{*}{512$\times$512} & \attr{155}{155}{155}{60} \\
    & & Impulse noise & & & & \attr{155}{155}{155}{60} \\
    & & Deadline noise & & & & \attr{155}{155}{155}{60} \\
    & & Gaussian noise & & & & \attr{155}{155}{155}{60} \\
    & & Spatially correlated noise & & & & \attr{155}{155}{155}{60} \\
    \cmidrule{2-7}

    & \multirow{4}{*}{Correction} & Color band attenuation & \multirow{4}{*}{GEE} & \multirow{4}{*}{0.3$\sim$10m} & \multirow{4}{*}{512$\times$512} & \attr{155}{155}{--}{60} \\
    & & Geometric compression & & & & \attr{155}{155}{--}{60} \\
    & & Geometric stretch & & & & \attr{155}{155}{--}{60} \\
    & & Color band switch & & & & \attr{155}{155}{--}{60} \\
    \cmidrule{2-7}

    & \multirow{4}{*}{Compression} & Jpeg2000 & \multirow{4}{*}{GEE} & \multirow{4}{*}{0.3$\sim$10m} & \multirow{4}{*}{512$\times$512} & \attr{155}{155}{155}{60} \\
    & & Webp & & & & \attr{155}{155}{155}{60} \\
    & & Jpeg & & & & \attr{155}{155}{155}{60} \\
    & & Spiht & & & & \attr{155}{155}{155}{60} \\
    \cmidrule{2-7}

    & \multirow{5}{*}{Cloud} & \multirow{2}{*}{Haze} & GEE & 0.3$\sim$10m & 512$\times$512 & \multirow{2}{*}{\attr{155}{155}{155}{60}} \\
    & & & RICE Dataset & N/A & 512$\times$512 & \\
    \cmidrule{3-7}
    & & \multirow{3}{*}{Cloud} & GEE & 0.3$\sim$10m & 512$\times$512 & \multirow{3}{*}{\attr{155}{155}{155}{60}} \\
    & & & Sentinel-2 & 10m & 982$\times$982 & \\
    & & & RICE Dataset & N/A & 512$\times$512 & \\
    \cmidrule{2-7}

    & Multi-dist. & Multiple distortions & -- & -- & 512$\times$512 & \attr{300}{300}{--}{--} \\
    \cmidrule{2-7}

    & \multirow{2}{*}{Blur} & Gaussian blur & \multirow{2}{*}{GEE} & \multirow{2}{*}{0.3$\sim$10m} & \multirow{2}{*}{512$\times$512} & \attr{155}{155}{155}{60} \\
    & & Motion blur & & & & \attr{155}{155}{155}{60} \\
\midrule

\multirow{2}{*}{SAR} 
    & Sidelobe & Sidelobe noise & Multiple Datasets & 0.5$\sim$3m & 214$\times$241 & \attr{200}{200}{--}{--} \\
    & Speckle & Speckle noise & Multiple Datasets & 0.5$\sim$3m & 2048$\times$2048 & \attr{200}{200}{--}{--} \\
\midrule[\heavyrulewidth]

\multicolumn{6}{c}{\textbf{Total}} & \textbf{10815} \\
\bottomrule
\end{tabular}
\end{table*}

\section{Dataset Construction Details}
\label{app:dataset_construction}
\subsection{Data Collection and Geographic Sampling}

To construct a geographically diverse remote sensing image pool, we adopted a \textit{continent-balanced sampling strategy} rather than collecting images from a small number of commonly used benchmark regions. Since low-level degradations may interact with scene content, land-cover patterns, imaging geometry, and regional characteristics, a globally distributed image pool is necessary to reduce regional bias in evaluating the low-level visual perception ability of VLMs.

We first used continent-level vector boundaries from \textbf{Natural Earth}\footnote{\url{https://www.naturalearthdata.com}} to define the global sampling regions. For each continent, a predefined sampling quota was assigned, and clustering-based spatial sampling was performed within the corresponding boundary to encourage spatially dispersed coverage. The resulting continent-level distribution of sampled cities is summarized in Table~\ref{tab:continent_distribution}. We then collected candidate cities from the world-cities database and filtered out cities with populations below 100,000, as cities with very small populations were empirically found to have limited availability of suitable high-quality imagery. The remaining cities were selected by jointly considering population scale and geographic location, so that the final city set covers representative urban regions across different continents while maintaining spatial diversity.

\begin{table}[!ht]
\centering
\small
\caption{Continent-level distribution of sampled cities.}
\label{tab:continent_distribution}
\renewcommand{\arraystretch}{1.15}
\setlength{\tabcolsep}{5pt}
\resizebox{\columnwidth}{!}{%
\begin{tabular}{lccccccc}
\toprule
\textbf{Continent} & \textbf{Africa} & \textbf{Asia} & \textbf{Europe} & \textbf{North America} & \textbf{Oceania} & \textbf{South America} & \textbf{Total} \\
\midrule
\textbf{Numbers} & 130 & 210 & 160 & 160 & 29 & 120 & 809 \\
\bottomrule
\end{tabular}
}
\end{table}

For each selected city, we obtained its urban boundary from \textbf{OpenStreetMap}\footnote{\url{https://www.openstreetmap.org}} and randomly sampled points within the boundary. Remote sensing image patches centered at these points were downloaded from \textbf{Google Earth}\footnote{\url{https://earth.google.com}}. All collected image patches were finally cropped or resized to a unified spatial size of $512 \times 512$ for subsequent distortion synthesis and benchmark construction. This process yielded 26,493 RS image patches, which formed the initial source \textit{image pool} for the subsequent filtering and benchmark construction pipeline.

\subsection{Data Filtering and Cleaning}
After image collection, we applied a three-stage filtering procedure to refine the source \textit{image pool} before distortion synthesis. The filtering process was designed to remove invalid files, low-information samples, and visually unsuitable images.

\begin{itemize}
    \item \textbf{File-level filtering.}
    We first checked the basic file properties of the collected images. Non-TIFF files were discarded to maintain a consistent data format. Files with abnormally small storage sizes were removed, as they usually indicate failed downloads or incomplete files. We also excluded images with insufficient spatial resolution and removed duplicated image patches to avoid redundant samples in the source pool.

    \item \textbf{Information-level filtering.}
    We then filtered the images according to their effective visual content. Pure-white images were removed because they contain little or no usable scene information. Images dominated by ocean coverage or snow coverage were also excluded. Although these images may be valid remote sensing observations, they provide limited structural and land-cover details, making them less suitable for evaluating low-level visual degradations.

    \item \textbf{Human filtering.}
    Finally, the remaining images were manually inspected. Samples with abnormal visual content, severe acquisition artifacts, or other obvious quality issues were removed. This final screening step further improved the reliability of the clean image pool used for subsequent degradation synthesis.
\end{itemize}

Through the above filtering process, the initial source pool was reduced from 26,493 to 19,768 image patches, with approximately 25.38\% of the collected images removed. The remaining images formed the clean source image pool for subsequent benchmark construction.

\begin{figure*}[!t]
  \centering
  \resizebox{0.9\linewidth}{!}{
   \includegraphics{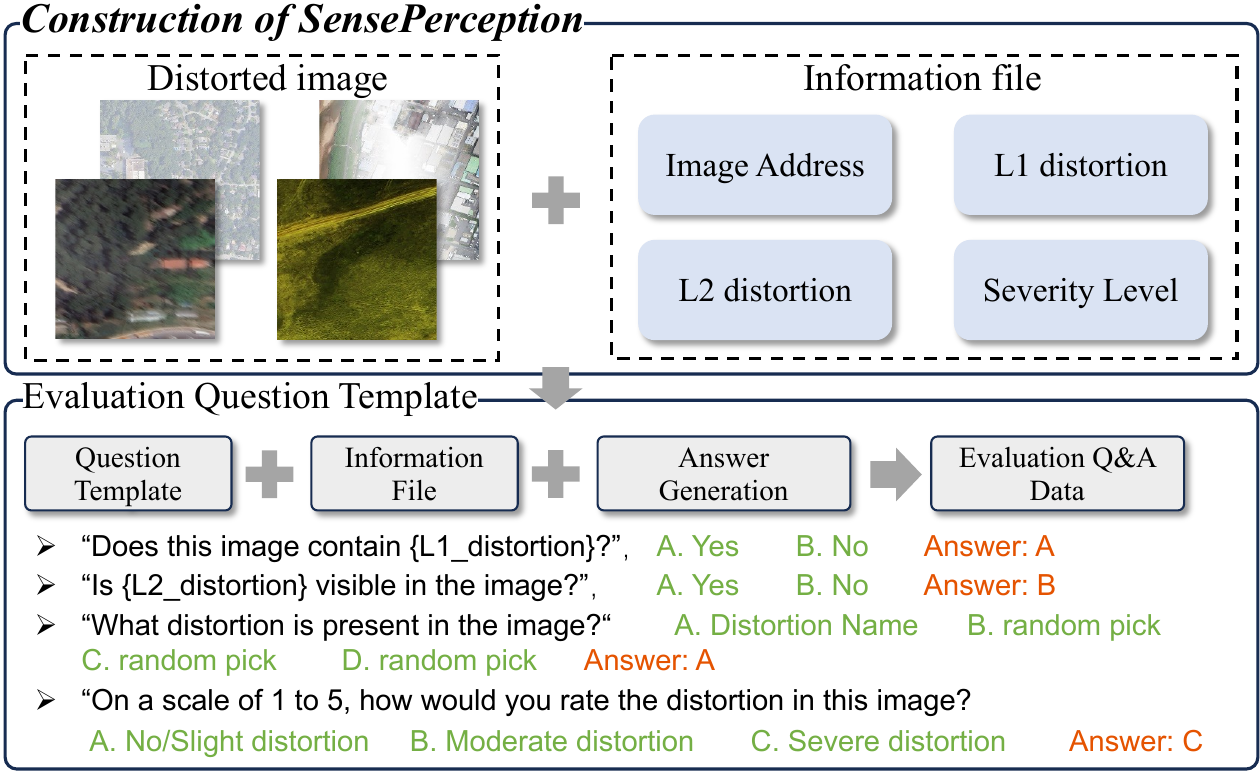}}
   \caption{Details of the construction of \textit{SensePerception}.}
   \label{fig: construction_per}
   \vspace{-5mm}
\end{figure*}

\subsection{Distortion Synthesis and Benchmark Construction}
Based on the cleaned \textit{image pool}, we synthesized distorted remote sensing images according to the distortion taxonomy described in Appendix~\ref{benchmark_taxonomy}. For each degradation type, distorted samples were generated with predefined parameter settings so that the corresponding distortion category and severity level could be explicitly recorded. Each distorted image was paired with an information file, which contains the \texttt{Image Address}, \texttt{L1\_distortion}, \texttt{L2\_distortion}, and \texttt{Severity Level}. These image--metadata pairs were then used as the basic data units for constructing the subsequent \textit{SensePerception} and \textit{SenseDescription} tasks.

\paragraph{Construction of SensePerception.}
As illustrated in Fig.~\ref{fig: construction_per}, SensePerception is constructed by combining each distorted image with its associated information file. The information file records the image address, L2 distortion category, L3 distortion type, and severity level, which are used to instantiate predefined multiple-choice question templates and automatically derive the corresponding answers. Specifically, we design three types of questions: whether-type, what-type, and how-type questions. Whether-type questions evaluate whether a model can determine the presence or absence of a specified L2 or L3 degradation, such as ``Does this image contain haze?'' or ``Is Gaussian noise visible in the image?'' What-type questions assess whether a model can identify the correct distortion category or fine-grained distortion type from several candidate options, where distractors are randomly sampled from other distortion types. How-type questions evaluate whether a model can judge the degradation severity according to predefined quality levels, such as no/slight, moderate, and severe distortion. Through this template-based generation process, the distorted images and their metadata are converted into standardized evaluation Q\&A pairs for low-level visual perception assessment.

\paragraph{Construction of SenseDescription.}
\begin{figure*}[!t]
  \centering
  \resizebox{1\linewidth}{!}{
   \includegraphics{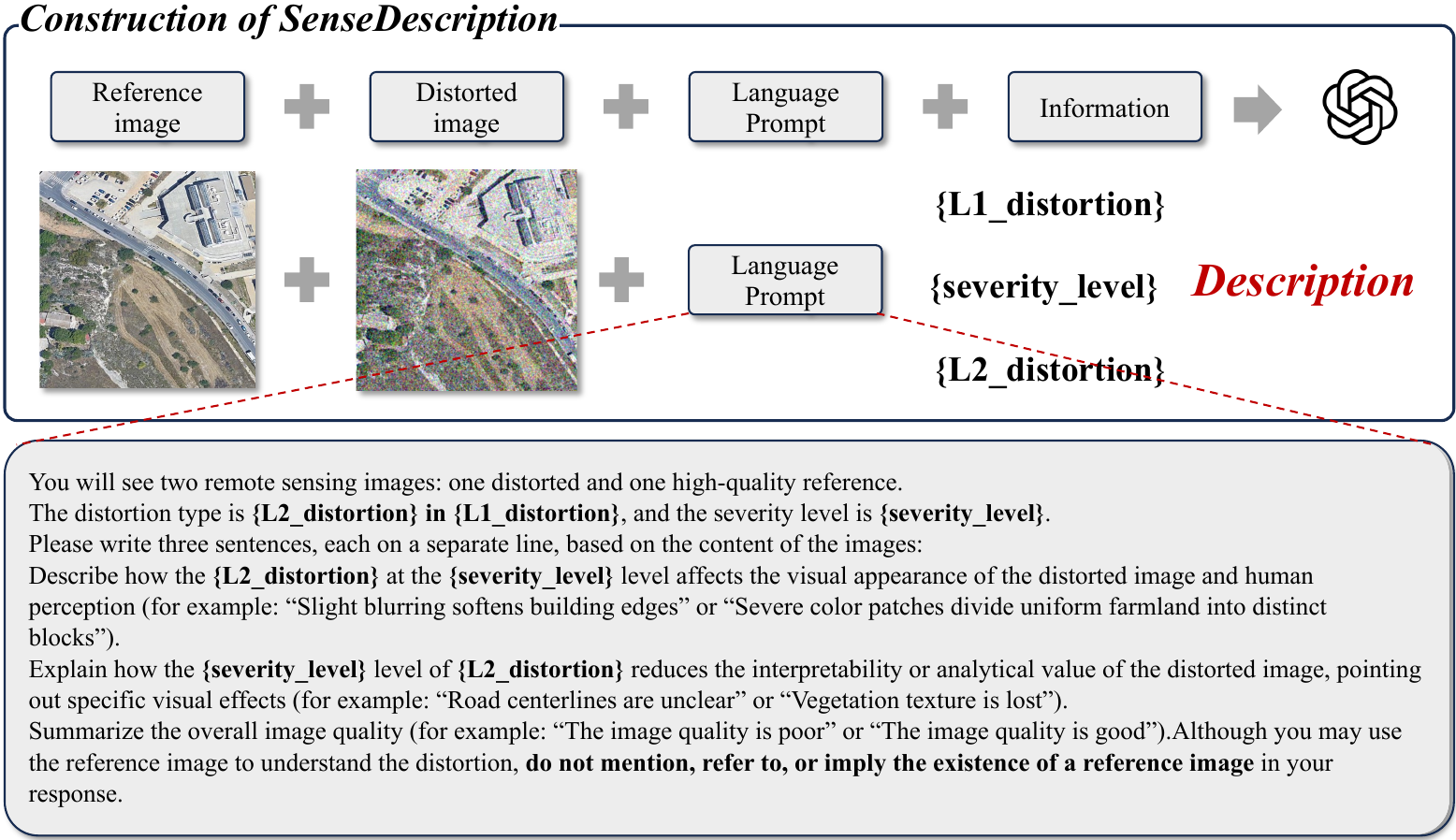}}
   \caption{Details of the construction of \textit{SenseDescription}.}
   \label{fig: construction_des}
   \vspace{-5mm}
\end{figure*}
As shown in Figure~\ref{fig: construction_des}, we further construct description-based diagnostic tasks to evaluate whether VLMs can generate faithful and informative quality descriptions beyond MCQ-style perception. For each selected distorted image, we provide the LVLM-assisted annotation pipeline with its corresponding high-quality reference image, a structured language prompt, and the associated distortion metadata, including \texttt{L2\_distortion}, \texttt{L3\_distortion}, and \texttt{severity\_level}. The reference image is used only to help the annotator identify the visual effect introduced by the distortion, while the metadata constrains the generated description to the intended distortion category and severity level.

The annotator is instructed to generate a three-sentence diagnostic report. The first sentence describes the visible characteristics of the \texttt{L3\_distortion} under the given severity level and its effect on human visual perception. The second sentence explains how the distortion reduces the interpretability or analytical value of the remote sensing image, with concrete visual consequences such as blurred road boundaries, lost texture, or corrupted object structures. The third sentence summarizes the overall image quality. Although the reference image is used during annotation, the final description is required to describe only the distorted image and must not mention, refer to, or imply the existence of a reference image.

After generation, we further check the descriptions to remove unsupported speculation, incorrect distortion descriptions, and vague quality assessments. This human-LVLM collaborative process enables SenseDescription to provide detailed diagnostic labels while maintaining consistency across different distortion categories and severity levels.

\section{Evaluation Details}
\label{evaluation_details}
\begin{table*}[t]
\caption{Details of the evaluated open-source vision-language models.}
\label{tab:vlm_info}
\centering
\small
\renewcommand{\arraystretch}{1.1}
\setlength{\tabcolsep}{5pt}

\begin{tabular}{lccc}
\Xhline{1.5pt}
\textbf{Model} &
\textbf{Vision Backbone} &
\textbf{Language Backbone} &
\textbf{Overall Parameters} \\
\Xhline{1.5pt}

\multicolumn{4}{c}{\textbf{General-domain Large Vision-Language Models}} \\
\hline

Ovis1.6-9B & SigLIP-400M & Gemma2-9B-It & 9B \\
Ovis2.5-9B & SigLIP2-400M & Qwen2.5-7B-Instruct & 9B \\
Llama-3.2-11B & -- & Llama 3.1 & 10.6B \\
Molmo-7B-D & CLIP ViT-L/14 & Qwen2-7B & 7B \\
Phi4-vision & SigLIP2-400M & Phi-4 & 14B \\
SenseNova-SI-1.1-8B & CLIP ViT-L/14 & Qwen3-8B & 8B \\
DeepSeekVL-7B & SigLIP-400M & DeepSeek-7B & 7.3B \\
LLaVA1.6-7B & CLIP ViT-L/14 & Vicuna-v1.5-7B & 7.1B \\
LLaVA1.6-13B & CLIP ViT-L/14 & Vicuna-v1.5-13B & 13.4B \\
LLaVA2-ov-7B & SigLIP-400M & Qwen2-7B & 8B \\
InternVL3.5-8B & InternViT-300M & Qwen3-8B & 8B \\
InternVL3.5-14B & InternViT-300M & Qwen3-14B & 14B \\
InternVL3.5-38B & InternViT-6B & Qwen3-32B & 38B \\
InternVL3.5-30B-A3B & InternViT-300M & Qwen3-30B-A3B & 30B/A3B \\
Qwen3VL-8B & SigLIP2-400M & Qwen3-8B & 8B \\
Qwen3VL-32B & SigLIP2-400M & Qwen3-32B & 32B \\
Qwen3VL-30B-A3B & SigLIP2-400M & Qwen3-30B-A3B & 30B/A3B \\
KimiVL-A3B & ViT (MoonViT) & Moonshot-MoE-A3B & 16B/A3B \\

\hline
\multicolumn{4}{c}{\textbf{Remote Sensing Large Vision-Language Models}} \\
\hline

GeoChat & CLIP ViT-L/14 & Vicuna-v1.5-7B & 7B \\
TEOChat & CLIP ViT-L/14 & LLaMA2-7B & 7B \\
EarthDial & InternViT-300M & Phi-3-mini & 7B \\
LHRS-Bot & CLIP ViT-L/14 & LLaMA2-7B & 7B \\
LHRS-Bot-nova & SigLIP-L/14 & LLaMA3-8B & 8B \\

\Xhline{1.5pt}
\end{tabular}
\vspace{-4mm}
\end{table*}
\subsection{Evaluation Settings}
Table \ref{tab:vlm_info} provides details of all open-source VLMs evaluated in our work.

\paragraph{Implementation Details.}
All general-domain open-source VLMs are deployed with the ms-swift framework\footnote{\url{https://github.com/modelscope/ms-swift}} from ModelScope, using vLLM\footnote{\url{https://github.com/vllm-project/vllm}} as the inference backend whenever supported. Remote sensing VLMs and IQA-oriented VLMs are evaluated using their official open-source implementations and are integrated into the same evaluation pipeline to ensure consistent data loading, prompting, answer parsing, and metric computation. Closed-source models are evaluated through their official APIs with the same zero-shot prompts and deterministic decoding settings whenever exposed by the API.

\paragraph{Generation Configuration.}
To ensure consistency and reproducibility, we use a unified generation configuration across models whenever supported: \texttt{temperature=0.0}, \texttt{max\_new\_tokens=128}, \texttt{truncation\_strategy=right}, and \texttt{enable\_thinking=False}. For models or APIs that do not expose a specific option, we keep the official default setting and record the corresponding backend configuration. Each sample is evaluated once. For \textit{SensePerception}, model outputs are parsed by the deterministic extraction pipeline described in Appendix~\ref{sec: EvaluationProtocolSensePerception}; invalid, empty, or unparseable responses are counted as incorrect. For \textit{SenseDescription}, generated reports are scored by the LLM-as-a-Judge protocol described in Appendix~\ref{sec: EvaluationProtocolSenseDescription}.

\paragraph{Input Formatting.}
For single-image tasks, each model receives one remote sensing image together with the corresponding zero-shot prompt. For paired-image tasks, models with native multi-image support receive the two images as separate visual inputs in a fixed \textit{Image~1}/\textit{Image~2} order. Models without a compatible multi-image inference interface are not evaluated under paired-image settings, which are marked as ``--'' in the result tables. All prompts explicitly refer to \textit{Image~1} and \textit{Image~2} for paired-image questions to avoid ambiguity.

\paragraph{Computational Cost.}
All open-source inference experiments are conducted on up to eight NVIDIA RTX 3090 GPUs, with the number of devices allocated according to the parameter scale of each VLM. Each complete model evaluation takes approximately 1.5 wall-clock hours on its allocated devices, resulting in about 34.5 hours of total wall-clock inference time for all evaluated open-source models. The total GPU-hours scale with the number of allocated devices. No additional large-scale training or unreported compute-intensive experiments are conducted.

\subsection{Evaluation Protocol for \textit{SensePerception}}
\label{sec: EvaluationProtocolSensePerception}
The \textit{SensePerception} task is formulated as a MCQ  answering problem, where each sample consists of one or more remote sensing images, a perception-oriented question, and a set of candidate options with exactly one correct answer. This formulation enables an objective and reproducible evaluation across heterogeneous VLMs, while avoiding the ambiguity inherent in open-ended generation.

\paragraph{Prompt Design.} To eliminate prompt-induced bias across models, we adopt a unified instruction template for all evaluated VLMs. The template is composed of three components arranged in a fixed order: the question, the image token(s), and the list of candidate options. For multi-image samples, all image tokens are inserted sequentially at the same position, while the rest of the template remains unchanged. In the examples below, we use {\color{PromptGray}gray} to denote the format-constraining prompt, black for the question itself, and within the candidate options, {\color{CorrectGreen}green} indicates the correct answer(s) while {\color{WrongOrange}orange} indicates the incorrect ones. These colors are used only for illustration in the manuscript and are not included in the actual model input. Two representative examples are shown below:

\textit{{\small 
{{\tt{User Prompt:}}\\
\color{PromptGray}Please select the two correct answers for the following question from the given options. Respond only with the two letters in alphabetical order, separated by a comma with no space (e.g., A,B). Do not include any additional text.}
Select all types of distortions affecting this image. [IMAGE\_TOKEN] \\ 
{{\tt{Answer:}}\\
\color{WrongOrange}A. Missing tiles}\qquad
{\color{CorrectGreen}B. Motion blur}\qquad
{\color{WrongOrange}C. Gaussian noise}\qquad
{\color{CorrectGreen}D. Cloud}
} } 

\textit{{\small 
{{\tt{User Prompt:}}\\
\color{PromptGray}Please select the most appropriate answer for the following single-choice question from the given options. Only respond with the corresponding letter (A, B, C or D). Do not include any additional text.}
Based on the paired images, what is the dominant degradation in Image 2? [IMAGE1\_TOKEN] [IMAGE2\_TOKEN] \\ 
{{\tt{Answer:}}\\
\color{WrongOrange}A. Color distortion}\qquad
{\color{WrongOrange}B. Image stretching}\qquad
{\color{WrongOrange}C. Dead pixel}\qquad
{\color{CorrectGreen}D. Haze}
}}

\paragraph{Answer Parsing.}
The trailing instruction constrains the model to respond only with option letter(s), reducing output-format variance and enabling deterministic parsing. To accommodate diverse model outputs, both the ground-truth answer and the model response are normalized into sets of option letters, and a prediction is considered correct only when the two sets are identical. We first apply case-insensitive regular-expression matching to extract valid option letters from the response, tolerating common variants such as ``A.'', ``A)'', or ``The answer is A''. If no valid letter is detected, we perform a deterministic content-level match between the normalized response and the textual content of each candidate option, and map the matched option back to its corresponding letter. Responses that fail both stages are counted as incorrect. Representative parsing cases are shown in Table~\ref{tab:mcq_parsing_examples}.
\begin{table}[!ht]
\centering
\small
\setlength{\tabcolsep}{5pt}
\renewcommand{\arraystretch}{1.06}
\caption{Representative examples of robust option extraction for MCQ evaluation. Minor formatting variations in model responses do not affect answer parsing.}
\label{tab:mcq_parsing_examples}
\begin{tabular}{c@{\hspace{10pt}}p{0.50\linewidth}@{\hspace{12pt}}c@{\hspace{10pt}}p{0.22\linewidth}}
\toprule
\textbf{GT} & \textbf{Model Response} & \textbf{Parsed} & \textbf{Case} \\
\midrule
A & \texttt{A} & A & Standard \\
B & \texttt{The answer is B.} & B & Natural language \\
C & \texttt{answer: C} & C & Colon format \\
D & \texttt{I think it is D.} & D & Informal answer \\
A & \texttt{(A) The image shows visible blur.} & A & Parenthesized option \\
B & \texttt{B. Moderate distortion.} & B & With explanation \\
C & \texttt{My choice is option C because the image contains compression artifacts.} & C & With rationale \\
B & \texttt{b} & B & Lowercase option \\
A & \texttt{The image has severe Gaussian blur.} & A & Text fallback \\
\bottomrule
\end{tabular}
\vspace{-4mm}
\end{table}

\paragraph{Stratification Protocol.}
We adopt exact-match accuracy as the primary metric. A prediction is considered correct only when the parsed option-letter set is identical to the ground-truth option set. Beyond the overall average accuracy, we report task-specific stratified results for both single-image and paired-image evaluations. For the \textit{single-image evaluation}, degradation types are grouped into \textit{General} and \textit{RS-centric} categories according to their domain specificity, as summarized in Table~\ref{tab:distortion_split}. Samples are further divided into single-distortion and multi-distortion cases, depending on whether one or multiple degradation types are present in the input image. These results characterize model performance across different degradation domains and distortion compositions. For the \textit{paired-image evaluation}, we adopt the same \textit{General} and \textit{RS-centric} degradation grouping. Paired-image samples are further categorized into \textit{intra-image} and \textit{inter-temporal} settings, corresponding to comparisons within the same source image and across different acquisition times, respectively. These results characterize the pairwise discriminative ability of VLMs under domain-specific and temporal/contextual variations.

\subsection{Evaluation Protocol for \textit{SenseDescription}}
\label{sec: EvaluationProtocolSenseDescription}
The \textit{SenseDescription} task is formulated as an open-ended caption generation problem, where each sample consists of one or more remote sensing images and a description-oriented instruction, and the model is required to produce a free-form natural language description of the visual content. Unlike the MCQ formulation adopted in \textit{SensePerception}, this task evaluates a model's ability to articulate fine-grained low-level degradation characteristics, their visual manifestations, and their impact on image interpretability in fluent natural language.

\paragraph{Prompt Design.} To eliminate prompt-induced bias across models, we adopt a unified instruction template for all evaluated VLMs. The template is composed of two components arranged in a fixed order: the description instruction and the image tokens. For multi-image samples, all image tokens are inserted sequentially at the same position, while the rest of the template remains unchanged. {\color{CorrectGreen}green} indicates the reference label description. Each dimension is evaluated independently using a dedicated prompt. Two representative example is shown below:

\textit{{\small 
{\tt{User Prompt:}}\\
You are shown a single distorted remote sensing image. Write exactly three concise sentences that describe the scene and its quality:\\
1. Identify the dominant type of distortion (e.g., haze, motion blur, compression artifacts, sensor noise) and describe its visible characteristics.\\
2. Explain how this distortion impairs interpretability or analytical utility, citing concrete visual consequences.\\
3. Provide an overall assessment of the image quality.Your response must contain exactly three sentences, with no numbering, and must not speculate about how the image 'should' look.
[IMAGE1\_TOKEN]\\
{\tt{Answer:}}\\
{\color{CorrectGreen}Moderate gaussian blur softens the edges of buildings and roads, making structural details less distinct. The blurring reduces interpretability by obscuring fine textures and diminishing clarity in densely packed urban areas. The image quality is moderately degraded, affecting its usefulness for detailed analysis.}
} } 

\textit{{\small 
{\tt{User Prompt:}}\\
You will see two remote sensing images Image 1 and Image 2. Respond with a single paragraph containing exactly three concise sentences. Sentence one must describe the observable distortions in Image 1 and how they appear visually. Sentence two must describe the observable distortions in Image 2 and how they appear visually . Sentence three must determine which image has higher overall quality (or if they are equal) and briefly explain the reason.
[IMAGE1\_TOKEN] [IMAGE2\_TOKEN] \\
{\tt{Answer:}}\\
{\color{CorrectGreen}Image 1 has no observable distortions and appears visually clean and sharp. Image 2 contains severe Gaussian blur artifacts, making the details appear smudged and less distinct. Therefore, Image 1 has higher overall quality due to its clarity and lack of distortions.}
} } 

\begin{table*}[!t]
\centering
\caption{Split of distortion types in SenseBench. Distortions are divided into General distortions and RS-centric distortions for fine-grained result analysis.}
\label{tab:distortion_split}

\setlength{\tabcolsep}{3.0pt}
\renewcommand{\arraystretch}{1.12}
\small

\begin{tabularx}{\textwidth}{
>{\raggedright\arraybackslash}p{0.11\textwidth}
>{\hsize=0.85\hsize\raggedright\arraybackslash}X
>{\raggedright\arraybackslash}p{0.115\textwidth}
>{\hsize=1.15\hsize\raggedright\arraybackslash}X
}
\toprule
\multicolumn{2}{c}{\textbf{General Distortions}} &
\multicolumn{2}{c}{\textbf{RS-centric Distortions}} \\
\cmidrule(lr){1-2} \cmidrule(lr){3-4}
\textbf{Category} & \textbf{Distortion Types} &
\textbf{Category} & \textbf{Distortion Types} \\
\midrule

\textit{Blur}
& Gaussian blur; Motion blur
& \textit{Cloud}
& Cloud \\

\addlinespace[1.5pt]

\textit{Noise}
& Gaussian noise; Impulse noise; \newline
Spatially correlated noise
& \textit{Noise}
& Deadline noise; Stripe noise; Sidelobe noise; Speckle noise \\

\addlinespace[1.5pt]

\textit{Compression}
& JPEG; JPEG2000; WebP; \newline
SPIHT
& \textit{Correction}
& Color band attenuation; Color band switch; \newline
Geometric compression; Geometric stretching \\

\addlinespace[1.5pt]

\textit{Cloud}
& Haze
& \textit{Missing}
& Blind flickering; Linear dead pixels; Missing tiles \\

\bottomrule
\end{tabularx}
\vspace{-3mm}
\end{table*}

\paragraph{Evaluation Rubric.}
Following the \textit{LLM-as-a-Judge} evaluation protocol described in the main text, each generated diagnostic report is evaluated against the corresponding reference description along three dimensions: (1) \textit{Completeness}, which rewards the coverage of key diagnostic elements in the reference; (2) \textit{Correctness}, which measures semantic consistency with the reference in terms of distortion type, severity, visual effects, and quality judgment; and (3) \textit{Faithfulness}, which penalizes hallucinated objects, unsupported assumptions, and speculative explanations beyond the observable image content. Each dimension is scored in $\{0,1,2\}$, corresponding to failure, partial success, and full success, respectively. The prompts used for the LLM-as-a-Judge evaluator are listed below:

\textit{{\small \#System: You are a helpful assistant.}}

\textbf{Completeness.} 
\textit{{\small \#User: Evaluate whether the description [RESPONSE] fully covers the low-level degradation factors mentioned in the reference description [LABEL]. Focus specifically on degradation factors such as blur, noise, compression artifacts, exposure issues, color distortion, etc.
Rate the completeness as follows:\\
- Score 2: The description includes all or almost all degradation factors present in the reference.\\
- Score 1: The description misses some key degradation factors but captures others.\\
- Score 0: The description misses most or all of the key degradation factors mentioned in the reference.\\
Provide only the score, without additional explanations or formatting (e.g., 0, 1, or 2).}} \\

\textbf{Correctness.} 
\textit{{\small \#User: Evaluate whether the description [RESPONSE] is factually aligned with the reference description [LABEL] and free from contradictions regarding visual quality. The correctness metric strictly penalizes contradictory descriptions (e.g., saying "sharp" when the reference says "blur", or "clean" when the reference says "noisy").\\
Rate the correctness as follows:\\
- Score 2: The description is factually aligned with the reference and contains no contradictions.\\
- Score 1: The description contains minor contradictions or ambiguities regarding the visual attributes.\\
- Score 0: The description contains strong or direct contradictions to the reference (e.g., opposite attributes).\\
Provide only the score, without additional explanations or formatting (e.g., 0, 1, or 2).}} \\

\textbf{Faithfulness.} 
\textit{{\small \#User: Evaluate whether the description [RESPONSE] is faithful to the low-level visual quality, avoiding hallucinations and irrelevant high-level semantics. Reference description: [LABEL]. The faithfulness metric verifies that the generated claims are visually grounded.\\
- It strictly penalizes "hallucinations" (mentioning artifacts or flaws not present in the reference).\\
- It penalizes irrelevant high-level semantics (describing the image content/scene rather than its quality/degradation).\\
Rate the faithfulness as follows:\\
- Score 2: The description is strictly grounded in low-level visual quality, with no hallucinations or irrelevant high-level content descriptions.\\
- Score 1: The description contains minor hallucinations or mixes in some irrelevant high-level semantics.\\
- Score 0: The description is dominated by hallucinations or focuses entirely on high-level content (e.g., describing objects) instead of visual quality.\\
Provide only the score, without additional explanations or formatting (e.g., 0, 1, or 2).}} \\

\section{Quality Control}
\label{sec:quality_control}
\subsection{Quality Control Procedure}

\begin{figure*}[!t]
  \centering
  \resizebox{\linewidth}{!}{
   \includegraphics{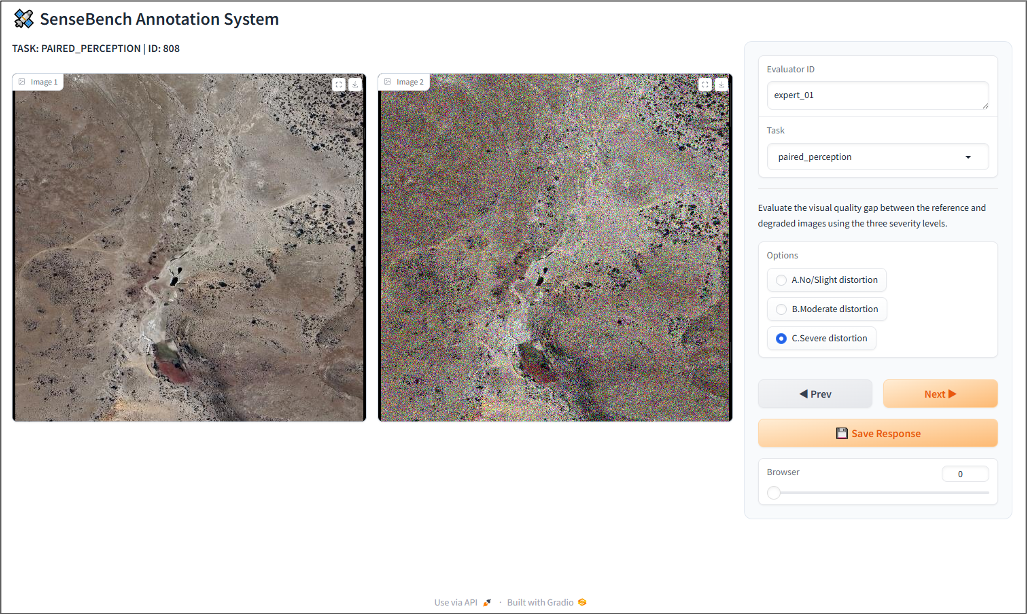}}
   \caption{SenseBench Annotation System used for human evaluator baseline collection and evaluation quality control.}
   \label{fig: Annotation}
   \vspace{-5mm}
\end{figure*}

To improve the reliability of \textbf{SenseBench}, we implemented a multi-stage quality control procedure during benchmark construction. Twelve evaluators with backgrounds in remote sensing or computer vision participated in the annotation and verification process. Following a team-based inspection protocol, the evaluators were divided into four groups, denoted as Group A, B, C, and D, with three members in each group. In each inspection stage, a sample was approved only when all three members of the assigned group confirmed its visual validity, task relevance, and internal consistency. Samples with any identified issue were returned for revision and further review, while unresolved cases were removed from the benchmark.

\paragraph{Perceptual Validity and Metadata Alignment.}
Since most degraded samples in \textbf{SenseBench} are generated through \textit{controlled synthesis}, \textbf{Group A} and \textbf{B} verified the perceptual validity of the synthesized distortions and their alignment with the associated metadata. Evaluators checked whether each sample contained clear and recognizable visual evidence consistent with the intended degradation mechanism, such as edge softening for blur, block or coding artifacts for compression, contrast attenuation for haze, directional structures for stripe noise, and missing regions for data loss. They further verified whether the L2 distortion category, L3 fine-grained distortion type, and severity label recorded in the information file were consistent with the observed degradation. Samples with imperceptible, excessive, or visually inconsistent distortions, as well as samples with mismatched metadata or unstable severity judgments, were corrected, re-synthesized, or removed.

\paragraph{\textit{SensePerception} Verification.}
For \textit{SensePerception}, \textbf{Group C} inspected the generated MCQ samples from both visual and semantic perspectives. A sample was approved only when the distorted image, metadata, question, candidate options, and ground-truth answer were mutually consistent. The \textit{Whether}, \textit{What}, and \textit{How} questions were respectively checked for degradation presence, distortion-type correctness, and severity-level consistency. Samples with incorrect labels, ambiguous distractors, or mismatched question-answer pairs were returned for revision and further review; unresolved cases were removed.

\paragraph{\textit{SenseDescription} Verification.}
For \textit{SenseDescription}, \textbf{Group D} manually reviewed the LVLM-assisted diagnostic annotations to ensure their factuality, specificity, and task alignment. A description was approved only when it correctly identified the dominant distortion, provided concrete visual evidence, explained its impact on interpretability or analytical utility, and gave a consistent overall quality assessment. Descriptions with incorrect distortion claims, vague or unsupported assessments, hallucinated content, or reference-image leakage were revised, regenerated, or removed.

\subsection{Human Evaluation}
\label{human_evaluator_baseline}

\paragraph{Human Evaluator Baseline.}
To provide a human reference baseline, we conducted human evaluation with annotators from \textbf{Group A} and \textbf{Group B}. For each evaluation task, we constructed a distortion-stratified human-evaluation subset by sampling 5\% of benchmark instances from each distortion category, with category-level sample counts rounded up when necessary to ensure coverage of all distortion categories. Each sampled instance was independently completed by every annotator, so each question received six human responses. The human evaluation followed the same protocols described in Sec.~\ref{evaluation_details}, and the evaluation interface is illustrated in Fig.~\ref{fig: Annotation}. The HUMAN EVALUATOR rows in Tables~\ref{perceptionsingle}, \ref{perceptionmulti}, and~\ref{tab:description_all} report the mean performance averaged over all annotators and sampled instances.

\subsection{Evaluation Quality Control}
\label{evaluation_quality_control}
\begin{table*}[!t]
\centering
\caption{Results on the LLM-As-Judge Reliability on Assessing Description Ability of MLLMs.}
\label{tab:llm_as_judge_quality}
\footnotesize
\setlength{\tabcolsep}{5pt}
\renewcommand{\arraystretch}{1}
\setlength{\dashlinedash}{3pt}
\setlength{\dashlinegap}{2pt}

\begin{tabular}{
@{\hspace{3pt}}l
!{\hspace{6pt}\vrule width 0.4pt\hspace{6pt}}
cccc
!{\hspace{6pt}\vrule width 0.4pt\hspace{6pt}}
c
@{\hspace{3pt}}
}
\toprule
\textbf{Dimensions/Model} &
\textbf{Completeness} &
\textbf{Correctness} &
\textbf{Faithfulness} &
\textbf{Average} &
\textbf{Rank} \\
\midrule

\multicolumn{6}{@{\hspace{2pt}}l}{\emph{Reliability performance for single image description}} \\
\hdashline
Qwen3-VL-8B
& 0.930/0.945 & 0.911/0.979 & 0.945/0.998 & 0.929/0.974 & 1 / 1 \\

InternVL3.5-8B
& 0.903/0.951 & 0.913/0.984 & 0.915/0.992 & 0.910/0.976 & 2 / 2 \\

InternVL3.5-30B-A3B
& 0.919/0.944 & 0.927/0.992 & 0.949/0.995 & 0.932/0.977 & 4 / 4 \\

DepictQA
& 0.901/0.924 & 0.922/0.981 & 0.935/0.948 & 0.919/0.951 & 3 / 3 \\

TEOChat
& 0.952/0.991 & 0.933/0.992 & 0.928/0.955 & 0.938/0.979 & 5 / 5 \\

\midrule
\multicolumn{6}{@{\hspace{2pt}}l}{\emph{Reliability performance for paired image description}} \\
\hdashline

Qwen3-VL-8B
& 0.908/0.945 & 0.931/0.993 & 0.931/0.985 & 0.923/0.974 & 1 / 1 \\

InternVL3.5-8B
& 0.900/0.954 & 0.922/0.994 & 0.940/0.968 & 0.921/0.972 & 4 / 4 \\

InternVL3.5-30B-A3B
& 0.903/0.941 & 0.932/0.994 & 0.940/0.958 & 0.925/0.964 & 3 / 3 \\

DepictQA
& 0.925/0.946 & 0.928/0.995 & 0.912/0.974 & 0.922/0.972 & 2 / 2 \\

TEOChat
& 0.913/0.948 & 0.982/0.999 & 0.933/0.984 & 0.943/0.977 & 5 / 5 \\

\bottomrule
\end{tabular}
\vspace{-3mm}
\end{table*}

\paragraph{Reliability of \textit{LLM-as-judge} Evaluation.}
To verify the reliability of \textit{LLM-as-judge} for description-quality evaluation, we selected five representative MLLMs and evaluated them under both single-image and paired-image settings. Human evaluation was conducted by \textbf{Group A} and \textbf{Group C}, each consisting of three annotators, and the average score within each group was used as the corresponding human score. The two groups performed the evaluation independently, and their scores were used to assess the consistency between human judgments and \textit{LLM-as-judge} scores. We then measured this consistency across the evaluation dimensions using SRCC/PLCC. As shown in Table~\ref{tab:llm_as_judge_quality}, the results demonstrate strong agreement between \textit{LLM-as-judge} scores and human judgments.

\subsection{Information of the Human Annotators}
\label{sec:human_annotators}

We enlisted twelve human annotators with academic backgrounds in remote sensing, computer vision, and electronic information engineering to participate in the annotation, verification, and quality control of \textbf{SenseBench}. Following the team-based protocol described in Sec.~\ref{sec:quality_control}, the annotators were organized into four groups, denoted as Groups A, B, C, and D, with three annotators in each group. The same grouping was used consistently across the benchmark construction and evaluation-quality-control procedures.

\paragraph{Education and Research Background.}
Among the twelve annotators, 2 were at the bachelor's level, 7 were at the master's level, and 3 were at the doctoral level. In terms of research background, 7 annotators specialized in photogrammetry and remote sensing, 3 specialized in computer science or related areas, and 2 specialized in electronic information engineering. These backgrounds ensured that the annotators were familiar with remote sensing image interpretation, visual degradation analysis, and vision-related annotation tasks.

\paragraph{Age Distribution.}
The annotators were between 21 and 32 years old. Specifically, 6 annotators were aged 21--25, 5 were aged 26--29, and 1 was aged 30--32.

\paragraph{Ethical Considerations.}
All annotators were informed of the purpose and intended use of \textbf{SenseBench} before participating in the annotation and quality-control process. Their participation was voluntary and appropriately compensated. Specifically, each annotator received an hourly compensation of 60 RMB for part-time annotation and inspection work, which is consistent with the compensation level for comparable academic assistance tasks in our local context. The annotators were instructed to focus on low-level visual quality factors in remote sensing images and to flag samples with ambiguous labels, unsupported descriptions, or other annotation-quality issues for revision or removal.

\section{Additional Results}
\label{sec:additional_results}

\subsection{Complete SensePerception Results}

To further expand the scale of our evaluation and the coverage of tested models, we conduct a comprehensive analysis on a broader set of VLMs. 
This section reports the complete \textit{SensePerception} results and provides additional analysis of model behavior under different input settings and evaluation dimensions. 
In contrast to the averaged results presented in Sec.~\ref{04_experiment}, Table~\ref{tab: perceptionsingle_all} and Table~\ref{tab: perceptionmulti_all} offer a fine-grained breakdown across question types, distortion categories, and input conditions. 

\begin{table*}[!t]
    
    \centering
    \small 
    \renewcommand\arraystretch{1} 
    \renewcommand\tabcolsep{4.5pt}
    
    \caption{Complete quantitative results of VLMs on low-level perception for single images.}
    \label{tab: perceptionsingle_all}
    \newcolumntype{C}{>{\centering\arraybackslash}p{1.3cm}}
    \resizebox{\textwidth}{!}{%
    \begin{tabular}{l | CCC | CC | CC | C}
    \toprule
    
    \textbf{Sub-categories} & \multicolumn{3}{c|}{\textbf{Question Type}} & \multicolumn{2}{c|}{\textbf{Domain}} & \multicolumn{2}{c|}{\textbf{Context}} & \multirow{2}{*}{\textit{Average$\uparrow$}} \\ 
    
    \cdashline{1-8} 
    
    \rule{0pt}{12pt}\textbf{Model} & \textit{Whether$\uparrow$} & \textit{What$\uparrow$} & \textit{How$\uparrow$} & \textit{General$\uparrow$} & \textit{RS$\uparrow$} & \textit{Single$\uparrow$} & \textit{Multi$\uparrow$} & \\ 
    
    \midrule
    \textcolor{gray}{\textsc{Human Evaluator}} & \textcolor{gray}{90.58\%} & \textcolor{gray}{88.62\%} & \textcolor{gray}{90.26\%} & \textcolor{gray}{90.11\%} & \textcolor{gray}{89.32\%} & \textcolor{gray}{89.75\%} & \textcolor{gray}{90.00\%} & \textcolor{gray}{89.82\%}\\
    \noalign{\vskip 1pt}
    \cdashline{1-9}
    \noalign{\vskip 1pt}

    GPT-5.4 & 68.46\% & 60.43\% & 58.54\% & 64.83\% & \underline{62.87\%} & 64.23\% & 41.50\% & 62.48\% \\
    Qwen3.5-Plus & \underline{71.32}\% & 60.62\% & 57.92\% & 68.10\% & 60.39\% & 64.87\% & 42.67\% & 63.29\% \\
    Gemini-3.1-pro-preview & \textbf{87.17\%} & \textbf{80.17\%} & \underline{59.92\%} & \textbf{79.47\%} & \textbf{80.06\%} & \textbf{79.68\%} & 41.17\% & \textbf{75.75\%} \\

    \midrule
    Ovis1.6-9B\cite{ovis1.6} & 68.54\% & 60.88\% & 50.38\% & 65.43\% & 57.80\% & 62.36\% & 54.17\% & 59.93\% \\
    Ovis2.5-9B\cite{ovis2.5} & 50.39\% & 41.48\% & 37.31\% & 46.33\% & 41.48\% & 43.87\% & 29.50\% & 43.06\% \\
    Llama3.2-11B\cite{llama3.2} & 54.17\% & 42.21\% & 38.62\% & 49.70\% & 42.02\% & 45.57\% & 46.67\% & 45.00\% \\
    Molmo-7B-D\cite{molmo} & 49.30\% & 41.59\% & 53.15\% & 52.57\% & 41.17\% & 46.87\% & 46.50\% & 48.01\% \\
    Phi4-vision\cite{phi4} & 51.75\% & 37.33\% & 38.38\% & 43.97\% & 42.19\% & 42.42\% & 44.83\% & 42.49\% \\
    SenseNova-SI-1.1-8B\cite{sensenova} & 58.94\% & 50.72\% & 33.77\% & 48.03\% & 53.74\% & 50.36\% & 34.67\% & 47.81\% \\
    DeepseekVL-7B\cite{deepseekvl} & 60.31\% & 38.48\% & 33.46\% & 47.50\% & 44.98\% & 45.17\% & 33.83\% & 44.08\% \\
    LLaVA1.6-7B\cite{llava} & 50.90\% & 37.06\% & 49.54\% & 49.30\% & 40.63\% & 44.85\% & 45.50\% & 45.83\% \\
    LLaVA1.6-13B\cite{llava} & 50.04\% & 38.53\% & 35.08\% & 45.50\% & 38.26\% & 41.32\% & 47.50\% & 41.22\% \\
    LLaVA2-ov-7B\cite{llava-ov} & 70.58\% & \underline{67.57\%} & 54.31\% & \underline{71.07\%} & 61.46\% & \underline{67.26\%} & 46.00\% & 64.15\% \\
    InternVL3.5-8B\cite{internvl3.5} & 56.59\% & 40.70\% & 42.38\% & 48.89\% & 44.52\% & 47.19\% & 46.83\% & 46.56\% \\
    InternVL3.5-14B\cite{internvl3.5} & 53.14\% & 41.77\% & 42.23\% & 46.47\% & 44.64\% & 46.47\% & 39.67\% & 45.71\% \\
    InternVL3.5-38B\cite{internvl3.5} & 57.70\% & 50.27\% & 25.46\% & 49.92\% & 50.52\% & 46.43\% & 36.67\% & 44.48\% \\
    InternVL3.5-30B-A3B\cite{internvl3.5} & 58.99\% & 47.96\% & 47.77\% & 53.30\% & 51.61\% & 52.26\% & 44.17\% & 51.57\% \\
    Qwen3VL-8B\cite{qwen3vl} & 64.35\% & 54.20\% & 52.38\% & 59.80\% & 53.29\% & 58.02\% & 50.33\% & 56.98\% \\
    Qwen3VL-32B\cite{qwen3vl} & 69.94\% & 63.65\% & \textbf{60.15\%} & 70.06\% & 56.40\% & 65.92\% & \underline{58.33\%} & \underline{64.58\%} \\
    Qwen3VL-30B-A3B\cite{qwen3vl} & 65.66\% & 54.75\% & 50.46\% & 60.93\% & 54.83\% & 57.13\% & \textbf{58.50\%} & 56.96\% \\
    KimiVL-A3B\cite{kimivl} & 66.11\% & 51.73\% & 49.31\% & 58.82\% & 55.89\% & 56.85\% & 47.17\% & 55.72\% \\
    
    \midrule 
    DepictQA\cite{depictqa} & 50.06\% & 31.49\% & 21.00\% & 39.00\% & 33.20\% & 35.87\% & 41.00\% & 34.18\% \\
    Q-instruct-llava-7b\cite{qinstruct} & 47.17\% & 28.11\% & 34.31\% & 37.40\% & 36.31\% & 36.57\% & 37.50\% & 36.53\% \\
    Q-instruct-llava-13b\cite{qinstruct} & 53.95\% & 36.41\% & 48.69\% & 45.37\% & 46.98\% & 46.11\% & 40.83\% & 46.35\% \\
    
    \midrule 
    GeoChat\cite{geochat} & 46.13\% & 29.80\% & 20.92\% & 34.60\% & 33.83\% & 33.57\% & 33.50\% & 32.28\% \\
    TEOChat\cite{teochat} & 54.57\% & 29.99\% & 27.38\% & 37.70\% & 41.52\% & 38.51\% & 24.50\% & 37.31\% \\
    EarthDial\cite{soni2025earthdial} & 58.96\% & 48.06\% & 30.31\% & 46.23\% & 52.87\% & 48.89\% & 20.50\% & 45.78\% \\
    LHRS-Bot\cite{lhrs} & 54.96\% & 27.01\% & 26.15\% & 35.47\% & 40.56\% & 37.40\% & 33.17\% & 36.04\% \\
    LHRS-Bot-nova\cite{li2025lhrsbotnava} & 47.83\% & 23.54\% & 21.54\% & 30.83\% & 33.85\% & 30.85\% & 41.33\% & 30.97\% \\
    \bottomrule
    \end{tabular}
    }
    \vspace{-4mm}
\end{table*}

\paragraph{Question-type difficulty reveals a perception hierarchy.}
As defined in Sec.~\ref{sec: SensePerception}, \textit{SensePerception} evaluates distortion existence, distortion category, and distortion severity through \textit{Whether}, \textit{What}, and \textit{How} questions, respectively. The complete results show a consistent performance decrease from \textit{Whether} to \textit{What} and then to \textit{How}, suggesting that current VLMs are generally more capable of detecting whether a visible degradation exists than identifying its category or estimating its severity. For example, Table~\ref{tab: perceptionsingle_all} shows that Gemini-3.1-pro-preview drops from 87.17\% on \textit{Whether} to 80.17\% on \textit{What} and 59.92\% on \textit{How} in the single-image setting, while Table~\ref{tab: perceptionmulti_all} shows a similar pattern for GPT-5.4 in the paired-image setting, decreasing from 85.64\% and 80.45\% on \textit{Whether} and \textit{What} to 62.36\% on \textit{How}.This trend indicates that low-level remote-sensing perception is not a single ability, but a hierarchy from anomaly detection to degradation diagnosis and severity calibration. While \textit{Whether} questions mainly rely on salient visual abnormalities, \textit{What} questions require distinguishing visually similar degradation patterns, and \textit{How} questions further require a stable quality scale, making severity estimation the most challenging and exposing the limitations of current VLMs in fine-grained visual calibration.

\begin{table*}[!t]
    \centering
    \small 
    \renewcommand\arraystretch{1} 
    \renewcommand\tabcolsep{4.5pt}
    
    \caption{Complete quantitative results of VLMs on low-level perception for paired images.}
    \label{tab: perceptionmulti_all}
    \newcolumntype{C}{>{\centering\arraybackslash}p{1.3cm}}
    
    \resizebox{\textwidth}{!}{%
    \begin{tabular}{l | CCC | CC | CC | C}
    \toprule
    
    \textbf{Sub-categories} & \multicolumn{3}{c|}{\textbf{Question Type}} & \multicolumn{2}{c|}{\textbf{Domain}} & \multicolumn{2}{c|}{\textbf{Context}} & \multirow{2}{*}{\textit{Average$\uparrow$}} \\ 
    
    \cdashline{1-8} 

    \rule{0pt}{12pt}\textbf{Model} & \textit{Whether$\uparrow$} & \textit{What$\uparrow$} & \textit{How$\uparrow$} & \textit{General$\uparrow$} & \textit{RS$\uparrow$} & \textit{Intra.$\uparrow$} & \textit{Inter.$\uparrow$} & \\ 

    \midrule
    \textcolor{gray}{\textsc{Human Evaluator}} & \textcolor{gray}{95.28\%} & \textcolor{gray}{89.44\%} & \textcolor{gray}{82.05\%} & \textcolor{gray}{89.81\%} & \textcolor{gray}{89.87\%} & \textcolor{gray}{81.92\%} & \textcolor{gray}{89.62\%} & \textcolor{gray}{88.92\%} \\
    \noalign{\vskip 1pt}
    \cdashline{1-9}
    \noalign{\vskip 1pt}

    GPT-5.4 & 85.64\% & \underline{80.45\%} & 62.36\% & 76.73\% & 75.58\% & \underline{77.50\%} & 78.00\% & 76.15\% \\
    Qwen3.5-PLUS & 87.00\% & 78.00\% & \textbf{65.64}\% & \underline{77.21}\% & \underline{76.55}\% & \underline{77.50}\% & 79.78\% & \underline{76.88\%} \\
    Gemini-3.1-pro-preview & \underline{89.64\%} & \textbf{90.00\%} & \underline{64.21}\% & \textbf{81.94\%} & \textbf{80.63\%} & \textbf{82.55\%} & \textbf{84.33\%} & \textbf{81.28\%} \\

    \midrule
    Ovis1.6-9B\cite{ovis1.6} & 73.55\% & 45.82\% & 30.91\% & 54.12\% & 50.67\% & 54.43\% & 47.80\% & 50.09\% \\
    Ovis2.5-9B\cite{ovis2.5} & 58.36\% & 60.55\% & 36.64\% & 51.52\% & 56.92\% & 51.13\% & 61.13\% & 51.85\% \\
    Llama3.2-11B\cite{llama3.2} & 72.73\% & 32.82\% & 31.33\% & 46.00\% & 49.49\% & 45.42\% & 53.08\% & 45.63\% \\
    Molmo-7B-D\cite{molmo} & 94.82\% & 37.91\% & 29.93\% & 54.55\% & 61.19\% & 53.30\% & 68.43\% & 54.22\% \\
    Phi4-vision\cite{phi4} & 83.73\% & 34.64\% & 36.36\% & 52.24\% & 55.34\% & 52.64\% & 56.10\% & 51.58\% \\
    SenseNova-SI-1.1-8B\cite{sensenova} & 45.45\% & 58.55\% & 54.83\% & 50.55\% & 55.49\% & 49.34\% & 61.64\% & 52.94\% \\
    DeepseekVL-7B\cite{deepseekvl} & 76.73\% & 34.00\% & 33.43\% & 50.85\% & 48.85\% & 49.01\% & 52.58\% & 48.05\% \\
    LLaVA2-ov-7B\cite{llava-ov} & 69.00\% & 52.09\% & 37.62\% & 57.21\% & 51.94\% & 51.70\% & 63.52\% & 52.90\% \\
    InternVL3.5-8B\cite{internvl3.5} & 51.27\% & 57.55\% & 63.92\% & 52.61\% & 62.13\% & 56.37\% & 57.74\% & 57.58\% \\
    InternVL3.5-14B\cite{internvl3.5} & 56.91\% & 61.27\% & 56.64\% & 54.85\% & 63.24\% & 57.92\% & 60.00\% & 58.27\% \\
    InternVL3.5-38B\cite{internvl3.5} & 56.00\% & 62.73\% & 52.73\% & 51.94\% & 65.30\% & 59.62\% & 52.70\% & 57.15\% \\
    InternVL3.5-30B-A3B\cite{internvl3.5} & 68.73\% & 66.64\% & 62.80\% & 63.58\% & 70.28\% & 65.52\% & 69.06\% & 66.06\% \\
    Qwen3VL-8B\cite{qwen3vl} & 72.00\% & 66.55\% & 51.33\% & 63.64\% & 66.48\% & 63.25\% & 69.18\% & 63.29\% \\
    Qwen3VL-32B\cite{qwen3vl} & 79.27\% & 75.82\% & 57.62\% & 70.30\% & 75.73\% & 70.57\% & 78.24\% & 70.90\% \\
    Qwen3VL-30B-A3B\cite{qwen3vl} & \textbf{91.55\%} & 60.55\% & 54.69\% & 69.82\% & 72.09\% & 66.60\% & \underline{82.01\%} & 68.93\% \\
    KimiVL-A3B\cite{kimivl} & 71.45\% & 53.73\% & 49.51\% & 58.18\% & 60.95\% & 54.95\% & 71.19\% & 58.23\% \\

    \midrule 
    DepictQA\cite{depictqa} & 60.18\% & 23.09\% & 23.55\% & 41.82\% & 29.39\% & 37.03\% & 42.44\% & 35.61\% \\
    \midrule
    TEOChat\cite{teochat} & 34.55\% & 30.00\% & 37.06\% & 31.64\% & 35.81\% & 35.33\% & 28.43\% & 33.87\% \\
    EarthDial\cite{soni2025earthdial} & 26.91\% & 50.73\% & 40.28\% & 38.06\% & 40.63\% & 43.92\% & 26.54\% & 39.31\% \\

    \bottomrule
    \end{tabular}
    
    \vspace{-5mm}}
\end{table*}

\paragraph{Model categories reveal mismatched capability biases.}
Using open-source general-domain VLMs as a reference, closed-source commercial models achieve a higher overall upper bound, suggesting that stronger general visual representations and instruction-following ability benefit low-level degradation perception. For example, Table~\ref{tab: perceptionsingle_all} shows that Gemini-3.1-pro-preview achieves an average score of 75.75\%, outperforming most open-source models, while Table~\ref{tab: perceptionmulti_all} shows that the three commercial models remain within a relatively high average range of 76\%--81\% in the paired-image setting. However, this advantage does not fully resolve fine-grained degradation diagnosis. RS-VLMs exhibit a certain remote-sensing bias, as EarthDial improves from 46.23\% on the General subset to 52.87\% on the RS subset in Table~\ref{tab: perceptionsingle_all}, with TEOChat and LHRS-Bot showing similar tendencies, but their overall averages remain low. This indicates that remote-sensing semantic adaptation helps models understand ``what is in the image'', but does not necessarily transfer to identifying ``what degradation is present'' or ``how severe it is''. Similarly, IQA-VLMs do not show a stable advantage despite being trained for image-quality assessment: Table~\ref{tab: perceptionsingle_all} reports only 34.18\%, 36.53\%, and 46.35\% average scores for DepictQA, Q-instruct-llava-7b, and Q-instruct-llava-13b, respectively. These results suggest that natural-image quality priors do not fully generalize to remote-sensing degradation scenarios, and that \textit{SensePerception} requires a joint capability of general visual perception, remote-sensing domain understanding, and fine-grained degradation calibration.

\paragraph{Paired-image context reveals different cross-image attribution behaviors.}
Model performance in the paired-image setting depends not only on the availability of a reference image, but also on the contextual relationship between the reference and degraded images. \textit{Intra-image} comparison emphasizes fine-grained degradation comparison, whereas \textit{Inter-temporal} comparison may involve both degradation differences and temporal or scene-level changes. Note as expected, as shown in Table~\ref{tab: perceptionmulti_all}, \textit{Inter-temporal} comparison is not consistently harder than \textit{Intra-image} comparison: Qwen3VL-30B-A3B improves from 66.60\% to 82.01\%, and KimiVL-A3B increases from 54.95\% to 71.19\%. This suggests that some models may benefit from salient global changes in cross-temporal pairs, rather than relying purely on fine-grained degradation evidence. In contrast, \textit{Intra-image} pairs share nearly identical semantic content, making them better suited to exposing models' ability to compare low-level degradations such as noise, blur, compression artifacts, and missing regions. Thus, paired-image evaluation is not merely a test of whether a reference image is provided, but a test of whether models can establish controllable cross-image visual attribution by separating degradation-induced changes from scene or temporal variation.

\begin{table*}[!t]
    
    \centering
    \footnotesize
    \renewcommand\arraystretch{1} 
    \renewcommand\tabcolsep{2pt}
    
    \caption{Complete Quantitative results of VLMs on low-level description for single images.}
    \label{descriptionsingle}
    
    \newcolumntype{C}{>{\centering\arraybackslash}p{0.9cm}}
    
    \resizebox{\textwidth}{!}{%
    \begin{tabular}{l | CCCC | CCCC | CCCC | C}
    \toprule
    
    \textbf{Dimensions} & \multicolumn{4}{c|}{\textbf{Completeness}} & \multicolumn{4}{c|}{\textbf{Correctness}} & \multicolumn{4}{c|}{\textbf{Faithfulness}} & \multirow{2}{*}{\textit{Sum$\uparrow$}} \\ 
    
    \cdashline{1-13}
    
    \textbf{Model} & \textit{P0} & \textit{P1} & \textit{P2} & \textit{Score$\uparrow$} & \textit{P0} & \textit{P1} & \textit{P2} & \textit{Score$\uparrow$} & \textit{P0} & \textit{P1} & \textit{P2} & \textit{Score$\uparrow$} & \\ 
    \midrule 
    
    \textcolor{gray}{\textit{Human Evaluator}} & \textcolor{gray}{4.58\%} & \textcolor{gray}{21.25\%} & \textcolor{gray}{74.17\%} & \textcolor{gray}{1.70} & \textcolor{gray}{15.42\%} & \textcolor{gray}{12.08\%} & \textcolor{gray}{72.50\%} & \textcolor{gray}{1.57} & \textcolor{gray}{0.83\%} & \textcolor{gray}{6.25\%} & \textcolor{gray}{92.92\%} & \textcolor{gray}{1.92} & \textcolor{gray}{5.19} \\
    \noalign{\vskip 1pt}
    \cdashline{1-14}
    \noalign{\vskip 1pt}
    
    GPT-5.4 & 3.33\% & 9.50\% & 87.17\% & \textbf{1.84} & 25.17\% & 2.67\% & 72.17\% & \textbf{1.47} & 18.83\% & 6.00\% & 75.17\% & 1.56 & \textbf{4.87} \\
    Qwen3.5-Plus & 10.50\% & 7.83\% & 81.67\% & 1.71 & 35.33\% & 0.67\% & 64.00\% & 1.29 & 28.17\% & 5.83\% & 66.00\% & 1.38 & 4.38 \\
    Gemini-3.1-pro-preview & 8.67\% & 9.50\% & 81.83\% & \underline{1.73} & 27.17\% & 0.83\% & 72.00\% & \underline{1.45} & 20.33\% & 6.67\% & 73.00\% & 1.53 & 4.71 \\
    
    \midrule
    Ovis1.6-9B\cite{ovis1.6} & 5.00\% & 52.67\% & 42.33\% & 1.37 & 32.33\% & 2.83\% & 64.83\% & 1.32 & 3.83\% & 3.83\% & 92.33\% & 1.89 & 4.58 \\
    Ovis2.5-9B\cite{ovis2.5} & 8.17\% & 45.33\% & 46.50\% & 1.38 & 18.67\% & 18.67\% & 62.67\% & 1.44 & 0.00\% & 7.50\% & 92.50\% & \textbf{1.93} & \underline{4.75} \\
    Llama-3.2-11B\cite{llama3.2} & 33.33\% & 54.17\% & 12.50\% & 0.79 & 67.00\% & 3.17\% & 29.83\% & 0.63 & 39.33\% & 10.00\% & 50.67\% & 1.11 & 2.53 \\
    Molmo-7B-D\cite{molmo} & 30.67\% & 46.17\% & 23.17\% & 0.93 & 59.00\% & 1.00\% & 40.00\% & 0.81 & 5.33\% & 1.00\% & 93.67\% & 1.88 & 3.62 \\
    Phi4-vision\cite{phi4} & 19.67\% & 55.67\% & 24.67\% & 1.05 & 62.33\% & 0.67\% & 37.00\% & 0.75 & 11.33\% & 4.17\% & 84.50\% & 1.73 & 3.53 \\
    SenseNova-SI-1.1-8B\cite{sensenova} & 13.67\% & 59.67\% & 26.67\% & 1.13 & 63.33\% & 0.33\% & 36.33\% & 0.73 & 17.67\% & 2.67\% & 79.67\% & 1.62 & 3.48 \\
    DeepseekVL-7B\cite{deepseekvl} & 18.50\% & 63.00\% & 18.50\% & 1.00 & 70.17\% & 0.83\% & 29.00\% & 0.59 & 23.83\% & 9.00\% & 67.17\% & 1.43 & 3.02 \\
    LLaVA1.6-7B\cite{llava} & 38.83\% & 58.33\% & 2.83\% & 0.64 & 90.83\% & 0.50\% & 8.67\% & 0.18 & 71.33\% & 16.00\% & 12.67\% & 0.41 & 1.23 \\
    LLaVA1.6-13B\cite{llava} & 17.00\% & 68.83\% & 14.17\% & 0.97 & 58.17\% & 2.83\% & 39.00\% & 0.81 & 7.50\% & 5.50\% & 87.00\% & 1.79 & 3.57 \\
    LLaVA2-ov-7B\cite{llava-ov} & 28.67\% & 50.67\% & 20.67\% & 0.92 & 39.17\% & 2.67\% & 58.17\% & 1.19 & 9.33\% & 5.00\% & 85.67\% & 1.76 & 3.87 \\
    InternVL3.5-8B\cite{internvl3.5} & 9.50\% & 61.67\% & 28.83\% & 1.19 & 56.67\% & 1.17\% & 42.17\% & 0.85 & 10.33\% & 5.83\% & 83.83\% & 1.74 & 3.78 \\
    InternVL3.5-14B\cite{internvl3.5} & 8.00\% & 63.67\% & 28.33\% & 1.20 & 59.17\% & 1.00\% & 39.83\% & 0.81 & 4.17\% & 4.00\% & 91.83\% & 1.88 & 3.89 \\
    InternVL3.5-38B\cite{internvl3.5} & 17.83\% & 63.67\% & 18.50\% & 1.01 & 64.17\% & 0.83\% & 35.00\% & 0.71 & 10.00\% & 6.67\% & 83.33\% & 1.73 & 3.45 \\
    InternVL3.5-30B-A3B\cite{internvl3.5} & 24.50\% & 53.50\% & 22.00\% & 0.97 & 66.83\% & 0.33\% & 32.83\% & 0.66 & 18.00\% & 3.83\% & 78.17\% & 1.60 & 3.23 \\
    Qwen3VL-8B\cite{qwen3vl} & 20.50\% & 35.17\% & 44.33\% & 1.24 & 50.00\% & 0.50\% & 49.50\% & 0.99 & 7.50\% & 0.17\% & 92.33\% & 1.85 & 4.08 \\
    Qwen3VL-32B\cite{qwen3vl} & 19.17\% & 29.17\% & 51.67\% & 1.32 & 46.50\% & 0.83\% & 52.67\% & 1.06 & 4.33\% & 0.33\% & 95.33\% & \underline{1.91} & 4.29 \\
    Qwen3VL-30B-A3B\cite{qwen3vl} & 22.50\% & 25.50\% & 52.00\% & 1.29 & 45.50\% & 0.00\% & 54.50\% & 1.09 & 9.00\% & 0.33\% & 90.67\% & 1.82 & 4.20 \\
    KimiVL-A3B\cite{kimivl} & 40.83\% & 42.67\% & 16.50\% & 0.76 & 59.83\% & 2.50\% & 37.67\% & 0.78 & 43.83\% & 3.83\% & 52.33\% & 1.08 & 2.62 \\
    
    \midrule
    GeoChat\cite{geochat} & 24.00\% & 69.33\% & 6.67\% & 0.83 & 81.83\% & 4.67\% & 13.50\% & 0.32 & 47.83\% & 14.17\% & 38.00\% & 0.90 & 2.05 \\
    TEOChat\cite{teochat} & 16.17\% & 83.50\% & 0.33\% & 0.84 & 76.67\% & 7.50\% & 15.83\% & 0.39 & 14.33\% & 33.00\% & 52.67\% & 1.38 & 2.61 \\
    EarthDial\cite{soni2025earthdial} & 78.17\% & 21.83\% & 0.00\% & 0.22 & 81.00\% & 3.33\% & 15.67\% & 0.35 & 70.17\% & 7.50\% & 22.33\% & 0.52 & 1.09 \\
    LHRS-Bot\cite{lhrs} & 40.83\% & 58.67\% & 0.50\% & 0.60 & 89.00\% & 4.00\% & 7.00\% & 0.18 & 24.00\% & 43.83\% & 32.17\% & 1.08 & 1.86 \\
    LHRS-Bot-nova\cite{li2025lhrsbotnava} & 87.67\% & 11.67\% & 0.67\% & 0.13 & 93.33\% & 1.17\% & 5.50\% & 0.12 & 74.67\% & 6.00\% & 19.33\% & 0.45 & 0.70 \\
    
    \midrule
    DepictQA\cite{depictqa} & 4.17\% & 41.83\% & 54.00\% & 1.50 & 49.33\% & 3.83\% & 46.83\% & 0.97 & 23.17\% & 34.67\% & 42.17\% & 1.19 & 3.66 \\
    Q-Instruct-LLaVA1.5-13B\cite{qinstruct} & 15.83\% & 62.17\% & 22.00\% & 1.06 & 43.17\% & 6.50\% & 50.33\% & 1.07 & 2.67\% & 76.33\% & 21.00\% & 1.18 & 3.31 \\
    Q-Instruct-LLaVA1.5-7B\cite{qinstruct} & 17.83\% & 59.83\% & 22.33\% & 1.04 & 44.50\% & 3.83\% & 51.67\% & 1.07 & 6.33\% & 66.17\% & 27.50\% & 1.21 & 3.32 \\
    
    \bottomrule
    \end{tabular}%
    }
    \vspace{-4mm}
\end{table*}

\subsection{Complete SenseDescription Results}
\begin{table*}[!t]
    
    \centering
    \footnotesize
    \renewcommand\arraystretch{1} 
    \renewcommand\tabcolsep{2pt}
    
    \caption{Complete Quantitative results of VLMs on low-level description for paired images.}
    \label{descriptionmulti}
    
    \newcolumntype{C}{>{\centering\arraybackslash}p{0.9cm}}
    
    \resizebox{\textwidth}{!}{%
    \begin{tabular}{l | CCCC | CCCC | CCCC | C}
    \toprule
    
    \textbf{Dimensions} & \multicolumn{4}{c|}{\textbf{Completeness}} & \multicolumn{4}{c|}{\textbf{Correctness}} & \multicolumn{4}{c|}{\textbf{Faithfulness}} & \multirow{2}{*}{\textit{Sum$\uparrow$}} \\ 
    
    \cdashline{1-13}
    
    \textbf{Model} & \textit{P0} & \textit{P1} & \textit{P2} & \textit{Score$\uparrow$} & \textit{P0} & \textit{P1} & \textit{P2} & \textit{Score$\uparrow$} & \textit{P0} & \textit{P1} & \textit{P2} & \textit{Score$\uparrow$} & \\ 
    \midrule 
    
    \textcolor{gray}{\textit{Human Evaluator}} & \textcolor{gray}{3.92\%} & \textcolor{gray}{21.25\%} & \textcolor{gray}{74.83\%} & \textcolor{gray}{1.70} & \textcolor{gray}{22.17\%} & \textcolor{gray}{11.00\%} & \textcolor{gray}{66.83\%} & \textcolor{gray}{1.45} & \textcolor{gray}{8.75\%} & \textcolor{gray}{18.33\%} & \textcolor{gray}{72.92\%} & \textcolor{gray}{1.64} & \textcolor{gray}{4.79} \\
    \noalign{\vskip 1pt}
    \cdashline{1-14}
    \noalign{\vskip 1pt}
    
    GPT-5.4 & 1.83\% & 21.33\% & 76.83\% & \textbf{1.75} & 42.33\% & 2.50\% & 55.17\% & 1.13 & 36.83\% & 4.33\% & 58.83\% & 1.22 & 4.10 \\
    Qwen3.5-Plus & 5.83\% & 16.17\% & 78.00\% & \underline{1.72} & 31.00\% & 0.33\% & 68.67\% & \underline{1.38} & 29.50\% & 1.50\% & 69.00\% & 1.40 & \underline{4.50} \\
    Gemini-3.1-pro-preview & 2.50\% & 24.83\% & 72.67\% & 1.70 & 25.33\% & 0.17\% & 74.50\% & \textbf{1.49} & 24.17\% & 1.33\% & 74.50\% & 1.50 & \textbf{4.69} \\
    
    \midrule
    Ovis1.6-9B\cite{ovis1.6} & 11.83\% & 83.83\% & 4.33\% & 0.93 & 65.33\% & 0.67\% & 34.00\% & 0.69 & 49.33\% & 10.83\% & 39.83\% & 0.91 & 2.53 \\
    Ovis2.5-9B\cite{ovis2.5} & 26.04\% & 37.56\% & 36.39\% & 1.10 & 15.33\% & 34.67\% & 50.00\% & 1.35 & 0.00\% & 23.50\% & 76.50\% & \textbf{1.76} & 4.21 \\
    Llama-3.2-11B\cite{llama3.2} & 12.83\% & 76.83\% & 10.33\% & 0.97 & 65.00\% & 2.83\% & 32.17\% & 0.67 & 48.67\% & 8.67\% & 42.67\% & 0.94 & 2.58 \\
    Molmo-7B-D\cite{molmo} & 33.17\% & 66.67\% & 0.17\% & 0.67 & 96.83\% & 0.67\% & 2.50\% & 0.06 & 63.50\% & 29.50\% & 7.00\% & 0.43 & 1.16 \\
    Phi4-vision\cite{phi4} & 18.00\% & 81.17\% & 0.83\% & 0.83 & 94.83\% & 1.33\% & 3.83\% & 0.09 & 46.33\% & 44.83\% & 8.83\% & 0.62 & 1.54 \\
    SenseNova-SI-1.1-8B\cite{sensenova} & 71.33\% & 28.67\% & 0.00\% & 0.29 & 91.50\% & 0.17\% & 8.33\% & 0.17 & 67.50\% & 18.67\% & 13.83\% & 0.46 & 0.92 \\
    DeepseekVL-7B\cite{deepseekvl} & 47.00\% & 52.67\% & 0.33\% & 0.53 & 93.17\% & 1.00\% & 5.83\% & 0.13 & 60.67\% & 17.50\% & 21.83\% & 0.61 & 1.27 \\
    LLaVA2-ov-7B\cite{llava-ov} & 8.33\% & 82.33\% & 9.33\% & 1.01 & 76.33\% & 1.17\% & 22.50\% & 0.46 & 46.33\% & 11.83\% & 41.83\% & 0.95 & 2.42 \\
    InternVL3.5-8B\cite{internvl3.5} & 27.33\% & 65.33\% & 7.33\% & 0.80 & 81.00\% & 1.33\% & 17.67\% & 0.37 & 50.50\% & 24.00\% & 25.50\% & 0.75 & 1.92 \\
    InternVL3.5-14B\cite{internvl3.5} & 38.33\% & 52.83\% & 8.83\% & 0.70 & 81.83\% & 0.83\% & 17.33\% & 0.35 & 53.83\% & 18.33\% & 27.83\% & 0.74 & 1.79 \\
    InternVL3.5-38B\cite{internvl3.5} & 33.17\% & 49.83\% & 17.00\% & 0.84 & 66.67\% & 3.00\% & 30.33\% & 0.64 & 48.50\% & 13.83\% & 37.67\% & 0.89 & 2.37 \\
    InternVL3.5-30B-A3B\cite{internvl3.5} & 29.83\% & 58.67\% & 11.50\% & 0.82 & 77.00\% & 4.00\% & 19.00\% & 0.42 & 44.33\% & 25.17\% & 30.50\% & 0.86 & 2.10 \\
    Qwen3VL-8B\cite{qwen3vl} & 9.83\% & 57.33\% & 32.83\% & 1.23 & 66.83\% & 0.17\% & 33.00\% & 0.66 & 34.67\% & 6.67\% & 58.67\% & 1.24 & 3.13 \\
    Qwen3VL-32B\cite{qwen3vl} & 13.17\% & 27.33\% & 59.50\% & 1.46 & 43.33\% & 2.50\% & 54.17\% & 1.11 & 16.83\% & 5.33\% & 77.83\% & \underline{1.61} & 4.18 \\
    Qwen3VL-30B-A3B\cite{qwen3vl} & 21.33\% & 42.00\% & 36.67\% & 1.15 & 58.67\% & 1.67\% & 39.67\% & 0.81 & 35.50\% & 7.00\% & 57.50\% & 1.22 & 3.18 \\
    KimiVL-A3B\cite{kimivl} & 27.00\% & 48.83\% & 24.17\% & 0.97 & 56.50\% & 2.67\% & 40.83\% & 0.84 & 37.17\% & 16.50\% & 46.33\% & 1.09 & 2.90 \\
    
    \midrule
    TEOChat\cite{teochat} & 63.00\% & 36.67\% & 0.33\% & 0.37 & 88.50\% & 3.83\% & 7.67\% & 0.19 & 67.67\% & 19.83\% & 12.50\% & 0.45 & 1.01 \\
    EarthDial\cite{soni2025earthdial} & 100.00\% & 0.00\% & 0.00\% & 0.00 & 99.83\% & 0.00\% & 0.17\% & 0.00 & 88.50\% & 0.00\% & 11.50\% & 0.23 & 0.23 \\
    
    \midrule
    DepictQA\cite{depictqa} & 19.17\% & 50.83\% & 30.00\% & 1.11 & 83.83\% & 1.17\% & 15.00\% & 0.31 & 11.83\% & 71.00\% & 17.17\% & 1.05 & 2.47 \\
    
    \bottomrule
    \end{tabular}%
    }
    \vspace{-4mm}
\end{table*}
To better interpret how models obtain their diagnostic description scores, we report the complete \textit{SenseDescription} results in Table~\ref{descriptionsingle} and Table~\ref{descriptionmulti} for the single-image and paired-image settings, respectively. Beyond the averaged scores, these tables provide the full distributions over $P_0$, $P_1$, and $P_2$, corresponding to failure, partial alignment, and full alignment for each evaluation dimension. This distributional view is important for open-ended evaluation, because similar average scores may reflect different response patterns: a model may be consistently partially correct, or alternate between fully grounded descriptions and complete failures. Thus, the complete results provide a clearer view of the reliability and failure modes of VLMs in low-level diagnostic description.

\paragraph{Distributional results reveal a partial-diagnosis pattern.}
The complete distributions show that many models concentrate on $P_1$ rather than consistently reaching $P_2$, indicating that they often capture partial visual evidence but fail to complete the full degradation diagnosis. For example, in the single-image setting,Table~\ref{descriptionsingle} shows that InternVL3.5-8B obtains 61.67\% on $P_1$ but only 28.83\% on $P_2$ for Completeness; in the paired-image setting, Table~\ref{descriptionmulti} shows that Ovis1.6-9B reaches 83.83\% on $P_1$ but only 4.33\% on $P_2$. This suggests that current VLMs are not simply unable to describe degradations, but often remain partially correct: they identify some visible cues while failing to jointly cover degradation type, severity, and impact on remote-sensing interpretation.

\paragraph{Similar aggregate scores can hide different error structures.} The full distributions show that a single \textit{Sum} score cannot fully characterize diagnostic behavior. For example, Table~\ref{descriptionmulti} shows that GPT-5.4 and Ovis2.5-9B obtain comparable \textit{Sum} scores of 4.10 and 4.21, but with different distributions: GPT-5.4 achieves higher $P_2$ in \textit{Completeness} (76.83\%), whereas Ovis2.5-9B achieves higher $P_2$ in \textit{Faithfulness} (76.50\%) but much lower $P_2$ in \textit{Completeness} (36.39\%). This suggests that similar aggregate scores may reflect different bottlenecks, making the $P_0/P_1/P_2$ distributions necessary for interpreting open-ended diagnostic performance.

\subsection{Distortion-wise Analysis}
\begin{table*}[!t]
    \centering
    \footnotesize
    \renewcommand\arraystretch{1}
    \setlength{\tabcolsep}{6pt}
    
    \caption{Distortion-wise quantitative results of VLMs on low-level perception for single images.}
    \label{tab:distortion_single}
    
    \resizebox{0.75\textwidth}{!}{%
    \begin{tabular}{l*{7}{c}}
    \toprule
    \textbf{Model} & \textit{Blur$\uparrow$} & \textit{Cloud$\uparrow$} & \textit{Compre.$\uparrow$} & \textit{Correct.$\uparrow$} & \textit{Missing$\uparrow$} & \textit{Noise$\uparrow$} & \textit{Avg.$\uparrow$} \\
    \midrule
    
    GPT-5.4 & 0.710 & \underline{0.652} & 0.563 & 0.606 & 0.578 & 0.677 & 0.631 \\
    Qwen3.5-Plus & 0.708 & 0.573 & 0.631 & 0.630 & 0.580 & 0.666 & 0.631 \\
    Gemini-3.1-pro-preview & \textbf{0.752} & \textbf{0.675} & \textbf{0.812} & \textbf{0.814} & \textbf{0.840} & \textbf{0.819} & \textbf{0.785} \\
    
    \midrule
    Ovis1.6-9B\cite{ovis1.6} & \underline{0.720} & 0.610 & 0.579 & 0.561 & 0.498 & 0.653 & 0.604 \\
    Ovis2.5-9B\cite{ovis2.5} & 0.553 & 0.368 & 0.371 & 0.369 & 0.433 & 0.522 & 0.436 \\
    Llama-3.2-11B\cite{llama3.2} & 0.558 & 0.343 & 0.443 & 0.468 & 0.387 & 0.503 & 0.450 \\
    Molmo-7B-D\cite{molmo} & 0.508 & 0.367 & 0.561 & 0.464 & 0.338 & 0.487 & 0.454 \\
    Phi4-vision\cite{phi4} & 0.465 & 0.310 & 0.343 & 0.388 & 0.365 & 0.549 & 0.403 \\
    SenseNova-SI-1.1-8B\cite{sensenova} & 0.413 & 0.395 & 0.473 & 0.513 & 0.502 & 0.587 & 0.480 \\
    DeepseekVL-7B\cite{deepseekvl} & 0.542 & 0.388 & 0.418 & 0.463 & 0.313 & 0.551 & 0.446 \\
    LLaVA1.6-7B\cite{llava} & 0.493 & 0.303 & 0.491 & 0.475 & 0.312 & 0.496 & 0.428 \\
    LLaVA2-ov-7B\cite{llava-ov} & 0.712 & 0.612 & \underline{0.663} & \underline{0.651} & 0.550 & 0.686 & \underline{0.646} \\
    InternVL3.5-8B\cite{internvl3.5} & 0.495 & 0.327 & 0.477 & 0.461 & 0.473 & 0.509 & 0.457 \\
    InternVL3.5-14B\cite{internvl3.5} & 0.445 & 0.465 & 0.460 & 0.519 & 0.420 & 0.468 & 0.463 \\
    InternVL3.5-38B\cite{internvl3.5} & 0.513 & 0.512 & 0.407 & 0.521 & 0.485 & 0.526 & 0.494 \\
    InternVL3.5-30B-A3B\cite{internvl3.5} & 0.510 & 0.432 & 0.564 & 0.533 & 0.508 & 0.540 & 0.514 \\
    Qwen3VL-8B\cite{qwen3vl} & 0.663 & 0.513 & 0.487 & 0.561 & 0.518 & 0.647 & 0.565 \\
    Qwen3VL-32B\cite{qwen3vl} & 0.718 & 0.588 & 0.663 & 0.575 & \underline{0.597} & \underline{0.689} & 0.638 \\
    Qwen3VL-30B-A3B\cite{qwen3vl} & 0.613 & 0.437 & 0.602 & 0.565 & 0.497 & 0.647 & 0.560 \\
    KimiVL-A3B\cite{kimivl} & 0.570 & 0.465 & 0.540 & 0.633 & 0.483 & 0.624 & 0.553 \\
    
    \midrule
    GeoChat\cite{geochat} & 0.358 & 0.310 & 0.322 & 0.396 & 0.270 & 0.373 & 0.338 \\
    TEOChat\cite{teochat} & 0.412 & 0.412 & 0.336 & 0.445 & 0.310 & 0.445 & 0.393 \\
    EarthDial\cite{soni2025earthdial} & 0.457 & 0.518 & 0.479 & 0.534 & 0.595 & 0.476 & 0.510 \\
    LHRS-Bot\cite{lhrs} & 0.322 & 0.430 & 0.290 & 0.451 & 0.362 & 0.417 & 0.379 \\
    LHRS-Bot-nova\cite{li2025lhrsbotnava} & 0.305 & 0.265 & 0.240 & 0.324 & 0.353 & 0.410 & 0.316 \\
    
    \bottomrule
    \end{tabular}%
    }
    \vspace{-4mm}
\end{table*}
\begin{table*}[!t]
    \centering
    \footnotesize
    \renewcommand\arraystretch{1}
    \setlength{\tabcolsep}{6pt}
    
    \caption{Distortion-wise quantitative results of VLMs on low-level perception for paired images.}
    \label{tab:distortion_paired}
    
    \resizebox{0.75\textwidth}{!}{%
    \begin{tabular}{l*{7}{c}}
    \toprule
    \textbf{Model} & \textit{Blur$\uparrow$} & \textit{Cloud$\uparrow$} & \textit{Compre.$\uparrow$} & \textit{Correct.$\uparrow$} & \textit{Missing$\uparrow$} & \textit{Noise$\uparrow$} & \textit{Avg.$\uparrow$} \\
    \midrule
    
    GPT-5.4 & \textbf{0.764} & \underline{0.852} & 0.689 & \underline{0.806} & 0.709 & 0.834 & 0.776 \\
    Qwen3.5-Plus & \underline{0.752} & 0.824 & \underline{0.715} & 0.776 & \underline{0.755} & \underline{0.846} & \underline{0.778} \\
    Gemini-3.1-pro-preview & \underline{0.752} & \textbf{0.876} & \textbf{0.794} & \textbf{0.830} & \textbf{0.861} & \textbf{0.861} & \textbf{0.829} \\
    
    \midrule
    Ovis1.6-9B\cite{ovis1.6} & 0.609 & 0.467 & 0.495 & 0.470 & 0.439 & 0.606 & 0.515 \\
    Ovis2.5-9B\cite{ovis2.5} & 0.576 & 0.503 & 0.448 & 0.570 & 0.567 & 0.582 & 0.541 \\
    Llama-3.2-11B\cite{llama3.2} & 0.536 & 0.473 & 0.405 & 0.593 & 0.382 & 0.482 & 0.479 \\
    Molmo-7B-D\cite{molmo} & 0.542 & 0.515 & 0.580 & 0.739 & 0.630 & 0.496 & 0.584 \\
    Phi4-vision\cite{phi4} & 0.509 & 0.515 & 0.442 & 0.564 & 0.494 & 0.632 & 0.526 \\
    SenseNova-SI-1.1-8B\cite{sensenova} & 0.397 & 0.542 & 0.524 & 0.475 & 0.467 & 0.627 & 0.505 \\
    DeepseekVL-7B\cite{deepseekvl} & 0.524 & 0.536 & 0.508 & 0.502 & 0.403 & 0.507 & 0.497 \\
    LLaVA2-ov-7B\cite{llava-ov} & 0.621 & 0.524 & 0.558 & 0.459 & 0.439 & 0.616 & 0.536 \\
    InternVL3.5-8B\cite{internvl3.5} & 0.497 & 0.633 & 0.506 & 0.577 & 0.506 & 0.638 & 0.560 \\
    InternVL3.5-14B\cite{internvl3.5} & 0.530 & 0.697 & 0.544 & 0.652 & 0.512 & 0.588 & 0.587 \\
    InternVL3.5-38B\cite{internvl3.5} & 0.494 & 0.682 & 0.539 & 0.536 & 0.667 & 0.585 & 0.584 \\
    InternVL3.5-30B-A3B\cite{internvl3.5} & 0.579 & 0.712 & 0.692 & 0.632 & 0.748 & 0.642 & 0.668 \\
    Qwen3VL-8B\cite{qwen3vl} & 0.658 & 0.691 & 0.559 & 0.673 & 0.627 & 0.696 & 0.651 \\
    Qwen3VL-32B\cite{qwen3vl} & 0.727 & 0.803 & 0.620 & 0.700 & 0.736 & 0.792 & 0.730 \\
    Qwen3VL-30B-A3B\cite{qwen3vl} & 0.700 & 0.727 & 0.680 & 0.750 & 0.742 & 0.690 & 0.715 \\
    KimiVL-A3B\cite{kimivl} & 0.548 & 0.591 & 0.594 & 0.557 & 0.570 & 0.642 & 0.584 \\
    
    \midrule
    TEOChat\cite{teochat} & 0.336 & 0.430 & 0.308 & 0.364 & 0.255 & 0.333 & 0.338 \\
    EarthDial\cite{soni2025earthdial} & 0.336 & 0.539 & 0.400 & 0.348 & 0.400 & 0.368 & 0.399 \\
    
    \bottomrule
    \end{tabular}%
    }
    \vspace{-4mm}
\end{table*}
This section provides a distortion-wise analysis of low-level perception performance. Table~\ref{tab:distortion_single} and Table~\ref{tab:distortion_paired} report results under the single-image and paired-image settings, respectively, across six major distortion categories: \textit{blur}, \textit{cloud}, \textit{compression}, \textit{correction}, \textit{missing}, and \textit{noise}. We include this breakdown because averaged scores can obscure category-specific weaknesses, where a model may perform well on salient distortions but fail on subtle or domain-specific artifacts. This analysis therefore helps determine whether model performance reflects general low-level perception or sensitivity to particular degradation patterns.

\paragraph{Distortion categories show uneven difficulty.}
The distortion-wise results show that VLMs perceive different degradation categories unevenly. Overall, \textit{blur} and \textit{noise} are relatively easier for strong models, especially in the single-image setting. For example, Table~\ref{tab:distortion_single} shows that Qwen3VL-32B achieves 0.718 and 0.689 on \textit{blur} and \textit{noise}, respectively, while GPT-5.4 obtains 0.710 and 0.677. In contrast, \textit{cloud}, \textit{missing}, and some \textit{compression}/\textit{correction} degradations more often reduce performance, as they involve occlusion, structural loss, compression artifacts, or remote-sensing-specific imaging changes. Although paired images alleviate part of this difficulty, e.g., Gemini-3.1-pro-preview reaches 0.876, 0.861, and 0.861 on \textit{cloud}, \textit{missing}, and \textit{noise} in Table~\ref{tab:distortion_paired}, clear gaps remain across categories. This indicates that low-level degradation perception is not a uniform capability, but is strongly affected by the visual form and comparability of each distortion type.

\paragraph{Reference gain is distortion-dependent.}
The performance gain brought by paired images is not uniform across distortion categories, but strongly depends on the specific degradation type. As shown in Table~\ref{tab:distortion_single} and Table~\ref{tab:distortion_paired}, Qwen3VL-32B improves from 0.638 in the single-image setting to 0.730 in the paired-image setting, with large gains on \textit{cloud} (0.588 to 0.803), \textit{missing} (0.597 to 0.736), and \textit{noise} (0.689 to 0.792), but only a small gain on \textit{blur} (0.718 to 0.727). Similarly, InternVL3.5-30B-A3B improves from 0.432 to 0.712 on \textit{cloud} and from 0.508 to 0.748 on \textit{missing}. These results suggest that reference images are more helpful for degradations with clear spatial discrepancies or structural changes, such as cloud occlusion, missing regions, and noise, while providing more limited gains for \textit{blur}, which relies more on edge sharpness and fine-grained texture comparison. Thus, the advantage of the paired setting does not simply come from additional visual input, but depends on whether models can attribute cross-image differences to specific degradation types.

\section{Limitations \& Solutions}
\label{limitation}

SenseBench provides a systematic framework for evaluating low-level visual perception in remote sensing vision-language models. Nevertheless, its benchmark construction and evaluation protocol inevitably involve several practical constraints and potential sources of bias. We therefore discuss the main limitations below, along with the mitigation strategies adopted to improve the reliability of the benchmark.

\subsection{Biases in synthetic degradations.}
\paragraph{Limitation.} Part of SenseBench is constructed using algorithmically synthesized degradations. We acknowledge that such synthetic degradations may not fully match the distribution of real degraded remote sensing images. In particular, degradation algorithms may introduce \textbf{systematic biases}, causing synthesized samples to deviate from real-world degradation patterns. \textbf{Incidental biases} may also arise from parameter sampling, scene content, or unsuccessful synthesis cases, resulting in unnatural, ambiguous, or low-quality samples. These biases may affect benchmark data quality and, consequently, the reliability of model evaluation.

\paragraph{Current Solution.}
In real-world remote sensing scenarios, degradations are diverse, irregular, and difficult to exhaustively annotate at scale. It is therefore challenging to manually collect a large-scale benchmark that simultaneously covers broad geographic regions, diverse degradation types, and multiple severity levels. Controlled degradation synthesis has thus been widely adopted in prior IQA benchmarks, such as LIVE\cite{LIVE}, TID2013\cite{TID2013}, and KADID-10k\cite{kadid10k}. SenseBench follows this common practice while \emph{avoiding exclusive reliance on synthetic data}. For degradation types that can be reliably collected, such as cloud occlusion, SAR speckle noise, and SAR sidelobe artifacts, we manually collect real degraded samples. For synthesized degradation types, we design the generation procedures according to standard image degradation mechanisms, including noise, blur, compression, missing regions, cloud or haze contamination, and geometric or radiometric correction errors, thereby reducing potential \textbf{systematic bias}. To further mitigate \textbf{incidental bias}, we conduct human inspection, remove abnormal samples, and re-synthesize invalid cases, as described in Section~\ref{sec:quality_control}. Although these measures cannot fully close the gap between synthetic and real degradations, controlled synthesis provides a \emph{practical, reproducible, and scalable} solution under the current difficulty of large-scale real degradation annotation.

\subsection{Annotation bias in diagnostic descriptions}
\paragraph{Limitation.}
The open-ended diagnostic description task in SenseBench involves LLM-assisted annotation during data construction. Specifically, we use GPT-5.2 to assist in generating diagnostic descriptions. While this improves annotation efficiency, it may also introduce \textbf{model-specific writing patterns}, \textbf{prompt sensitivity}, or annotation errors if the generated outputs are used without verification.

\paragraph{Current solution.}
Our annotation process is not a fully automatic generation pipeline. During data construction, the LLM is provided with the clean image, the degraded image, and structured ground-truth metadata, including the degradation type, degradation name, and severity level. Therefore, the descriptions are generated under \emph{explicit visual and semantic grounding}, rather than through unconstrained generation. We further adopt a human--LLM collaborative workflow: the LLM is used to improve annotation efficiency and linguistic consistency, while human annotators review, correct, and filter the generated descriptions. This design enables scalable annotation while reducing the cost of fully manual writing and mitigating both \textbf{LLM-induced bias} and the subjective bias of individual annotators.

\subsection{Judge-model bias in open-ended evaluation}
\paragraph{Limitation.}
SenseBench evaluates open-ended diagnostic responses using Prometheus, a model designed for LLM-as-judge evaluation, rather than relying entirely on human scoring. We acknowledge that such judge-based evaluation may inherit \textbf{evaluator-model biases} and cannot fully substitute expert human assessment.

\paragraph{Current solution.}
Exhaustive human evaluation is impractical at the scale of our benchmark. With more than 20 evaluated models, 1,200 open-ended questions per model, and three scoring dimensions per response, the evaluation requires tens of thousands of scalar judgments, approaching 100k scores when repeated trials and invalid runs are included. Assuming one minute per score, this would require approximately 1,667 person-hours. We therefore adopt an LLM-as-judge protocol, which has also been used in prior open-ended multimodal evaluation benchmarks, such as Q-Bench~\cite{qbench}, MM-Vet~\cite{mmvet}, and LLaVA-Bench~\cite{llava}. To assess its reliability, we conduct a sampled human-alignment study in Section~\ref{evaluation_quality_control}. The results show that the LLM judge achieves high correlation with human scores, and that the sampled model ranking is consistent between human evaluation and LLM-based evaluation. These results suggest that our judge-based protocol provides a \emph{scalable and sufficiently reliable} approximation to human evaluation.

\section{Broader impacts}
\label{broader_impacts}

\textbf{SenseBench} may contribute positively to the development of more reliable VLMs for low-level visual perception and diagnostic description in remote sensing. By providing a systematic evaluation protocol for image degradations, it can support image selection, quality control, and reliability assessment in Earth observation workflows. As SenseBench is an evaluation benchmark rather than a deployed decision-making system, its direct negative societal impact is limited. Nevertheless, VLM-generated diagnostic descriptions may be misused if treated as authoritative quality assessments. Such over-reliance could affect downstream image selection or interpretation, especially in applications requiring expert domain knowledge. Therefore, SenseBench should be used as an \emph{assistive evaluation tool} and combined with human expert judgment, rather than used for autonomous decision-making.




\clearpage
\section*{NeurIPS Paper Checklist}

\begin{enumerate}

\item {\bf Claims}
    \item[] Question: Do the main claims made in the abstract and introduction accurately reflect the paper's contributions and scope?
    \item[] Answer: \answerYes{} 
    \item[] Justification: We present the contribution and scope of the SenseBench benchmark, and summarize the main experimental findings in the abstract and introduction.

\item {\bf Limitations}
    \item[] Question: Does the paper discuss the limitations of the work performed by the authors?
    \item[] Answer: \answerYes{} 
    \item[] Justification: The paper discuss the limitations of the work in \ref{limitation}.

\item {\bf Theory assumptions and proofs}
    \item[] Question: For each theoretical result, does the paper provide the full set of assumptions and a complete (and correct) proof?
    \item[] Answer: \answerNA{} 
    \item[] Justification: The paper does not include theoretical results, theorems, or formal proofs; it focuses on benchmark construction and empirical evaluation.

    \item {\bf Experimental result reproducibility}
    \item[] Question: Does the paper fully disclose all the information needed to reproduce the main experimental results of the paper to the extent that it affects the main claims and/or conclusions of the paper (regardless of whether the code and data are provided or not)?
    \item[] Answer: \answerYes{} 
    \item[] Justification: The paper provides the necessary information for reproducing the main experimental results. Specifically, the dataset construction process is described in Appendix~\ref{app:dataset_construction}, and the evaluation settings, prompts, answer parsing rules, and implementation details are provided in Appendix~\ref{evaluation_details}.

\item {\bf Open access to data and code}
    \item[] Question: Does the paper provide open access to the data and code, with sufficient instructions to faithfully reproduce the main experimental results, as described in supplemental material?
    \item[] Answer: \answerYes{} 
    \item[] Justification: The dataset and code are provided \href{https://github.com/Zhong-Chenchen/SenseBench}{\textcolor{blue}{here}}. We provide detailed evaluation protocols, implementation settings, and reproduction details in Appendix~\ref{evaluation_details}.

\item {\bf Experimental setting/details}
    \item[] Question: Does the paper specify all the training and test details (e.g., data splits, hyperparameters, how they were chosen, type of optimizer) necessary to understand the results?
    \item[] Answer: \answerYes{} 
    \item[] Justification: We provide the experimental settings in Section~\ref{04_experiment}, and more details are provided in Appendix~\ref{evaluation_details}.

\item {\bf Experiment statistical significance}
    \item[] Question: Does the paper report error bars suitably and correctly defined or other appropriate information about the statistical significance of the experiments?
    \item[] Answer: \answerNo{} 
    \item[] Justification: The experiments are conducted under a fixed benchmark, fixed prompts, and fixed inference configurations to reduce evaluation variability, so repeated evaluations under the same setting yield identical results.

\item {\bf Experiments compute resources}
    \item[] Question: For each experiment, does the paper provide sufficient information on the computer resources (type of compute workers, memory, time of execution) needed to reproduce the experiments?
    \item[] Answer: \answerYes{} 
    \item[] Justification: We report the compute resources and approximate inference cost in Appendix~\ref{evaluation_details}.
    
\item {\bf Code of ethics}
    \item[] Question: Does the research conducted in the paper conform, in every respect, with the NeurIPS Code of Ethics \url{https://neurips.cc/public/EthicsGuidelines}?
    \item[] Answer: \answerYes{} 
    \item[] Justification: We have reviewed the NeurIPS Code of Ethics and confirm that the research conforms to it. The annotators' remuneration is described in Section~\ref{sec:quality_control}; the data mainly come from open data sources in Google Earth Engine, whose Terms of Service\footnote{\url{https://earthengine.google.com/terms/}} allow use for non-commercial activities and research or educational publications.

\item {\bf Broader impacts}
    \item[] Question: Does the paper discuss both potential positive societal impacts and negative societal impacts of the work performed?
    \item[] Answer: \answerYes{} 
    \item[] Justification: We discuss the positive societal impact and potential negative impact of SenseBench in Section~\ref{broader_impacts}.
    
\item {\bf Safeguards}
    \item[] Question: Does the paper describe safeguards that have been put in place for responsible release of data or models that have a high risk for misuse (e.g., pre-trained language models, image generators, or scraped datasets)?
    \item[] Answer: \answerNA{} 
    \item[] Justification: The paper does not release high-risk models, image generators, or scraped personal datasets; it only releases a remote sensing evaluation benchmark and code.

\item {\bf Licenses for existing assets}
    \item[] Question: Are the creators or original owners of assets (e.g., code, data, models), used in the paper, properly credited and are the license and terms of use explicitly mentioned and properly respected?
    \item[] Answer: \answerYes{} 
    \item[] Justification: We use several publicly available assets in this work, including 29 vision-language models, remote sensing datasets, and open-source software libraries. The evaluated models are obtained from their official releases, and we follow their corresponding licenses and usage terms, such as MIT, Apache-2.0, or other model-specific licenses. For data construction, we use publicly accessible satellite imagery, including Sentinel-2 and Landsat-8, which are provided under open-access data policies. We respect the terms of use of all data providers and provide proper attribution to the original sources. We also use open-source implementation frameworks and libraries, including MS-SWiFT and Vllm, and credit them according to their respective licenses.

\item {\bf New assets}
    \item[] Question: Are new assets introduced in the paper well documented and is the documentation provided alongside the assets?
    \item[] Answer: \answerYes{} 
    \item[] Justification: We introduce SenseBench as a new benchmark and provide documentation for its taxonomy, dataset construction, statistics, evaluation protocols, and quality control in Appendix~\ref{app:dataset_construction}, Appendix~\ref{evaluation_details}, and Section~\ref{sec:quality_control}.

\item {\bf Crowdsourcing and research with human subjects}
    \item[] Question: For crowdsourcing experiments and research with human subjects, does the paper include the full text of instructions given to participants and screenshots, if applicable, as well as details about compensation (if any)? 
    \item[] Answer: \answerYes{} 
    \item[] Justification: The paper provides the annotation and quality-control instructions, annotator information, compensation details, and screenshots of the annotation interface in Section~\ref{sec:quality_control}.

\item {\bf Institutional review board (IRB) approvals or equivalent for research with human subjects}
    \item[] Question: Does the paper describe potential risks incurred by study participants, whether such risks were disclosed to the subjects, and whether Institutional Review Board (IRB) approvals (or an equivalent approval/review based on the requirements of your country or institution) were obtained?
    \item[] Answer: \answerNA{} 
    \item[] Justification: This paper does not involve research with human subjects.

\item {\bf Declaration of LLM usage}
    \item[] Question: Does the paper describe the usage of LLMs if it is an important, original, or non-standard component of the core methods in this research? Note that if the LLM is used only for writing, editing, or formatting purposes and does \emph{not} impact the core methodology, scientific rigor, or originality of the research, declaration is not required.
    \item[] Answer: \answerYes{} 
    \item[] Justification: We describe the use of LLMs as part of the core benchmark construction and evaluation process, including GPT-assisted diagnostic description generation and LLM-as-a-Judge evaluation, in Appendix~\ref{app:dataset_construction} and Appendix~\ref{evaluation_details}.

\end{enumerate}

\end{document}